\documentclass[10pt, compsoc, peerreview, journal]{IEEEtran}

\title{Generalised Mutual Information: a Framework for Discriminative Clustering}

\author{Louis~Ohl,Pierre-Alexandre~Mattei, Charles~Bouveyron, Warith~Harchaoui, Micka\"el~Leclercq, Arnaud~Droit, Fr\'ed\'eric~Precioso
\IEEEcompsocitemizethanks{\IEEEcompsocthanksitem L.~Ohl, P-A.~Mattei, C.~Bouveyron and F.~Precioso are with the Universit\'e C\^ote d'Azur, INRIA Maasai team, CNRS
\IEEEcompsocthanksitem M.~Leclercq, A.~Droit and L.~Ohl are with the Universit\'e Laval, CHU de Qu\'ebec Research Centre
\IEEEcompsocthanksitem W.~Harchaoui is with Jellysmack, Artificial Intelligence Lab in Paris}%
\thanks{Manuscript under review}}
\usepackage[utf8]{inputenc} 
\usepackage[T1]{fontenc}    
\usepackage[pdfencoding=auto]{hyperref}       
\usepackage{url}            
\usepackage{amsfonts}       
\usepackage{nicefrac}       
\usepackage{microtype}      
\usepackage{wrapfig}      
\usepackage{graphicx}
\usepackage{url}            
\usepackage{booktabs}       
\usepackage{microtype}
\usepackage{amsmath}        
\usepackage{amssymb}        
\usepackage{amsthm}         
\usepackage{mathtools}
\usepackage{xcolor}

\usepackage{multirow}       
\usepackage{multicol}       

\usepackage[caption=false,font=normalsize,labelfont=sf,textfont=sf]{subfig}      

\usepackage[nocompress]{cite}
\renewcommand{\vec}{\pmb}

\newcommand{\p}{p_\theta}

\newcommand{\E}{\mathbb{E}}
\newcommand{\I}{\mathcal{I}}

\newtheorem{proposition}{Proposition}[section]
\newtheorem{corollary}{Corollary}[section]

\newcommand{\pdata}{p_\textup{data}(\vec{x})}
\newcommand{\px}{p(\vec{x})}
\newcommand{\ya}{y_a}
\newcommand{\yb}{y_b}
\newcommand{\py}{\p(y)}
\newcommand{\pya}{\p(\ya)}
\newcommand{\pyb}{\p(\yb)}
\newcommand{\pyx}{\p(y|\vec{x})}
\newcommand{\pyxa}{\p(\ya|\vec{x})}
\newcommand{\pyxb}{\p(\yb|\vec{x})}
\newcommand{\pxy}{\p(\vec{x}|y)}
\newcommand{\pxya}{\p(\vec{x}|\ya)}
\newcommand{\pxyb}{\p(\vec{x}|\yb)}

\newcommand{\pykx}{p(y=k|\vec{x})}\newcommand{\pxyk}{p(\vec{x}|y=k)}

\newcommand{\xa}{\vec{x}_a}

\newcommand{\xb}{\vec{x}_b}

\newcommand{\PX}{p(\vec{x})}
\newcommand{\PY}{p(y)}

\newcommand{\PXY}{p(\vec{x}|y)}

\newcommand{\Pjoint}{p(\vec{x},y)}

\begin{document}
\IEEEtitleabstractindextext{%
\begin{abstract}
In the last decade, recent successes in deep clustering majorly involved the Mutual Information (MI) as an unsupervised objective for training neural networks with increasing regularisations. While the quality of the regularisations have been largely discussed for improvements, little attention has been dedicated to the relevance of MI as a clustering objective. In this paper, we first highlight how the maximisation of MI does not lead to satisfying clusters. We identified the Kullback-Leibler divergence as the main reason of this behaviour. Hence, we generalise the mutual information by changing its core distance, introducing the Generalised Mutual Information (GEMINI): a set of metrics for unsupervised neural network training. Unlike MI, some GEMINIs do not require regularisations when training as they are geometry-aware thanks to distances or kernels in the data space. Finally, we highlight that GEMINIs can automatically select a relevant number of clusters, a property that has been little studied in deep discriminative clustering context where the number of clusters is a priori unknown.
\end{abstract}
}
\maketitle
\IEEEpeerreviewmaketitle


\IEEEraisesectionheading{\section{Introduction}
\label{sec:introduction}}
\IEEEPARstart{C}{lustering} is a fundamental learning task which consists in separating data samples into several categories, each named cluster. This task hinges on two main questions concerning the assessment of correct clustering and the actual number of clusters that may be contained within the data distribution. However, this problem is ill-posed since a cluster lacks formal definitions which makes it a hard problem~\cite{kleinberg_impossibility_2003}.

Model-based algorithms make assumptions about the true distribution of the data as a result of some latent distribution of clusters~\cite{bouveyron_model_2019}. These techniques are able to find the most likely cluster assignment to data points. These models are usually generative, exhibiting an explicit assumption of the prior knowledge on the data.

Early deep models to perform clustering first relied on autoencoders, based on the belief that an encoding space holds satisfactory properties~\cite{xie_unsupervised_2016,ghasedi_dizaji_deep_2017,ji_invariant_2019}. However, the drawback of these architectures is that they do not guarantee that data samples which should meaningfully be far apart remain so in the feature space. Early models that dropped decoders notably used the Mutual Information (MI)~\cite{krause_discriminative_2010,hu_learning_2017} as an objective to maximise. The MI can be written in two ways, either as measure of dependency between two variables $\vec{x}$ and $y$, e.g. data distribution $\PX$ and cluster assignment $\PY$:
\begin{equation}
\label{eq:mutual_information_1}
\I(\vec{x};y) = D_\text{KL} (\Pjoint || \PX\PY),
\end{equation}
or as an expected distance between implied distributions and the overall data:
\begin{equation}
\label{eq:mutual_information_2}
\I(\vec{x};y)= \mathbb{E}_{\PY} [ D_\text{KL} (\PXY||\PX)],
\end{equation}
with $D_\text{KL}$ being the Kullback-Leibler (KL) divergence. Related works often relied on the notion of MI as a measure of coherence between cluster assignments and data distribution~\cite{hjelm_learning_2019}. Regularisation techniques were employed to leverage the potential of MI, mostly by specifying model invariances, for example with data augmentation~\cite{ji_invariant_2019}.

The maximisation of MI thus gave way to contrastive learning objectives which aim at learning stable representations of data through such invariance specifications~\cite{chen_simple_2020,caron_unsupervised_2020}. The contrastive loss maximises the similarity between the features of a sample and its augmentation, while decreasing the similarity with any other sample. Clustering methods also benefited from recent successful deep architectures~\cite{li_contrastive_2021,tao_clustering-friendly_2021,huang_deep_2020} by encompassing regularisations in the architecture. These methods correspond to discriminative clustering where we seek to directly infer cluster given the data distribution.
Initial methods also focused on alternate schemes, for example with curriculum learning~\cite{chang_deep_2017} to iteratively select relevant data samples for training, by alternating K-means cluster assignment with supervised learning using the inferred labels~\cite{caron_deep_2018}, or by proceeding to multiple distinct training steps~\cite{van_gansbeke_scan_2020,dang_nearest_2021,park_improving_2021}.

However, most of the methods above rarely discuss their robustness when the number of clusters to find is different from the amount of preexisting known classes. While previous work was essentially motivated by considering MI as a dependence measure, we explore in this paper the alternative definition of the MI as the expected distance between data distribution implied by the clusters and the entire data. We extend it to incorporate cluster-wise comparisons of implied distributions, and question the choice of the KL divergence with other possible statistical distances.

Throughout the introduction of the Generalised Mutual Information (GEMINI), the contributions of this paper are:

\begin{itemize}
\item A demonstration of how the maxima of MI are not sufficient criteria for clustering. This extends the contribution of~\cite{tschannen_mutual_2019} to the discrete case.
\item The introduction of a set of metrics called GEMINIs involving different distances between distributions which can incorporate prior knowledge on the geometry of the data. Some of these metrics do not require regularisations.
\item A highlight of the implicit selection of clusters from GEMINIs which allows to select a relevant number of cluster during training.
\item A Python package using only for clustering with GEMINI: \emph{gemclus}
\end{itemize}

\section{Is MI a good clustering objective?}
\label{sec:about_mi}
We consider in this section a dataset consisting in $N$ unlabelled samples $\mathcal{D}=\{\vec{x}_i\}_{i=1}^N$. We distinguish two major use cases of the mutual information: one where we measure the dependence between two continuous variables, as is the case in representation learning, and one where the random variable is discrete. In representation learning, the goal is to construct a continuous representation $\vec{z}$ extracted from the data $\vec{x}$ using a learnable distribution of parameters $\theta$. In clustering, samples $\vec{x}$ are assigned to the discrete variable $y$ through another learnable distribution.


\subsection{Representation learning}
\label{ssec:mi_representation}

Representation learning consists in finding high-level features $\vec{z}_i$ extracted from the data $\vec{x}_i$ in order to perform a \emph{downstream task}, e.g. clustering or classification. MI between $\vec{x}$ and $\vec{z}$ is a common choice for learning features\cite{hjelm_learning_2019}. However, estimating correctly MI between two random variables in continuous domains is often intractable when $p(\vec{x}|\vec{z})$ or $p(\vec{z}|\vec{x})$ is unknown, thus lower bounds are preferred, e.g. variational estimators such as MINE~\cite{belghazi_mine_2018}, $\I_\text{NCE}$\cite{van_den_oord_representation_2018},$\I_\text{BA}$\cite{barber_im_2003},$\I_\text{DV}$\cite{donsker_asymptotic_1983}. These bounds require more parameters: additional discriminator networks are also trained to make a distinction between data and features issued from the joint distribution or product of marginals. Most of these lower bounds often present high-variance such as $\I_\text{NJW}$\cite{nguyen_estimating_2010}. Poole et al.~\cite{poole_variational_2019} eventually bridged the gap between this high-variance estimators and the high-bias low-variance $\I_\text{NCE}$ by introducing $\I_\alpha$, an interpolated lower bound. Another common choice of loss function to train features are contrastive losses such as NT-XENT~\cite{chen_simple_2020} where the similarity between the features $\vec{z}_i$ from data $\vec{x}_i$ is maximised with the features $\tilde{\vec{z}}$ from a data-augmented $\tilde{\vec{x}}_i$ against any other features $\vec{z}_j$. Recently, Do et al.~\cite{do_clustering_2021} achieved excellent performances in single-stage methods by highlighting the link between the $\I_\text{NCE}$ estimator~\cite{van_den_oord_representation_2018} and contrastive learning losses. Representation learning therefore comes at the cost of a complex lower bound estimator on MI, which often requires data augmentation.
Moreover, it was noticed that the MI is hardly predictive of downstream tasks~\cite{tschannen_mutual_2019} when the variable $y$ is continuous, i.e. a high value of MI does not clarify whether the discovered representations are insightful with regards to the target of the downstream task.

\subsection{Discriminative clustering}
\label{sssec:mi_clustering}

The MI has been first used as an objective for learning discriminative clustering models~\cite{bridle_unsupervised_1992}. Associated architectures went from simple logistic regression~\cite{krause_discriminative_2010} to deeper architectures~\cite{hu_learning_2017,ji_invariant_2019}. Beyond architecture improvement, the MI maximisation was also carried with several regularisations. These regularisations include penalty terms such as weight decay~\cite{krause_discriminative_2010} or Virtual Adversarial Training (VAT)~\cite{hu_learning_2017,miyato_virtual_2018}). Data augmentation was further used to provide invariances in clustering, as well as specific architecture designs like auxiliary clustering heads~\cite{ji_invariant_2019}. Rewriting the MI in terms of entropies:
\begin{equation}
\label{eq:mi_kl_entropies}
\I (\vec{x};y) = \mathcal{H}(y) - \mathcal{H}(y|\vec{x})
\end{equation}
highlights a requirement for balanced clusters, through the cluster entropy term $\mathcal{H}(y)$. Indeed, a uniform distribution maximises the entropy. This hints that an unregularised discrete mutual information for clustering can possibly produce uniformly distributed clusters among samples, regardless of how close they could be. We highlight this claim in section~\ref{ssec:local_maxima}. As an example of regularisation impact: maximising the MI with $\ell_2$ constraint can be equivalent to a soft and regularised K-Means in a feature space~\cite{jabi_deep_2019}. In clustering, the number of clusters to find is usually not known in advance. Therefore, an interesting clustering algorithm should be able to find a relevant number of clusters, i.e. perform model selection. However, model selection for parametric deep clustering models is expensive~\cite{ronen_deepdpm_2022}. Cluster selection through MI maximisation has been little studied in related works, since experiments usually tasked models to find the (supervised) classes of datasets. Furthermore, the literature diverged towards deep learning methods focusing mainly on images, yet rarely on other type of data such as tabular data~\cite{min_survey_2018}.

\subsection{Maximising the MI can lead to bad decision boundaries}
\label{ssec:local_maxima}

\begin{figure}
    \centering
    \subfloat[Good decision boundary]{
        \includegraphics[width=0.45\linewidth]{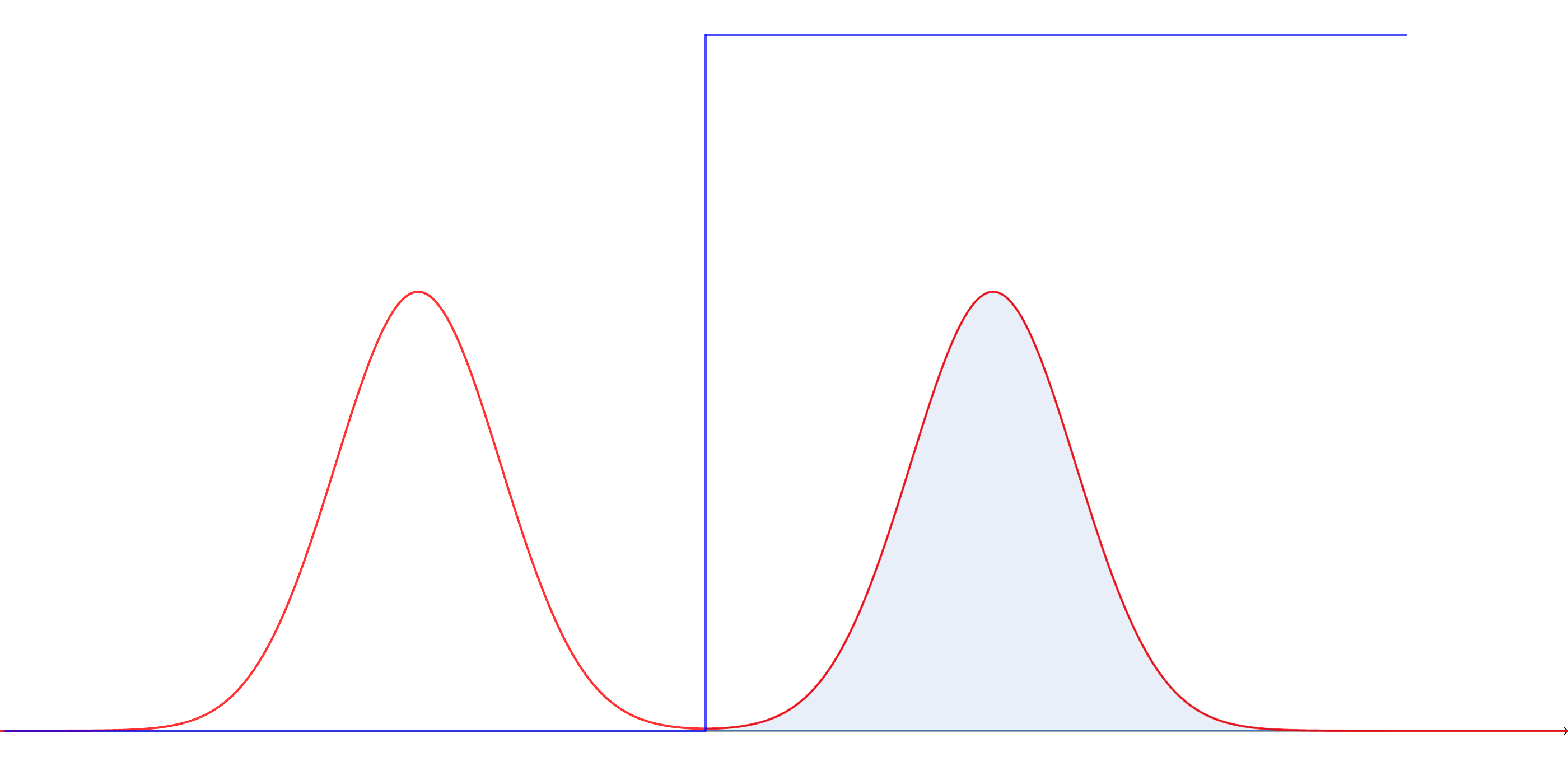}
        \label{sfig:good_decision_boundary}
    }\hfil
    \subfloat[Misplaced boundary]{
        \includegraphics[width=0.45\linewidth]{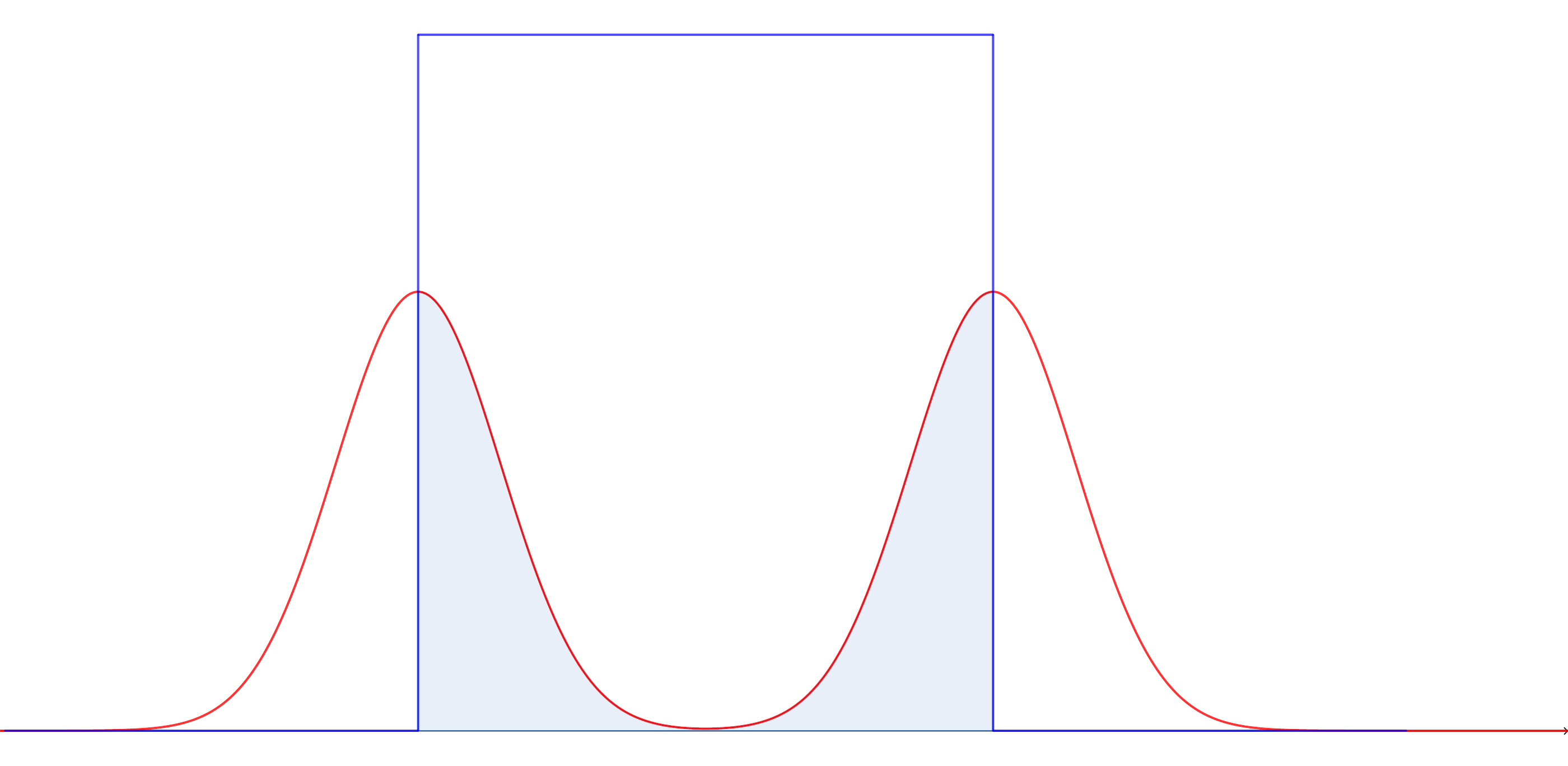}
        \label{sfig:odd_decision_boundary}
    }
    \caption{Example of maximised MI for a Gaussian mixture $\frac{1}{2} \mathcal{N}(\mu_0, \sigma^2)+\frac{1}{2}\mathcal{N}(\mu_1,\sigma^2)$. It is clear that figure \ref{sfig:good_decision_boundary} presents the best decision boundary and posterior between the two Gaussian distributions. Yet, as the posterior turns sharper, the difference between both MIs converges to 0.}
    \label{fig:example_good_odd_mi}
\end{figure}

Maximising the MI directly can be a poor objective: a high MI value is not necessarily predictive of the quality of the features regarding downstream tasks~\cite{tschannen_mutual_2019} when $y$ is continuous. We support a similar argument for the case where the data $\vec{x}$ is a continuous random variable and the cluster assignment $y$ a categorical variable. Indeed, the MI can be maximised by setting appropriately a sharp decision boundary which partitions evenly the data, i.e. when the distribution $\pyx$ converges to a Dirac distribution. This reasoning can be seen in the entropy-based formulation of the MI (Eq.~\ref{eq:mi_kl_entropies}): any sharp decision boundary minimises the negative conditional entropy, while ensuring balanced clusters maximises the entropy of cluster proportions. Consider for example Figure~\ref{fig:example_good_odd_mi}, where a mixture of Gaussian distributions with equal variances is separated by a sharp decision boundary. We highlight that both models will have the same mutual information on condition that the misplaced decision boundary of Figure~\ref{sfig:odd_decision_boundary} splits evenly the dataset (see Appendix~\ref{app:mi_convergence}).

Globally, MI misses the idea in clustering that any two points close to one another may be in the same cluster according to some chosen distance. Hence regularisations are required to ensure this constraint. An early sketch of these insights was mentioned by Bridle et al.~\cite{bridle_unsupervised_1992} or Corduneanu et al.~\cite{corduneanu_information_2012}. The non-predictiveness of MI was as well recently empirically highlighted by Zhang and Boykov~\cite{zhang_revisiting_2023} in discrete cases. This can be also be seen as a problem of invariance of the conditional distribution in low density areas~\cite{corduneanu_information_2012}.

\section{Extending the MI to the GEMINI}
\label{sec:gemini}
Given the identified limitations of MI, we now describe the discriminative clustering framework based on the expected distance equation of the mutual information. We then detail the different statistical distances we can use to extend MI to the Generalised Mutual Information (GEMINI).

\subsection{The discriminative clustering framework for GEMINIs}
\label{ssec:discriminative_clustering}

We only consider two random variables: the data $\vec{x}$ which can be continuous or discrete and the cluster assignment $y$ which is discrete. Instead of viewing the mutual information as a dependence-seeking objective (Eq.~\ref{eq:mutual_information_1}), we view it as a clustering objective that aims at separating the data distribution given cluster assignments $p(\vec{x}|y)$ from the data distribution $p(\vec{x})$ according to the KL divergence:
\begin{equation}\label{eq:base_mi}
    \I(\vec{x};y) = \E_{y\sim p(y)} \left[ D_\text{KL}(p(\vec{x}|y)\|p(\vec{x}))\right].
\end{equation}
To highlight the discriminative clustering design, we explicitly do not set any hypothesis on the data distribution by writing $\pdata$. The only part of the model that we design is a conditional distribution $\pyx$ that assigns a cluster $y$ to a sample $\vec{x}$ using the parameters $\theta$ of some learnable function $\psi$~\cite{minka_discriminative_2005}:
\begin{equation}\label{eq:conditional_model}
y|\vec{x} \sim \text{Categorical}(\psi_\theta(\vec{x})).
\end{equation}
This learnable function $\psi_\theta$ can typically be a neural network of adequate design regarding the data, e.g. a CNN, or a logistic regression. Consequently, the cluster proportions are controlled by $\theta$ because $\py=\mathbb{E}[\pyx]$ and so is the conditional distribution $\pxy$ even though intractable because we cannot compute the data distribution:

\begin{equation}
    \p(\vec{x},y) = \pdata \pyx.
\end{equation}

This questions how Eq. (\ref{eq:base_mi}) can be computed. Fortunately, well-known properties of MI can invert the distributions on which the KL divergence is computed~\cite{bridle_unsupervised_1992,krause_discriminative_2010} via Bayes' theorem:
\begin{equation} \label{eq:tractable_discriminative_mi}
    \I(\vec{x};y) = \E_{\vec{x} \sim \pdata} \left[ D_\text{KL} (\pyx \| \py)\right],
\end{equation}
which is possible to estimate. Since we highlighted earlier that the KL divergence in the MI can lead to inappropriate decision boundaries, we are interested in replacing it by other distances or divergences. However, changing it in Eq. (\ref{eq:tractable_discriminative_mi}) would focus on the separation of cluster assignments from the cluster proportions which may be irrelevant to the data distribution. We rather alter Eq. (\ref{eq:base_mi}) to clearly show that we separate data distributions given clusters from the entire data distribution because it allows us to take into account the data space geometry.

\subsection{The GEMINI}
\label{ssec:gemini}

\subsubsection{Replacing the Kullback-Leibler divergence with other distances}

The goal of the GEMINI is to separate data distributions according to an arbitrary distance $D$, i.e. changing the KL divergence for another divergence or distance in the MI. This brings the definition of our first GEMINI, the \emph{One-vs-All} (OvA):
\begin{equation}\label{eq:gemini_ova}
    \I^\text{OvA}_D(\vec{x};y) = \E_{y \sim \py} \left[ D(\pxy\|\pdata)\right],
\end{equation}
as it compares the distance between the distribution of a specific cluster $\pxy$ against the entire data distribution $p(x)$.  There exist other distances than the KL to measure how far two distributions $p$ and $q$ are one from the other. We can make a clear distinction between two types of distances, Csiszar's $f$-divergences~\cite{csiszar_information-type_1967} and Integral Probability Metrics (IPM)~\cite{sriperumbudur_integral_2009}. Unlike $f$-divergences, IPM-derived distances like the Wasserstein distance or the Maximum Mean Discrepancy (MMD)~\cite{gneiting_strictly_2007,gretton_kernel_2012} bring knowledge about the data throughout either a distance $c$ or a kernel $\kappa$: these distances are geometry-aware. 

\subsubsection{\texorpdfstring{$f$}{f}-divergence GEMINIs} These divergences involve a convex function $f:\mathbb{R}^+\rightarrow\mathbb{R}$ such that $f(1)=0$. This function is applied to evaluate the ratio between two distributions $p$ and $q$, as in Eq.~(\ref{eq:f_divergences_definition}):
\begin{equation}
\label{eq:f_divergences_definition}
D_\text{f-div}(p,q) = \E_{\vec{z} \sim q(\vec{z})} \left[ f\left(\frac{p(\vec{z})}{q(\vec{z})}\right)\right].
\end{equation}
We will focus on three $f$-divergences: the KL divergence, the Total Variation (TV) distance and the squared Hellinger distance. While the KL divergence is the usual divergence for the MI, the TV and the squared Hellinger distance present different advantages among $f$-divergences. First of all, both of them are bounded between 0 and 1. It is consequently easy to check when any GEMINI using those is maximised contrarily to the MI that is bounded by the minimum of the entropies of $\vec{x}$ and $y$~\cite{gray_maximum_1977}. When used as distance between data conditional distribution $\pxy$ and data distribution $\pdata$, we can apply Bayes' theorem in order to get an estimable equation to maximise (see App.~\ref{app:deriving_geminis}), which only involves cluster assignment $\pyx$ and marginals $\py$ as summarised in Table~\ref{tab:all_geminis}, generalising thus the work of Bridle et al.~\cite{bridle_unsupervised_1992}. Note that all $f$-divergences are maximised when the two distributions $p$ and $q$ have disjoint supports~\cite{liese_divergences_2011}. Common $f$-divergence like the KL, the squared Hellinger or the Pearson $\chi^2$ divergence, except the total variation distance, are specific cases of the $\alpha$-divergence subclass. The convex function of $\alpha$-divergence is parameterized by a real number $\alpha$ with:

\begin{equation}\label{eq:alpha_divergence_function}
    f_\alpha(t) = \left\{\begin{array}{cc}
        \frac{t^\alpha-\alpha t+(\alpha-1)}{\alpha(\alpha-1)}, &  \alpha\neq0, \alpha\neq1,\\
        t\ln{t}, &\alpha =1,\\
        -\ln{t},&\alpha=0.
    \end{array}\right.
\end{equation}

However, this class of $\alpha$-divergence is inappropriate in some cases for clustering. Indeed, we show with Proposition~\ref{prop:alpha_div_maximisation} (proof in App.~\ref{app:proof_alpha_div_maximisation}) that the maximisation of $\alpha$-divergences can lead to any clustering of the data space with balanced clusters as the discriminative model $\pyx$ converges to a Dirac distribution.

\begin{proposition}\label{prop:alpha_div_maximisation}
Let $\{\mathcal{X}_k\}_{k=1}^K$ a partition of $\mathcal{X}$ such that $\mathbb{P}(\vec{x} \in \mathcal{X}_k) = \frac{1}{K}$. Then for any $\alpha$-divergence with $\alpha>0$, the OvA GEMINI is upper bounded by a function which only depends on the proportions of the clusters. If the clustering model follows a Dirac distribution: $\p(y=k|\pmb{x})=\mathbf{1}_{[\vec{x}\in\mathcal{X}_k]}$, then the upper bound is tight and the GEMINI cannot be improved.
\end{proposition}

It is worth mentioning in Proposition~\ref{prop:alpha_div_maximisation} that the proportions of the cluster $p(y=k)$ do not matter for the specific case of $\alpha=2$ to achieve the global maximum, i.e. for the Pearson $\chi^2$-divergence. We can infer from Proposition~\ref{prop:alpha_div_maximisation} the specific Corollary~\ref{cor:mi_maximisation} since the MI is a case of OvA $\alpha$-divergence-GEMINI with $\alpha=1$. We conclude that MI maximisation is a poor objective when a discriminative model can converge to a Dirac distribution.

\begin{corollary}\label{cor:mi_maximisation}
Let $\{\mathcal{X}_k\}_{k=1}^K$ a partition of $\mathcal{X}$. Then the mutual information of a discriminative distribution $p(y|\vec{x})$ is upper bounded by the entropy of $\vec{x}$ and the upper bound is tight if the distribution is a Dirac model $p(y=k|\vec{x})=\pmb{1}_{[\vec{x}\in\mathcal{X}_k]}$. The highest upper bound is reached when the partition is balanced.
\end{corollary}

\subsubsection{IPM GEMINIs} The IPM is another class of distance that incorporates knowledge from the data through a function $f$:

\begin{equation}\label{eq:ipm}
    D_\text{ipm}(p,q) = \sup_{f\in\mathcal{F}} \E_{\vec{z} \sim p(\vec{z})}[f(\vec{z})] -\E_{\vec{z}\sim q(\vec{z})}[f(\vec{z})],
\end{equation}
where $\mathcal{F}$ is a set of functions. As backpropagation through suprema could be intractable, we choose to focus on two specific variations of the IPM for the GEMINI: the MMD and the Wasserstein distance. Note however that not all Wasserstein distances are IPMs and while some of our propositions are formulated for IPMs, we consider as well the entire class of the Wasserstein distances.

The MMD corresponds to the distance between the respective expected embedding of samples from the distribution $p$ and the distribution $q$ in a reproducible kernel Hilbert space (RKHS) $\mathcal{H}$:
\begin{equation}\label{eq:mmd_definition}
    \text{MMD}(p\|q) = \| \E_{\vec{z} \sim p(\vec{z})} [\varphi(\vec{z})] - \E_{\vec{z}\sim q(\vec{z})} [\varphi(\vec{z})]\|_\mathcal{H},
\end{equation}
where $\varphi$ is the RKHS embedding. To compute this distance we can use the kernel trick~\cite{gretton_kernel_2012} by involving the kernel function $\kappa(\vec{a},\vec{b})=\langle\varphi(\vec{a}),\varphi(\vec{b})\rangle$. We use Bayes' theorem to uncover a version of the MMD that can be estimated through Monte Carlo using only the predictions $\pyx$.

The Wasserstein distance is an optimal transport distance. It corresponds to the minimal amount of energy to transform a distribution into another according to an energy function yielding the cost $c$ of moving the mass of a sample from one location to another:
\begin{equation}\label{eq:wasserstein_definition}
    \mathcal{W}_c^d(p,q) = \left(\inf_{\gamma \in \Gamma(p,q)} \E_{\vec{x}, \vec{z} \sim \gamma(\vec{x},\vec{z})}\left[c(\vec{x},\vec{z})^d \right]\right)^{\frac{1}{d}},
\end{equation}
where $\Gamma(p,q)$ is the set of all couplings between $p$ and $q$, $c$ a distance function in $\mathcal{X}$ and $d$ a real positive number. Computing the Wasserstein-$d$ distance between two distributions $\p(\vec{x}|y=k)$ and $\p(\vec{x})$ is difficult in our discriminative context because we only have access to a finite set of samples $N$. Note that in the remainder of the paper, we will focus on the Wasserstein-1 metric. The idea of an expected Wasserstein distance was first proposed by Harchaoui~\cite[Eq. 48]{harchaoui_2020_learning} under the one-vs-rest name with an additional cluster proportion factor. However, we found this additional factor to be not grounded enough. Moreover, we can show that for the Wasserstein-1 metric the one-vs-rest Wasserstein preliminary work~\cite{harchaoui_2020_learning} is equivalent to the one-vs-all Wasserstein-GEMINI. To achieve the Wasserstein-GEMINI, we instead use approximations of the distributions with weighted sums of Diracs:

\begin{multline}\label{eq:dirac_approximation}
\p(\vec{x}|y=k) \approx \sum_{i=1}^N m_i^k \delta_{\vec{x}_i} = p_N^k,\\\text{with}\quad m_i^k = \frac{\p(y=k|\vec{x}_i)}{\sum_{j=1}^N\p(y=k|\vec{x}_j)},
\end{multline}
where $\delta_{\vec{x}_i}$ is a Dirac located on sample location $\vec{x}_i\in\mathcal{X}$. For the distribution $\pdata$, we set all importance weights to $1/N$. We state in Prop.~\ref{prop:wasserstein_convergence} that this empirical estimate of the Wasserstein distance converges to the correct Wasserstein distance. These importance weights are compatible with the \verb+emd2+ function of the python optimal transport package~\cite{flamary_pot_2021} which gracefully supports automatic differentiation. We describe other maximisation strategies in Appendix~\ref{app:other_wasserstein_distances}.

\begin{proposition}\label{prop:wasserstein_convergence}
    Let $p(\vec{x}|y=k_1)$ and $p(\vec{x}|y=k_2)$ two cluster distributions that we empirically approximate with importance-weighed Dirac estimators $p_N^{k_1}= \sum_{i=1}^N m_i^{k_1}\delta_{\vec{x}_i}$, resp. and  $p_N^{k_2}= \sum_{i=1}^N m_i^{k_2}\delta_{\vec{x}_i}$. Then the Wasserstein distance between estimates converges to the Wasserstein distance between the cluster distributions.
\end{proposition}

We refer to Appendix~\ref{app:wasserstein_convergence} for proof of convergence.

\subsection{The One-vs-One GEMINI}

We question the relevance of evaluating a distance between the distribution of the data given a cluster assumption $\pxy$ and the entire data distribution $\pdata$ when the geometry is taken into account. We argue that it is intuitive in clustering to compare the distribution of one cluster against the distribution of \emph{another cluster} rather than the data distribution. Indeed, considering the geometry of the data space through a kernel in the case of the MMD or a distance in the case of the Wasserstein metric implies that we can effectively measure how two distributions are close to one another. In the formal design of the mutual information, the distribution of each cluster $p(\vec{x}|y)$ is compared to the complete data distribution $p(\vec{x})$. Therefore, if one distribution of a specific cluster $p(\vec{x}|y)$ were to look alike the data distribution $p(\vec{x})$, for example up to a constant in some areas of the space, then its distance to the data distribution could be 0, making it unnoticed when maximising the OvA GEMINI.

\begin{figure}
\centering
    \subfloat[OvA]{
        \includegraphics[width=0.45\linewidth]{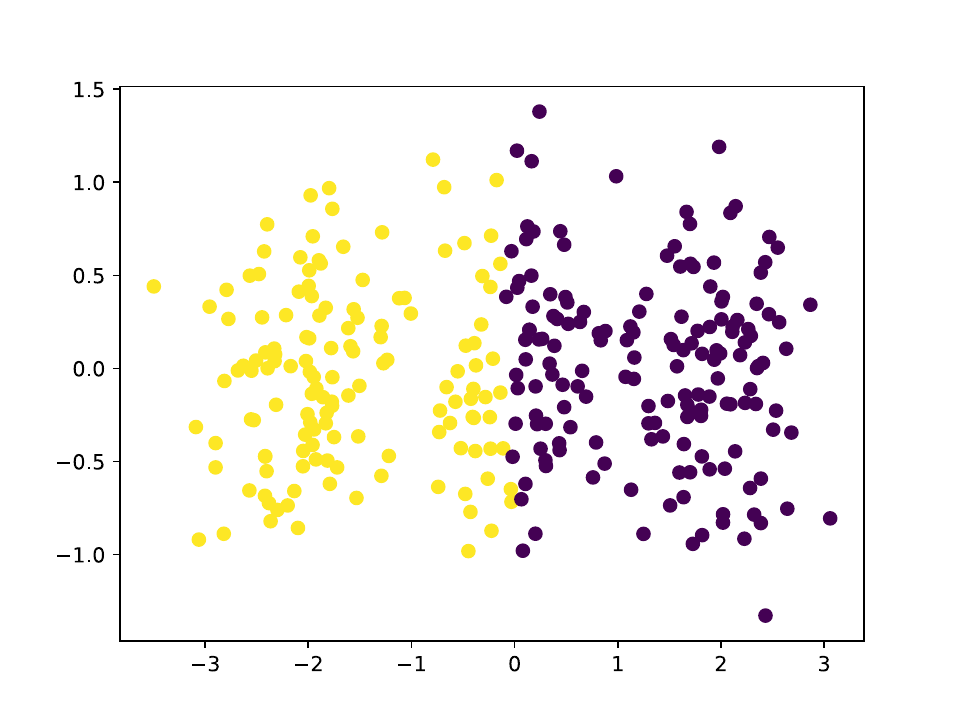}
        \label{sfig:example_ova}
    }\hfil
    \subfloat[OvO]{
        \includegraphics[width=0.45\linewidth]{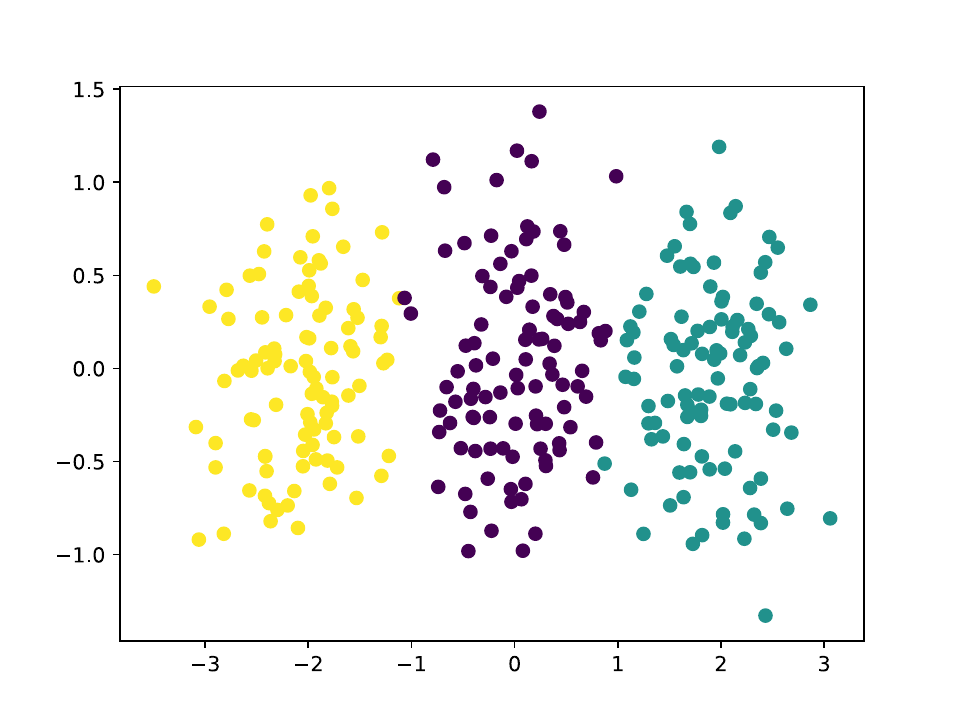}
        \label{sfig:example_ovo}
    }
\caption{Here, 3 clusters of equal proportions from isotropic Gaussian distributions are located in -2, 0 and 2 on the x-axis, with small covariance. The complete data distribution hence has its expectation in 0 on the x-axis. Consequently, maximising the OvA MMD-GEMINI with a logistic regression led to 2 clusters whereas the same model with the OvO MMD-GEMINI is able to see all 3 clusters.}\label{fig:example_ova_ovo}

\end{figure}

Take the example of 3 distributions $\{p(\vec{x}|y=i)\}_{i=1}^3$ with respective different expectations $\{\mu_i\}_{i=1}^3$. We want to separate them using the OvA MMD-GEMINI with linear kernel. The mixture of the 3 distributions creates a data distribution with expectation $\mu=\sum_{i=1}^3 p(y=i)\mu_i$. However, if the distributions satisfy that this data expectation $\mu$ is equal to one of the sub-expectations $\mu_i$, then the associated distribution $i$ will have will not provide any information since its MMD to the data distribution is equal to 0. We illustrate this example in figure~\ref{fig:example_ova_ovo}.

To address this issue, we introduce the second GEMINI named \emph{one-vs-one} (OvO) in which we compare cluster distributions from independently drawn cluster assignments $\ya$ and $\yb$:
\begin{equation}\label{eq:gemini_ovo}
    \I^\text{OvO}_D(\vec{x};y) = \E_{\ya,\yb \sim \py} \left[ D(\pxya \| \pxyb)\right].
\end{equation}

The example of Figure~\ref{fig:example_ova_ovo} is tackled by the OvO GEMINI since the distance between each pair of the 3 clusters is non-null. Conceptually, the idea of optimising the OvO MMD-GEMINI in clustering can be found as well by França et al.~\cite{franca_kernel_2020} who derived a regularised squared MMD in a one-vs-one setting through restrictions to Dirac distributions. Note that for most distances, the OvO GEMINI is an upper bound of the OvA GEMINI; proof of Proposition~\ref{prop:ovo_greater_ova} in App.~\ref{app:proof_ovo_greater_ova}.

\begin{proposition}\label{prop:ovo_greater_ova}
Let $D$ be an $f$-divergence or an IPM and $\p(y|\vec{x})$ a clustering distribution. Then: $\mathcal{I}^\text{ova}_D(\vec{x},y) \leq \mathcal{I}^\text{ovo}_D(\vec{x},y)$.
\end{proposition}

In the case of binary clustering, using an IPM distance implies equality between the OvA GEMINI and the OvO GEMINI; proof of Proposition~\ref{prop:equality_ova_ovo_ipm} in App.~\ref{app:proof_equality_ova_ovo_ipm}:

\begin{proposition}\label{prop:equality_ova_ovo_ipm}
Let $D$ be an IPM and $p(y|\vec{x})$ a clustering distribution $y$ taking $K=2$ values. Then: $\mathcal{I}^\text{ova}_D(\vec{x},y) = \mathcal{I}^\text{ovo}_D(\vec{x},y)$.
\end{proposition}

While the OvA GEMINI is maximised with Dirac clustering of the data space for some $\alpha$-divergence, we can extend Proposition~\ref{prop:alpha_div_maximisation} to all $f$-divergences for the OvO GEMINI with Proposition~\ref{prop:ovo_fdiv_maximisation} (Proof in App.~\ref{app:proof_ovo_fdiv_maximisation}). We notably conclude that for the total variation and the squared Hellinger distance, Dirac distributions on an even partition of the data space are the only optimal solutions.

\begin{proposition}\label{prop:ovo_fdiv_maximisation}
Let $D$ be an $f$-divergence, $\p(y|\vec{x})$ a clustering distribution such that $\p(y=k)=\frac{1}{K}$. The OvO GEMINI is then upper bounded by a function depending only on the cluster proportions. For the upper bound to be tight, a sufficient condition is to have disjoint supports between cluster distributions $\p(\vec{x}|y=k)$. The condition is necessary if the function $f$ satisfies $f(0) + g(0)< \infty$ where $g(t)=tf\left(\frac{1}{t}\right)$ is the convex conjugate of $f$.
\end{proposition}

\subsection{Using GEMINIs}

\begin{table*}[!tb]
\centering
\caption{Definition of the GEMINI for $f$-divergences, MMD and the Wasserstein distance. We directly write here the equation that can be optimised to train a discriminative model $\pyx$ via stochastic gradient descent since they are expectations over the data.}
\label{tab:all_geminis}
\begin{tabular}{c c}
\toprule
Name&Equation\\\hline\\
KL OvA/MI& $\E_{\pdata}\left[D_\text{KL}(\pyx \|\py)\right]$\\
KL OvO& $\E_{\pdata}[D_\text{KL}(\pyx \| \py))+D_\text{KL}(\py \| \pyx))]$\\
\begin{minipage}{0.2\linewidth}\centering Squared Hellinger\\OvA\end{minipage} & $1-\E_{\pdata}[\E_{\py}[\sqrt{\frac{\pyx}{\py}}]]$\\
\begin{minipage}{0.2\linewidth}\centering Squared Hellinger\\OvO \end{minipage} & $\E_{\pdata}[\mathbb{V}_{\py}[\sqrt{\frac{\pyx}{\py}}]]$\\
TV OvA& $\E_{\pdata} [D_\text{TV} (\pyx \| \py) ]$\\
TV OvO& $\frac{1}{2}\E_{\pdata}[\E_{\ya,\yb\sim\py}[|\frac{ \pyxa }{ \pya } - \frac{\pyxb}{\pyb}|]]$\\
\midrule\\
MMD OvA& $\E_{\py} \left[ \E_{\xa,\xb \sim \pdata} \left[ k(\xa, \xb) \left( \frac{\p(y|\xa)\p(y|\xb)}{\py^2} + 1 - 2\frac{\p(y|\xa)}{\py}\right) \right]^{\frac{1}{2}}\right]$ \\
MMD OvO& \begin{minipage}{0.7\linewidth}\centering\begin{multline*}\E_{\ya,\yb \sim \py} \left[ \E_{\xa,\xb \sim \pdata} \left[ k(\xa, \xb) \left( \frac{\p(\ya|\xb) \p(\ya|\xb)}{\pya^2} \right.\right.\right.\\\left.\left.\left. + \frac{\p(\yb |\xa)\p(\yb|\xb)}{\pyb^2} - 2\frac{\p(\ya |\xa)\p(\yb|\xb)}{\pya\pyb}\right) \right]^{\frac{1}{2}}\right] \end{multline*}\end{minipage}\\
Wasserstein OvA&$\mathbb{E}_{\py}\left[\mathcal{W}_c\left(\sum_{i=1}^N m_i^y\delta_{\vec{x}_i},\sum_{i=1}^N \frac{1}{N}\delta_{\vec{x}_i}\right)\right]$\\
Wasserstein OvO&$\mathbb{E}_{\ya,\yb\sim \py}\left[\mathcal{W}_c\left(\sum_{i=1}^N m_i^{\ya}\delta_{\vec{x}_i},\sum_{i=1}^N m_i^{\yb}\delta_{\vec{x}_i}\right)\right]$\\
\bottomrule
\end{tabular}
\end{table*}

\subsubsection{Choosing a GEMINI}

We stated in Section~\ref{ssec:discriminative_clustering} that we present the GEMINI as a distance between distributions evaluated in the data space $\mathcal{X}$ so that the distance $D$ can take into account the topology of the data. In practice, we only design a discriminative model $\pyx$. Thus, we need to compute all formulas of the GEMINI through Bayes' theorem to get equations depending on $\pyx$ and $\py$. We summarise the equations from all aforementioned GEMINIs in Table~\ref{tab:all_geminis} (see Appendix~\ref{app:deriving_geminis} for derivations). We give details on the complexity of GEMINIs in Appendix~\ref{app:exp_complexity} to help choose one. We propose as well some theoretical speed-ups in App.~\ref{app:wasserstein_speedup}. It is also important to consider the experimental purposes and context to choose a GEMINI. Indeed, when it is easier to design a distance than a kernel, the Wasserstein-GEMINI is more compatible than the MMD-GEMINI and vice-versa. Moreover, the MMD-GEMINI inherently computes expectations in a Hilbert Space which allows computing centroids deemed representative of the clusters. This notion of centroid is less straightforward when using the Wasserstein metric. 

\begin{figure*}[!t]
    \centering
    \subfloat[OvA GEMINIs]{
        \includegraphics[width=0.45\linewidth]{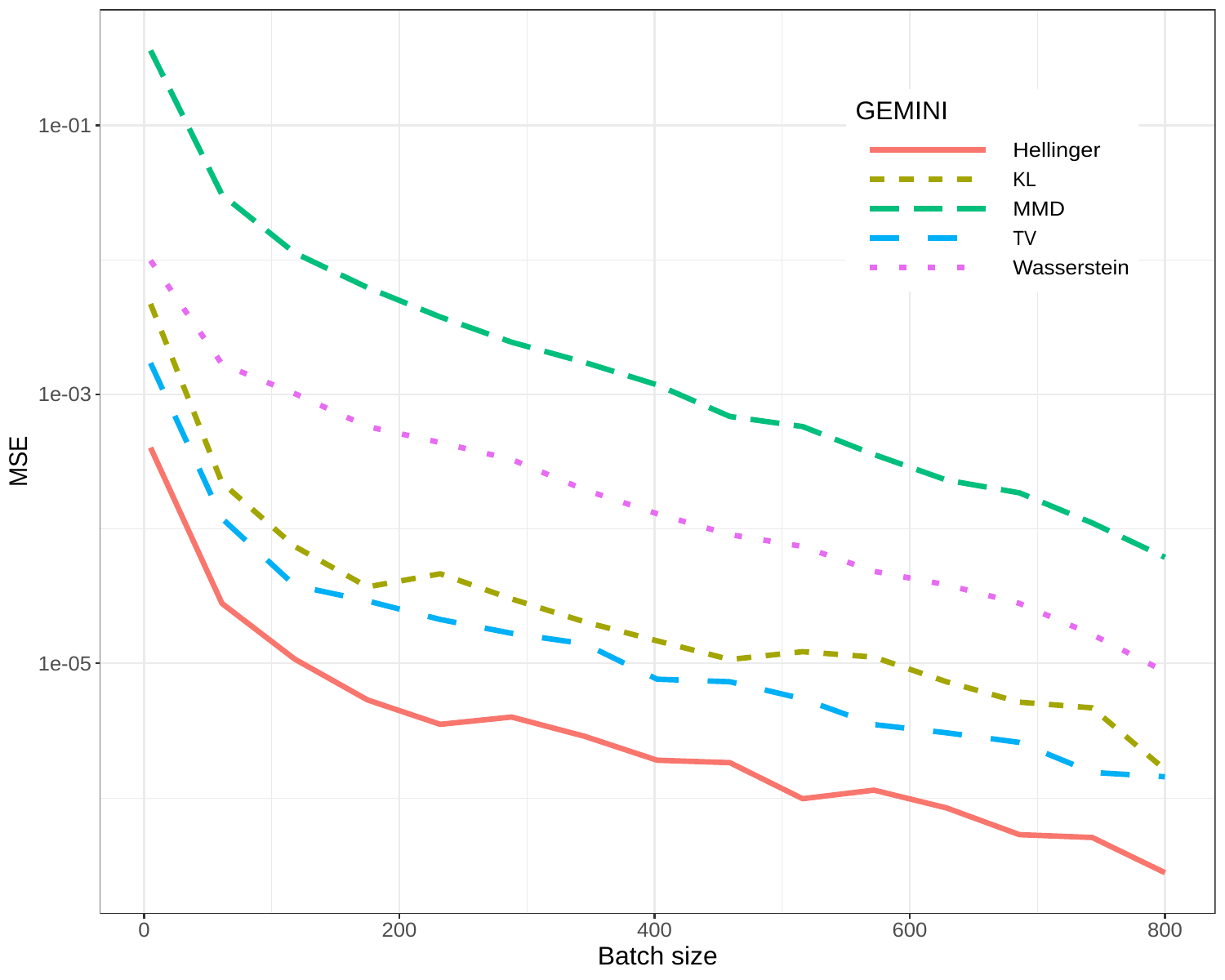}
        \label{sfig:mse_ova}
    }\hfil
    \subfloat[OvO GEMINIs]{
        \includegraphics[width=0.45\linewidth]{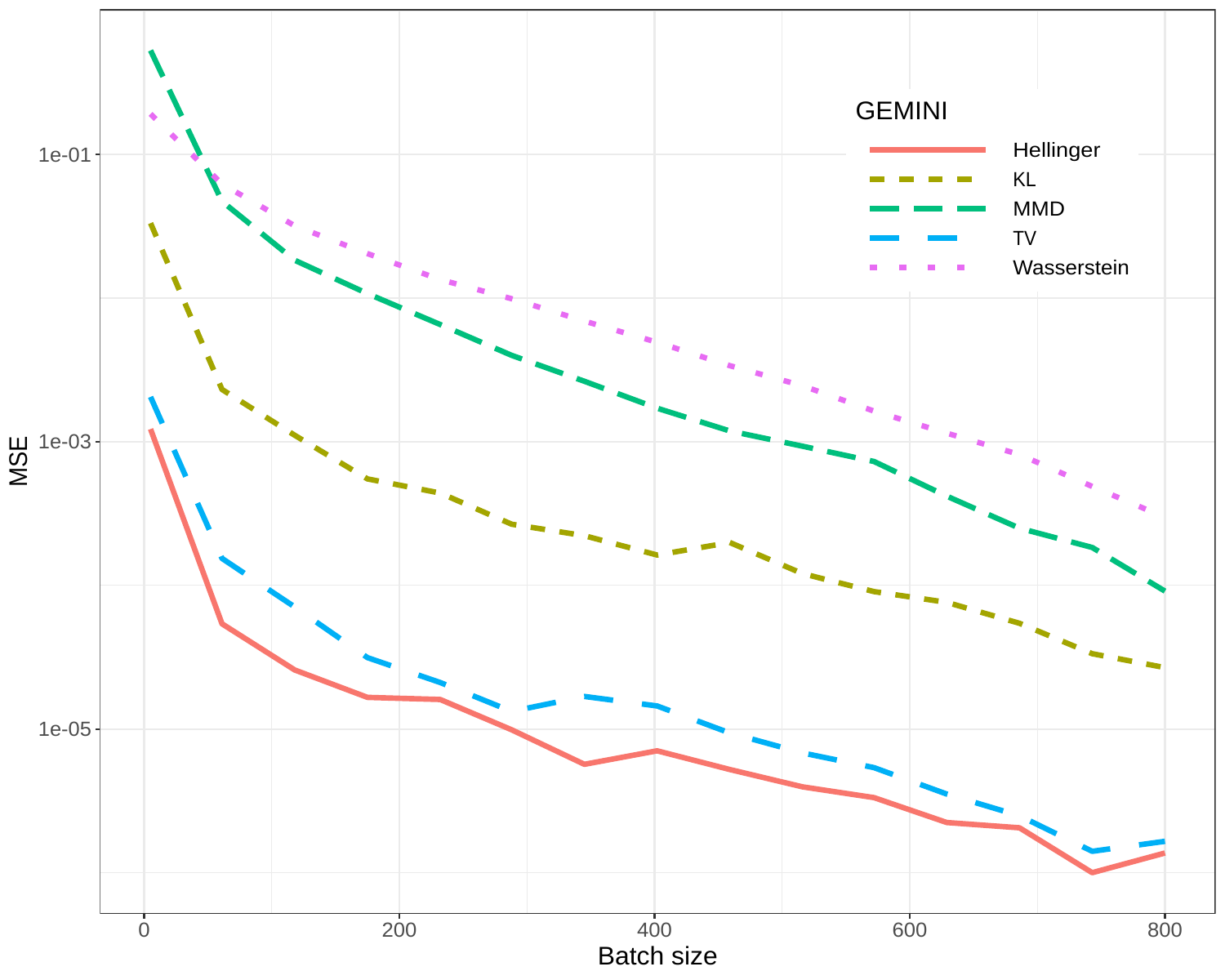}
        \label{sfig:mse_ovo}
    }
    \caption{Mean Squared Error (log scale) of estimates with varying batch sizes compared to the true value over a complete dataset of a 1000 samples. Each estimate was performed 50 times per batch size and GEMINI.}
    \label{fig:mse_gemini}
\end{figure*}

\subsubsection{Estimating a GEMINI}

All GEMINIs in Table~\ref{tab:all_geminis} can be estimated using Monte Carlo making them compatible with mini-batch learning, with batch sizes of a few hundred for large datasets similarly to prior works~\cite{hu_learning_2017,ji_invariant_2019, hjelm_learning_2019}. We highlight the importance of the batch size when using GEMINIs. With the use of mini-batch for training, the complete GEMINI is not evaluated on the entire dataset and hence a bias may rise from the empirical estimate. This bias then has consequences on the gradient, which in turn alters training. To illustrate this point, we generated 1000 predictions from a Dirichlet distribution with 10 clusters. These predictions are a proxy for the output of any discriminative model $\pyx$. We then compute the true GEMINI on all samples before evaluating it 50 times for different randomly sampled batches of increasing size. We report in Figure~\ref{fig:mse_gemini} the Mean Squared Error of all GEMINIs. We see that past 200 samples for both the OvA and the OvO models, the mean squared error is already close to or below $10^{-2}$, except for the OvO Wasserstein- and MMD-GEMINIs. This implies an upper bound of $10^{-2}$ for the bias of the estimates. We conclude that there is possibly a bias in GEMINIs estimates, but it remains small enough to be negligible.

\subsubsection{Code}

The original implementation for all experiments regarding GEMINI can be found here: \url{https://github.com/oshillou/GEMINI}. However, later works led to the development of python package for small-scale datasets on CPU: \emph{gemclus} at \url{https://gemini-clustering.github.io/}.

\section{Experiments}
\label{sec:experiments}
For all experiments below, we report the Adjusted Rand Index (ARI)~\cite{hubert_comparing_1985}, a common metric in clustering. This metric is external as it requires labels for evaluation. It ranges from 0, when labels are independent from cluster assignments, to 1, when labels are equivalent to cluster assignments up to permutations. An ARI close to 0 is equivalent to the best accuracy when voting constantly for the majority class, e.g. 10\% on a balanced 10-class dataset. Regarding the MMD- and Wasserstein-GEMINIs, we used by default a linear kernel and the Euclidean distance unless specified otherwise. All discriminative models are trained using the Adam optimiser~\cite{kingma_adam_2014}. We estimate a total of 450 hours of GPU consumption. (See Appendix~\ref{app:requirements} for the details of python packages for experiments).

\subsection{When the MI fails because of the modelling}
\label{ssec:exp_categorical}
\begin{figure}
    \centering
    \subfloat[OvA KL (MI)]{
        \fbox{ \includegraphics[height=0.1\paperheight,width=0.4\linewidth]{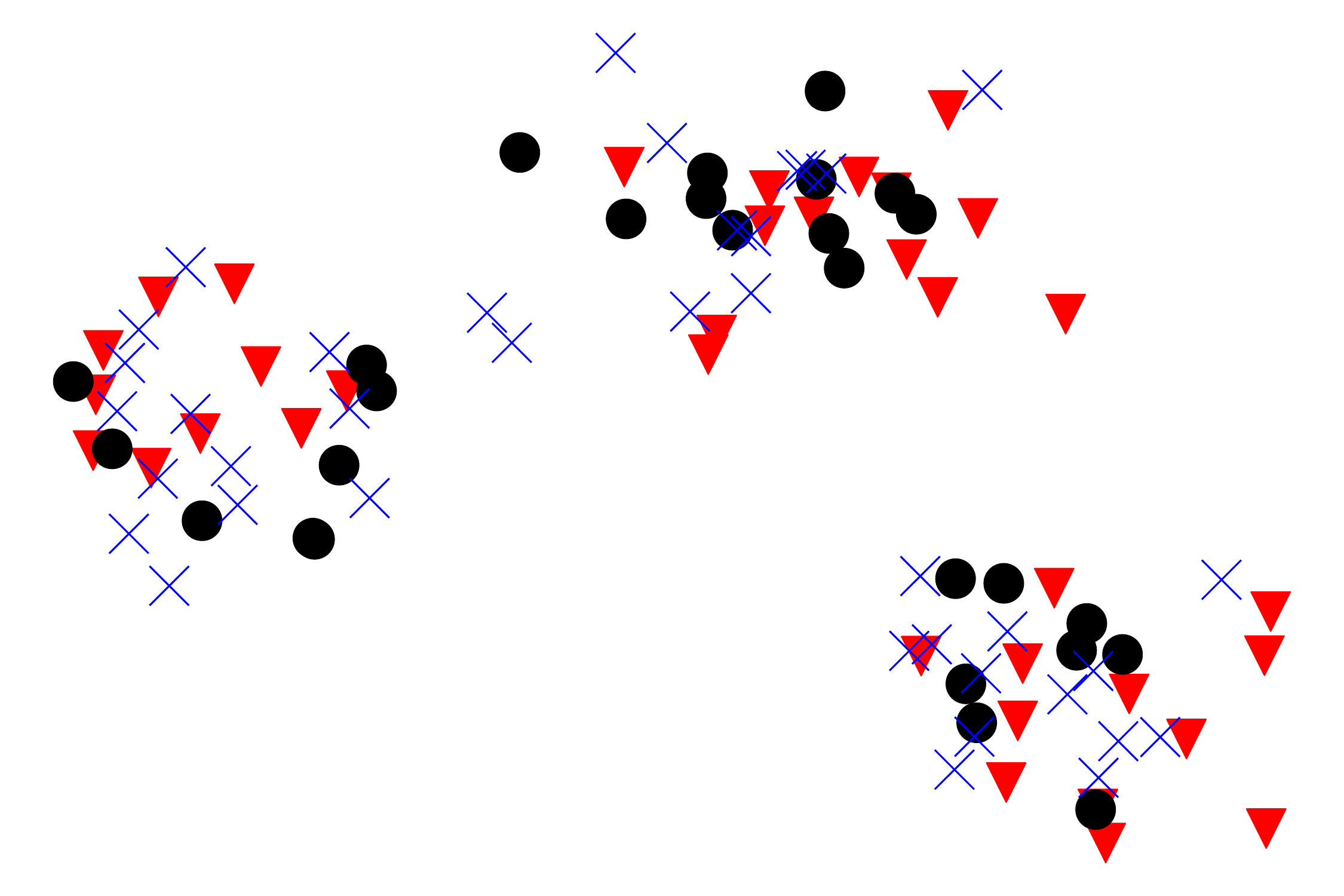}}
        \label{sfig:mi_categorical}
    }\hfil
    \subfloat[OvA MMD with linear kernel]{
        \fbox{ \includegraphics[height=0.1\paperheight,width=0.4\linewidth]{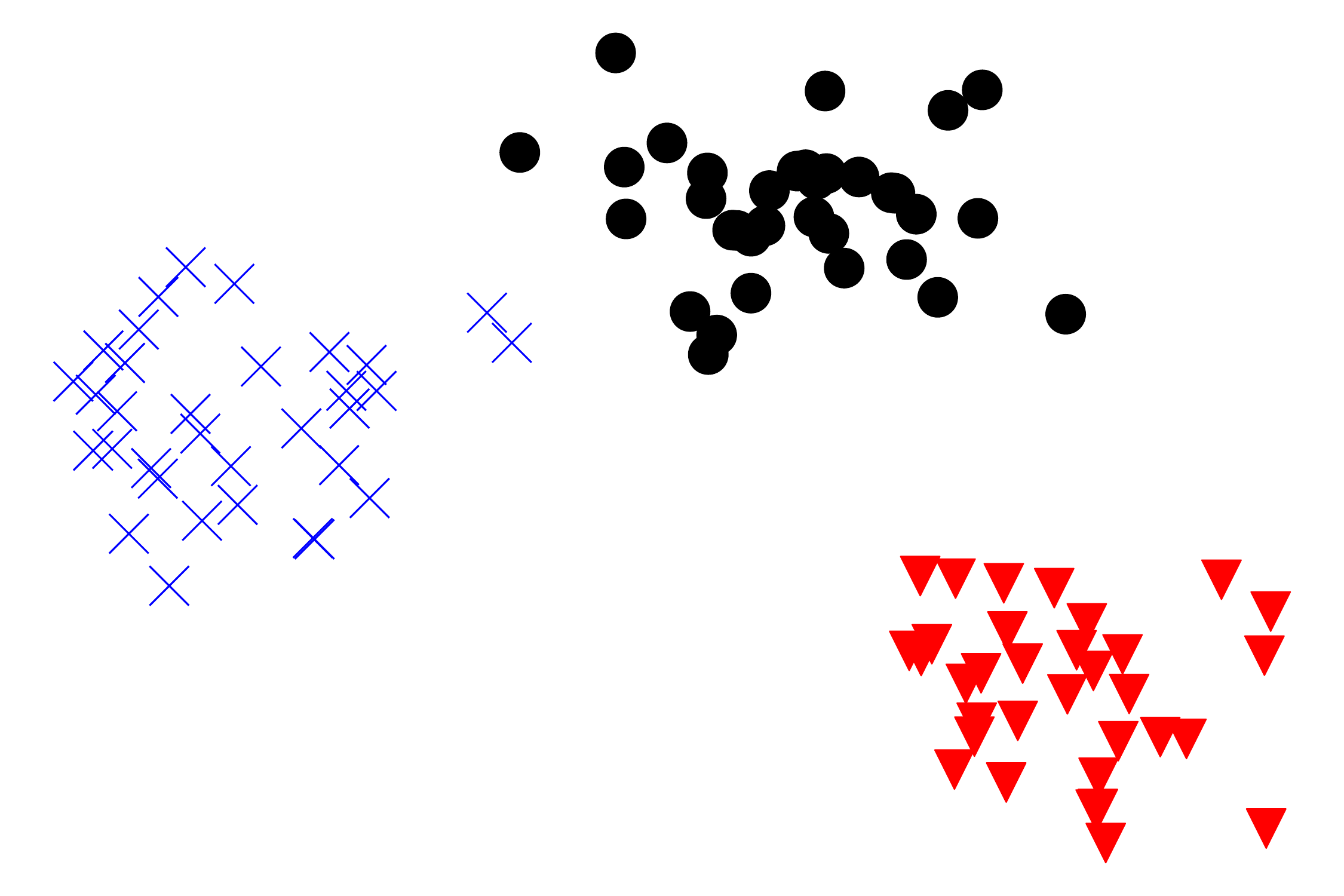}}
        \label{sfig:mmd_categorical}
    }
    \caption{Clustering of a mixture of 3 Gaussian distributions with MI (left) and a GEMINI (right) using categorical distributions. The MI does not have insights on the data shape because of the model, and clusters points uniformly between the 3 clusters (black dots, red triangles and blue crosses) whereas the MMD is aware of the data shape through its kernel.}
    \label{fig:categorical_boundaries}
\end{figure}

We first took the most simple discriminative clustering model, where each cluster assignment according to the input datum follows a categorical distribution:
\begin{equation*}
    y|\vec{x}=\vec{x}_i \sim \text{Cat}(\theta_i^1,\theta_i^2,\cdots,\theta_i^K).
\end{equation*}
We generated $N=100$ samples from a simple mixture of $K=3$ Gaussian distributions. Each model thus only consists in $NK$ parameters to optimise. This is a simplistic way of describing the most flexible deep neural network. We then maximised on the one hand the OvA KL (MI) and on the other hand the OvA MMD. Both clustering results can be seen in Figure~\ref{fig:categorical_boundaries}. We concluded that without any function, e.g. a neural network, to link the parameters of the conditional distribution with $\vec{x}$, the MI struggles to find the correct decision boundaries. Indeed, the position of $\vec{x}$ in the 2D space plays no role and the decision boundary becomes only relevant with regards to cluster entropy maximisation: a uniform distribution between 3 clusters. However, it plays a major role in the kernel of the MMD-GEMINI thus solving correctly the problem.

\subsection{Resistance to outliers}
\label{ssec:exp_gstm}
\begin{table*}
    \caption{Mean ARI (std) of a MLP fitting a mixture of 3 Gaussian and 1 Student-t multivariate distributions compared with Gaussian Mixture Models and K-Means. The model can be tasked to find either 4 or 8 clusters at best and the Student-t distribution has $\rho$=1 or 2 degrees of freedom. Bottom line presents the ARI for the maximum a posteriori of an oracle aware of all parameters of the data.}
    \label{tab:gstm_experiment_ari}
    \centering
    \begin{tabular}{c c c c c}
        \toprule
        \multirow{2}[3]{*}{Model}& \multicolumn{2}{c}{$\rho=2$}&\multicolumn{2}{c}{$\rho=1$}\\
        \cmidrule(lr){2-3}\cmidrule(lr){4-5}
        & 4 clusters&8 clusters&4 clusters&8 clusters\\
        \midrule
        K-Means&0.965 (0)&0.897 (0.040)&0 (0)&0.657 (0.008)\\
        GMM (full covariance)&0.972 (0)&0.868 (0.042)&0 (0)&0.610 (0.117)\\
        GMM (diagonal covariance)&{\bf 0.973 (0)}& 0.862 (0.048)&0.024 (0.107)& 0.660 (0.097)\\
        \midrule
        $\I_\text{KL}^\text{ova}$&0.883 (0.182)&0.761 (0.101)&{\bf 0.939 (0.006)}&0.742 (0.092)\\
        $\I_\text{KL}^\text{ovo}$&0.731 (0.140)&0.891 (0.129)&0.723 (0.114)&0.755 (0.163)\\
        $\I_{\text{H}^2}^\text{ova}$&0.923 (0.125)&{\bf 0.959 (0.043)}&0.906 (0.103)&0.86 (0.087)\\
        $\I_{\text{H}^2}^\text{ovo}$&0.926 (0.112)&0.951 (0.059)&0.858 (0.143)&0.887 (0.074)\\
        $\I_\text{TV}^\text{ova}$&0.940 (0.097)&0.973 (0.004)&0.904 (0.104)&{\bf 0.925 (0.103)}\\
        $\I_\text{TV}^\text{ovo}$&0.971 (0.005)&0.620 (0.053)&{\bf 0.938 (0.005)}&0.595 (0.055)\\
        \midrule
        $\I_\text{MMD}^\text{ova}$&0.953 (0.060)&0.940 (0.033)&0.922 (0.004)&{\bf 0.908 (0.016)}\\
        $\I_\text{MMD}^\text{ovo}$&0.968 (0.001)&0.771 (0.071)&0.921 (0.007)&0.849 (0.048)\\
        $\I_\mathcal{W}^\text{ova}$&0.897 (0.096)&0.896 (0.021)&0.915 (0.131)&0.889 (0.051)\\
        $\I_\mathcal{W}^\text{ovo}$&0.970 (0.002)&0.803 (0.067)&0.922 (0.006)&0.817 (0.042)\\
        \midrule
        Oracle&\multicolumn{2}{c}{0.991}&\multicolumn{2}{c}{0.989}\\
        \bottomrule
    \end{tabular}
\end{table*}
\begin{figure*}
    \centering
    \subfloat[OvA KL (MI) Entropy Map]{
        \includegraphics[width=0.3\linewidth]{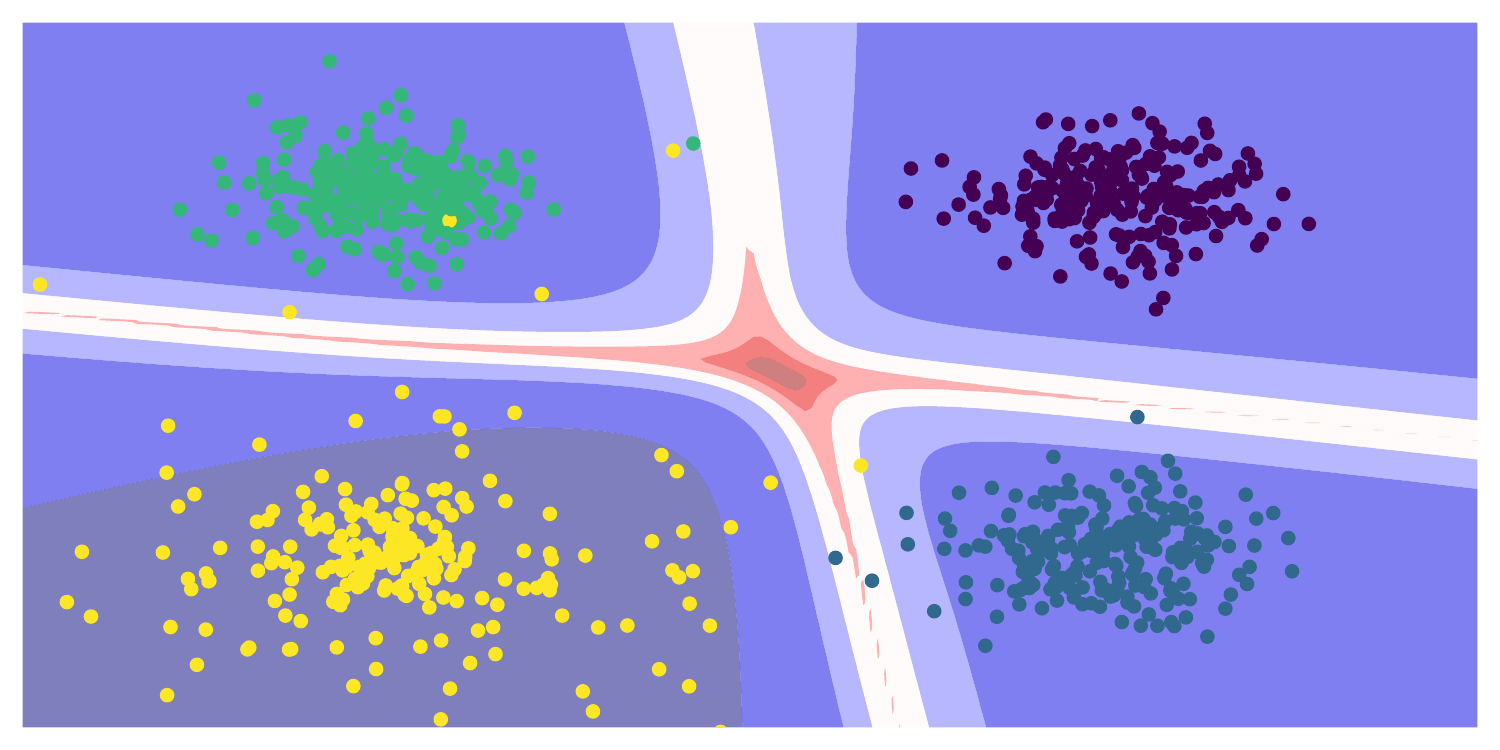}
        \label{sfig:mi_entropy_map}
    }\hfil
    \subfloat[OvO MMD Entropy Map]{
        \includegraphics[width=0.3\linewidth]{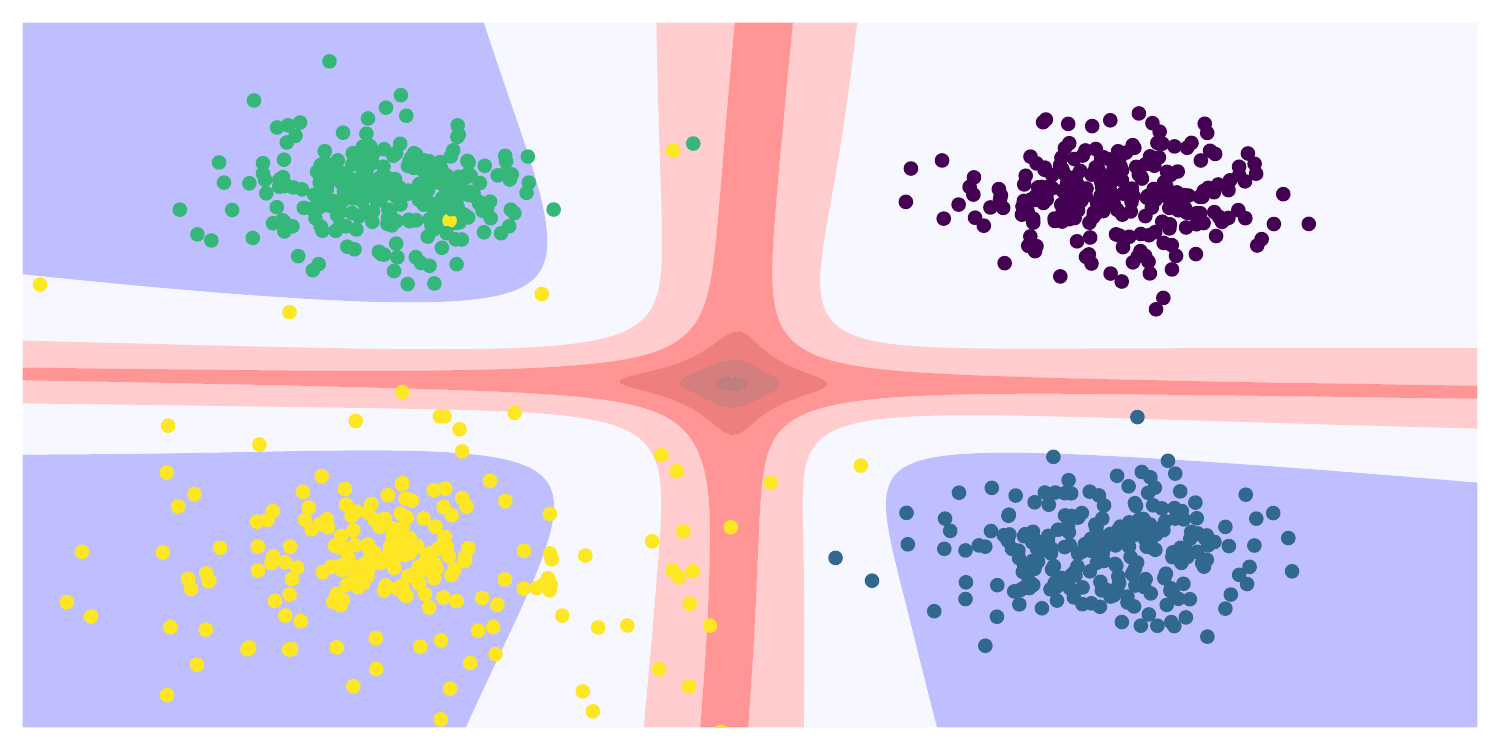}
        \label{sfig:mmd_entropy_map}
    }\hfil
    \subfloat[OvO Wasserstein Entropy Map]{
       \includegraphics[width=0.3\linewidth]{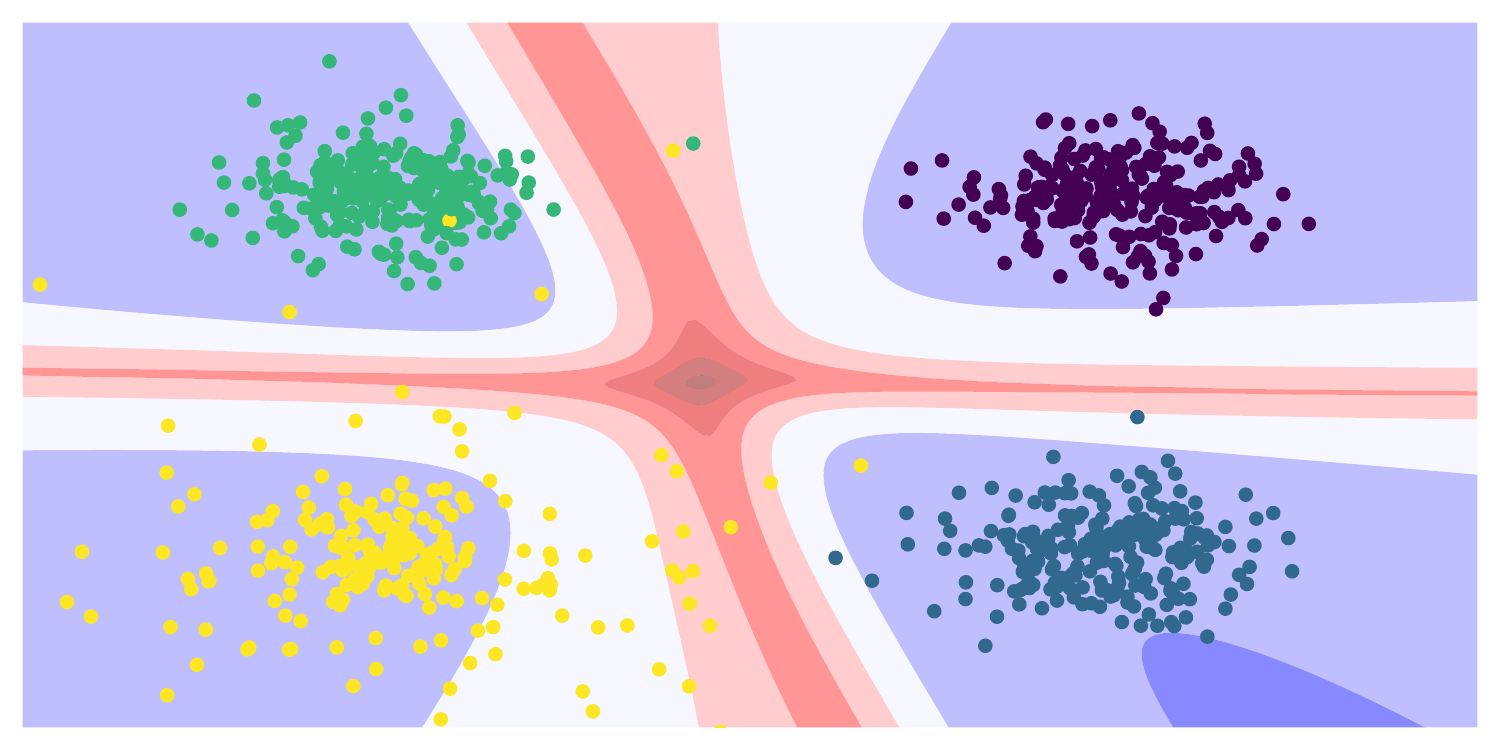}
       \label{sfig:wasserstein_entropy_map}
    }
    \caption{Entropy maps of the predictions of each MLP trained using a GEMINI or the MI. The bottom-left distribution (yellow) is a Student-t distribution with 1 degree of freedom that produces samples far from the origin. The Rényi entropy of prediction is highlighted from lowest (blue background) to highest (red background). MI on the left has the most confident predictions overall and the smallest uncertainty around the decision boundary, i.e. high entropy variations.}
    \label{fig:gstm_entropy_maps}
\end{figure*}


To prove the strength of using neural networks for clustering trained with GEMINI, we introduced extreme samples in Gaussian mixtures by replacing a Gaussian distribution with a Student-t distribution for which the degree of freedom $\rho$ is small. We fixed $K=4$ clusters, 3 being drawn from multivariate Gaussian distributions and the last one from a multivariate Student-t distribution in 2 dimensions for visualisation purposes with 1 degree of freedom or 2. Thus, the Student-t distribution produces samples that can be perceived as outliers regarding a Gaussian mixture owing to its heavy tail.

Each cluster distribution is centered around a mean $\mu_i$ which proximity is controlled by a scalar $\alpha$. For simplicity, all covariance matrices are the identity scaled by a scalar $\sigma$. We define:
\begin{align*}
    \mu_1 &= [\alpha, \alpha],&\mu_2 &= [\alpha, -\alpha],\\
\mu_3 &= [-\alpha, \alpha],&\mu_4 &= [-\alpha, -\alpha].
\end{align*}

To sample from a multivariate Student-t distribution, we first draw samples $\vec{x}$ from a centered multivariate Gaussian distribution. We then sample another variable $u$ from a $\chi^2$-distribution using the degrees of freedom $\rho$ as parameter. Finally, $\vec{x}$ is multiplied by $\sqrt{\frac{\rho}{u}}$, yielding samples from the Student-t distribution.

We report the ARIs of Multi-Layered Perceptron (MLP) trained 20 times with GEMINIs in Table~\ref{tab:gstm_experiment_ari}. The presence of "outliers" leads K-Means and Gaussian Mixture models to fail at grasping the 4 distributions when tring to find 4 clusters. Meanwhile, GEMINIs perform better. Note that the MMD and Wasserstein-GEMINI present lower standard deviation for high scores compared to $f$-divergence GEMINIs. We attribute these performances to both the MLP that tries to find separating hyperplanes in the data space and the absence of hypotheses regarding the data. Moreover, as mentioned in section~\ref{ssec:local_maxima}, the usual MI is best maximised when its decision boundary presents little entropy $\mathcal{H}(\pyx)$. As neural networks can be overconfident~\cite{guo_calibration_2017}, MI is likely to yield overconfident clustering by minimizing the conditional entropy. We highlight such behaviour in Figure~\ref{fig:gstm_entropy_maps} where the Rényi entropy~\cite{renyi_measures_1961} associated to each sample in the MI (Figure~\ref{sfig:mi_entropy_map}) is much lower, if not 0, compared to OvO MMD and OvO Wasserstein (figures~\ref{sfig:mmd_entropy_map} and~\ref{sfig:wasserstein_entropy_map}). We conclude that Wasserstein- and MMD-GEMINIs train neural networks not to be overconfident, hence yielding more moderate distributions $\pyx$.

\subsection{Leveraging a manifold geometry}
\label{ssec:exp_moons}
\begin{figure}
    \centering
    \subfloat[$\I^\text{ova}_\text{kl}$ for 2 clusters]{
        \includegraphics[width=0.45\linewidth]{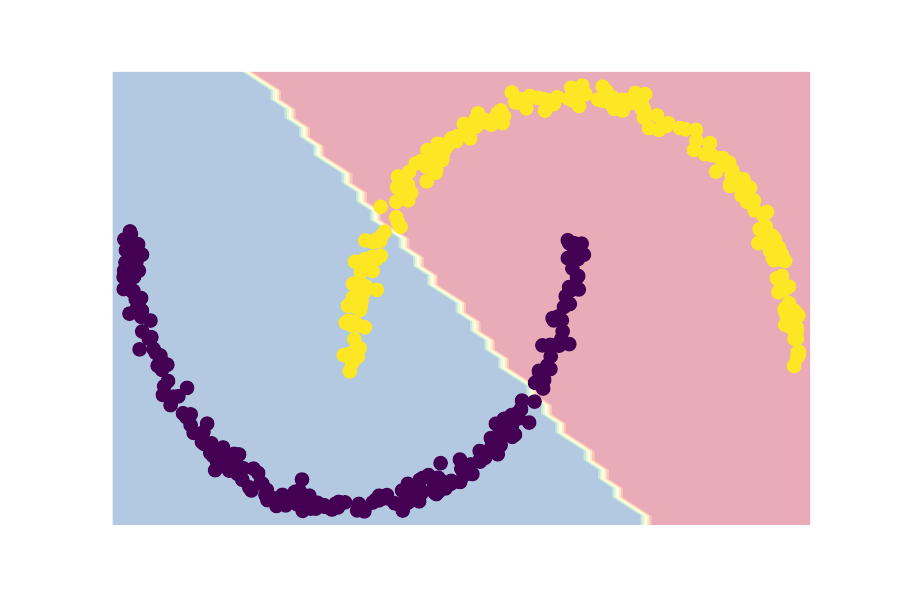}
        \label{sfig:moons_mi_2clusters}
    }\hfil
    \subfloat[$\I^\text{ova}_\text{kl}$ for 5 clusters]{
        \includegraphics[width=0.45\linewidth]{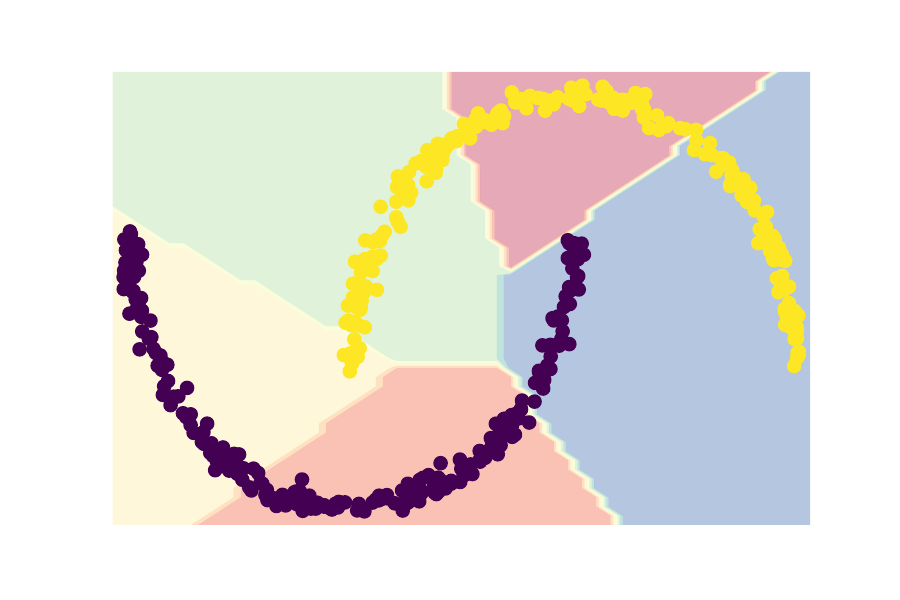}
        \label{sfig:moons_mi_5clusters}
    }\\
    \subfloat[$\I^\text{ovo}_\mathcal{W}$ for 2 clusters]{
        \includegraphics[width=0.45\linewidth]{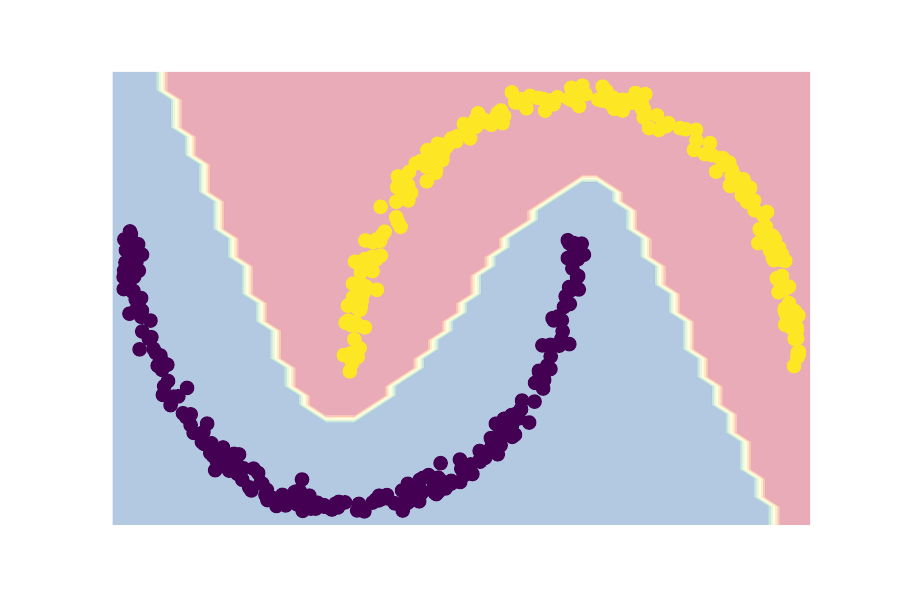}
        \label{sfig:moons_gemini_2clusters}
    }\hfil
    \subfloat[$\I^\text{ova}_\mathcal{W}$ for 5 clusters]{
        \includegraphics[width=0.45\linewidth]{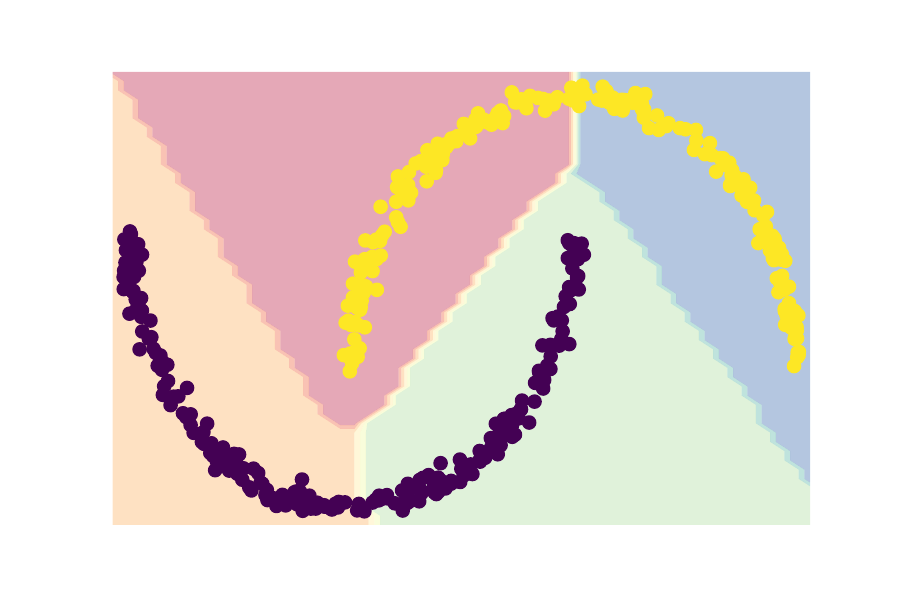}
        \label{sfig:moons_gemini_5clusters}
    }
    \caption{Fitting hand-generated moons using the GEMINI on top of an MLP for different amount of clusters. The Wasserstein-ovo model with 5 clusters eventually found 4 clusters.}
    \label{fig:moons_decision_boundary}
\end{figure}

We highlighted that MI can be maximised without requiring to find the suitable decision boundary. Here, we show how the provided distance to the OvO Wasserstein-GEMINI can leverage appropriate clustering when we have a good a priori on the data.

The cluster assignment is drawn according to a Bernoulli distribution with parameter 0.5. The points are then sampled using a uniformly distributed angle in $[0,\pi]$. Axis symmetry is applied depending on the clusters. All points are distributed at the same radius $\rho$ from some circle centre before adding some Gaussian noise. The samples then undergo some offset so that both moons are not linearly separable.

We generated a dataset consisting of two facing moons on which we trained a MLP using either the MI or the OvO Wasserstein-GEMINI. To construct a distance $c$ for the Wasserstein distance, we derived a distance from the Floyd-Warshall algorithm~\cite{warshall_theorem_1962,roy_transitivite_1959} on a sparse graph describing neighborhoods of samples. This distance describes how many neighbors are in between two samples, further details are provided in appendix~\ref{app:fw_distance}. We report the different decision boundaries in Figure~\ref{fig:moons_decision_boundary}. We observe that the insight on the neighborhood provided by our distance $c$ helped the MLP to converge to the correct solution with an appropriate decision boundary unlike the MI.  Note that the usual Euclidean distance in the Wasserstein metric would have converged to a solution similar to the MI. Indeed for 2 clusters, the optimal transport plan has a larger value, using a distribution similar to Figure~\ref{sfig:moons_mi_2clusters}, than in Figure~\ref{sfig:moons_gemini_2clusters}. This toy example shows how an insightful metric provided to the Wasserstein distance in GEMINIs can lead to correct decision boundaries while only designing a discriminative distribution $\pyx$ and a distance $c$.

In addition, we highlight an interesting behaviour of all GEMINIs. During optimisation, it is possible that the model converges to using fewer clusters than the number to find. For example in Figure~\ref{fig:moons_decision_boundary}, for 5 clusters, the model can converge to 4 balanced clusters and 1 empty cluster (Figure~\ref{sfig:moons_gemini_5clusters}) unlike MI that produced 5 misplaced clusters (Figure~\ref{sfig:moons_mi_5clusters}). Indeed, the entropy on the cluster proportion in the MI forces to use the maximum number of clusters.

\subsection{Fitting MNIST}
\label{ssec:exp_mnist}
\begin{table*}
    \caption{ARI for deep neural network trained with GEMINIs on MNIST for 500 epochs. Models were trained either with either 10 clusters to find or 15. We indicate in parentheses the number of used clusters by the model after training.}
    \label{tab:mnist_experiment}
    \centering
    \begin{tabular}{c c c c c c}
    \toprule
        \multicolumn{2}{c}{\multirow{2}[3]{*}{GEMINI}}& \multicolumn{2}{c}{10 clusters}&\multicolumn{2}{c}{15 clusters} \\
        \cmidrule(lr){3-4}\cmidrule(lr){5-6}
        \multicolumn{2}{c}{}&MLP&LeNet-5&MLP&LeNet-5\\
    \midrule
        \multirow{2}{*}{KL} & OvA&0.320 (10)&0.138 (8)&0.271 (15)&0.136 (12)\\
        &OvO&0.348 (7)&0.123 (4)&0.333 (8)&0.104 (4)\\
        
        \multirow{2}{*}{Squared Hellinger} & OvA&0.301 (10)&0.207 (6)&0.224 (13)&0.162 (7)\\
        &OvO&0.287 (10)&0.161 (6)&0.305 (13)&0.157 (7)\\
        
        \multirow{2}{*}{TV} & OvA&0.299 (10)&0.171 (6)&0.277 (15)& 0.140 (6)\\
        &OvO&0.422 (10)&0.161 (9)&0.330 (15)&0.182 (14)\\
        
    \midrule
        
        \multirow{2}{*}{MMD} & OvA&0.373 (10)&0.382 (10)&0.345 (15)&0.381 (15)\\
        &OvO&0.361 (10)&0.373 (10)&0.364 (15)&0.379 (15)\\
        
        \multirow{2}{*}{Wasserstein}& OvA&{\bf 0.471} (10)&{\bf 0.463} (10)&0.390 (15)&{\bf 0.446} (11)\\
        &OvO&0.450 (10)&0.383 (10)&{\bf 0.415} (15)&0.414 (15)\\
        \midrule
        \multicolumn{2}{c}{K-Means}&\multicolumn{2}{c}{0.367}&\multicolumn{2}{c}{0.385}\\
    \bottomrule
    \end{tabular}
\end{table*}

We trained a neural network using either MI or GEMINIs. Following Hu et al.~\cite{hu_learning_2017}, we first tried with a MLP with one single hidden layer of dimension 1200. To further illustrate the robustness of the method and its adaptability to other architectures, we also experimented using a LeNet-5 architecture~\cite{lecun_gradient_1998} since it is adequate to the MNIST dataset. We report our results in Table~\ref{tab:mnist_experiment}. Since we are dealing with a clustering method, we may not know the number of clusters a priori in a dataset. The only thing that can be said about MNIST is that there are \emph{at least} 10 clusters, one per digit. Indeed, writings of digits could differ leading to more clusters than the number of classes. That is why we further tested the same method with 15 clusters to find in Table~\ref{tab:mnist_experiment}. We first see that the scores of the MMD and Wasserstein GEMINIs are greater than the MI. We also observe that no $f$-divergence-GEMINI always yield best ARIs. Nonetheless, we observe better performances in the case of the TV GEMINIs owing to its bounded gradient. This results in controlled stepsize when doing gradient descent contrarily to KL- and squared Hellinger-GEMINIs. Notice that the change of architecture from a MLP to a LeNet-5 unexpectedly halves the scores for the $f$-divergences. We believe this drop is due to the change of notion of neighborhood implied by the network architecture.

\subsection{Number of non-empty clusters and architecture}
\label{ssec:cluster_selection}
\begin{figure}
\centering
    \subfloat[MLP]{
        \includegraphics[width=0.45\linewidth]{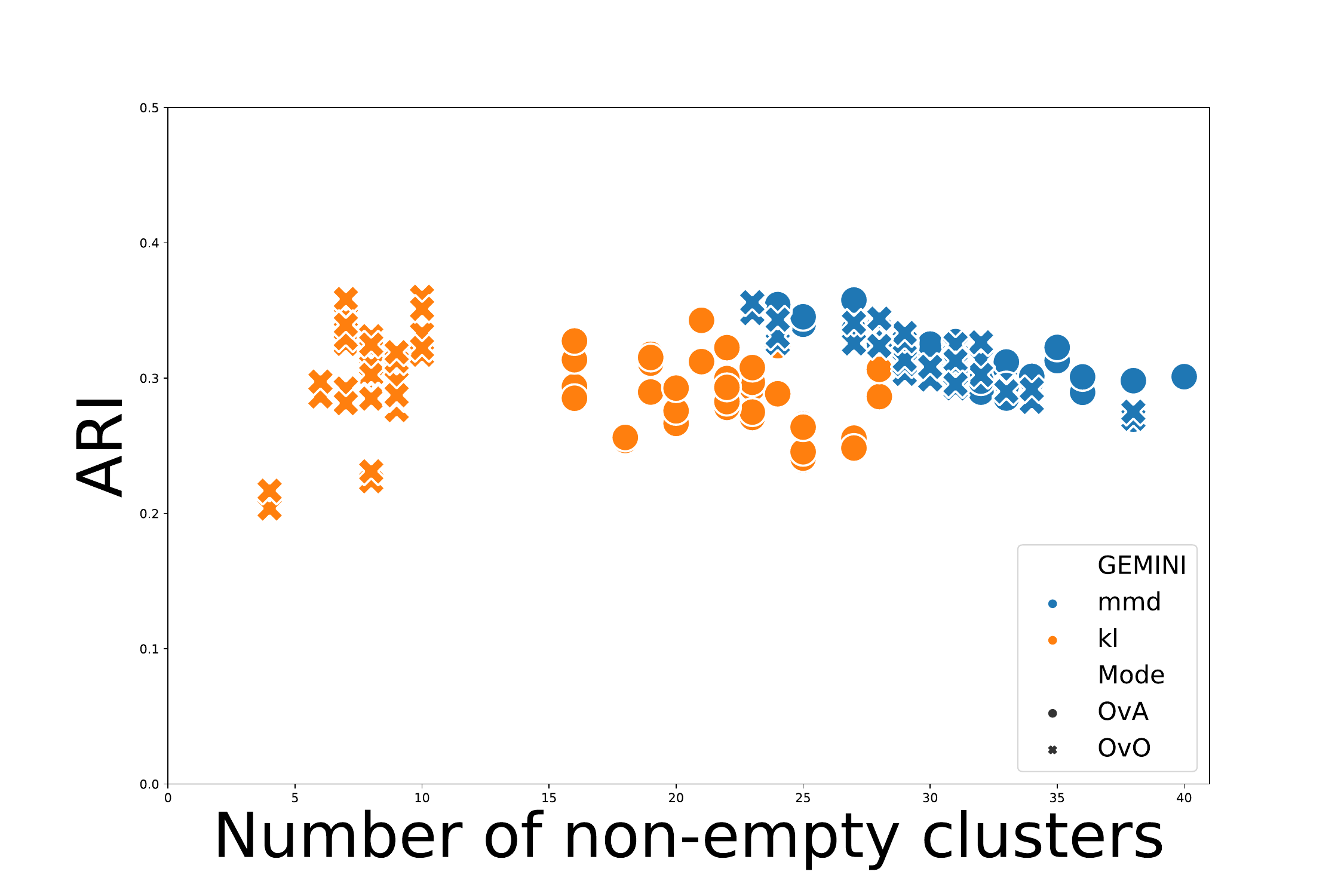}
        \label{sfig:mlp_clusters}
    }\hfil
    \subfloat[LeNet5]{
        \includegraphics[width=0.45\linewidth]{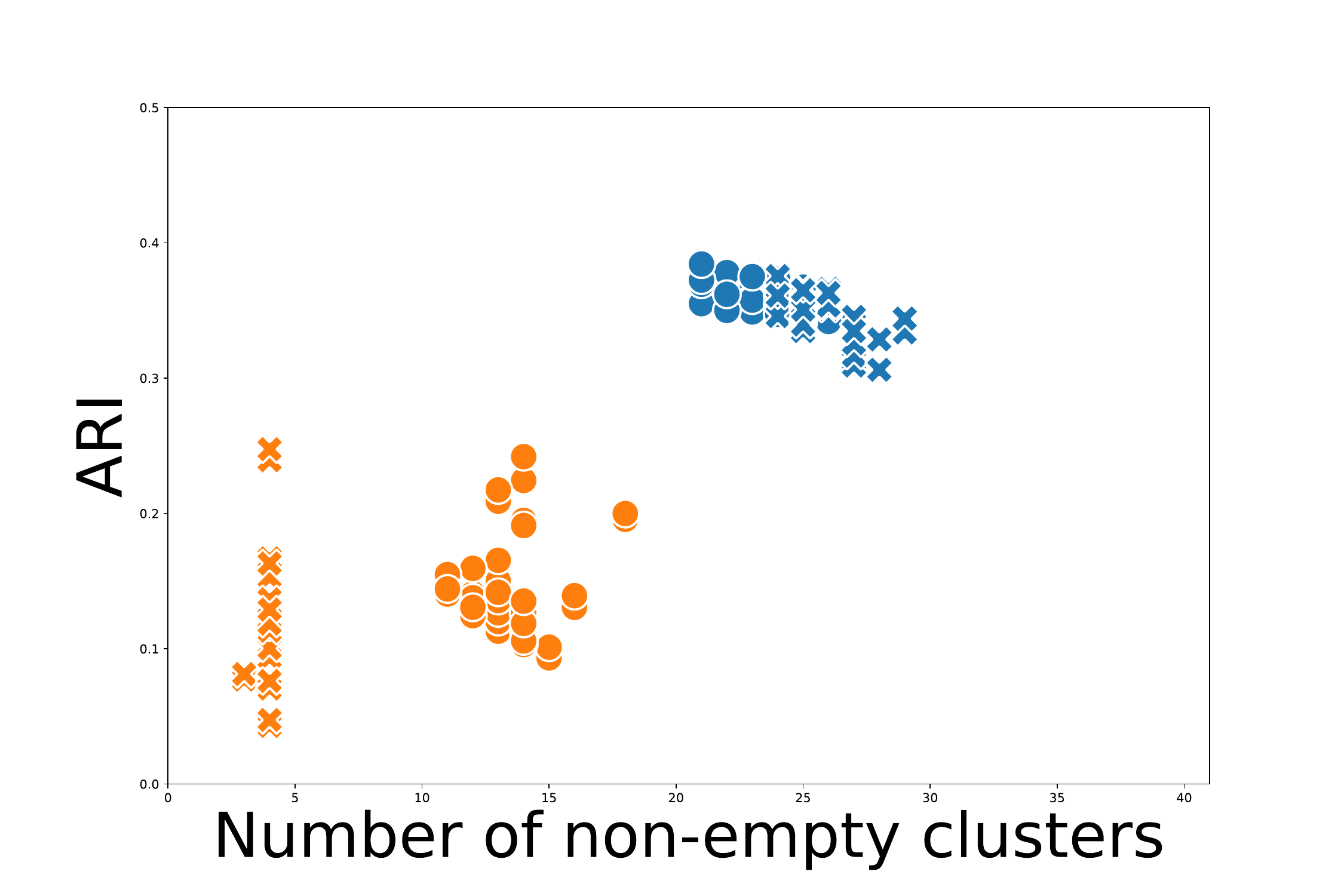}
        \label{sfig:lenet_clusters}
    }
    \caption{Distributions of the ARI scores given a number of non-empty clusters after 100 epochs of training on MNIST on two different architectures.}
    \label{fig:mnist_clusters}
\end{figure}

We repeat our previous experiment on the MNIST dataset from Sec.~\ref{ssec:exp_mnist}. We choose this time to get 50 clusters at best for both the MI and the MMD GEMINI and train the models for 100 epochs. We repeat the experiment 20 times per model and plot the resulting scores in figures~\ref{sfig:mlp_clusters} and~\ref{sfig:lenet_clusters}. We did not choose to test with the Wasserstein GEMINI because its complexity implies a long training time for 50 clusters, as explained in App.~\ref{app:exp_complexity}. We first observe in Fig.~\ref{fig:mnist_clusters} that the MMD-GEMINI with linear kernel has a tendency to exploit more clusters than the MI. The model converges to approximately 30 clusters in the case of the MLP and 25 for the LeNet-5 model with less variance. We can further observe that for all metrics the choice of architecture impacted the number of non-empty clusters after training. Indeed, by playing a key role in the decision boundary shape, the architecture may limit the number of clusters to be found: the MLP can draw more complex boundaries compared to the LeNet5 model. Moreover, we suppose that the cluster selection behaviour of GEMINI may be due to optimisation processes. Indeed, we optimise estimators of the GEMINI rather than the exact GEMINI.
Finally, Fig.~\ref{fig:mnist_clusters} also confirms from Table.~\ref{tab:mnist_experiment} the stability of the MMD-GEMINI regarding the ARI despite the change of architecture whereas the MI is affected and shows poor performance with the LeNet-5 architecture.

\subsection{Cifar10 clustering using a SIMCLR-derived kernel}
\label{ssec:exp_cifar10}
\begin{table*}
    \caption{ARI score of models trained for 200 epochs on CIFAR10 with different architectures using GEMINIs. The kernel for the MMD is either a linear kernel or the dot product between features extracted from a pretrained SIMCLR model. Both the Euclidean norm between images and SIMCLR features are considered for the Wasserstein metric. We report the ARI of related works when not using data augmentation for comparison.*: scores reported from Li et al., (2021)}
    \label{tab:simclr_kernel}
    \centering
    \begin{tabular}{c c c c c c c c c c}
    \toprule
    \multirow{2}{*}{Architecture}&No kernel&\multicolumn{4}{c}{Linear kernel / $\ell_2$ norm}&\multicolumn{4}{c}{SIMCLR \cite{chen_simple_2020}}\\
    \cmidrule(l){2-2}\cmidrule(lr){3-6}\cmidrule{7-10}
    &$\I_\text{KL}^\text{ova}$&$\I_\text{MMD}^\text{ova}$&$\I_\text{MMD}^\text{ovo}$&$\I_\mathcal{W}^\text{ova}$&$\I_\mathcal{W}^\text{ovo}$&$\I_\text{MMD}^\text{ova}$&$\I_\text{MMD}^\text{ovo}$&$\I_\mathcal{W}^\text{ova}$&$\I_\mathcal{W}^\text{ovo}$\\
    \midrule
    LeNet-5&0.026&0.049&0.048&0.043&0.041&{\bf 0.157}&0.145&0.079&0.138\\
    Resnet-18&0.008&0.047&0.044&0.037&0.036&0.122&{\bf 0.145}&0.052&0.080\\
    \midrule
    \multicolumn{3}{c}{KMeans (images / SIMCLR)}&0.041&0.147&\multicolumn{3}{c}{CC \cite{li_contrastive_2021}}&\multicolumn{2}{c}{0.030}\\
    \multicolumn{3}{c}{IDFD \cite{tao_clustering-friendly_2021}}&\multicolumn{2}{c}{0.060}&\multicolumn{3}{c}{JULE \cite{yang_joint_2016}*}&\multicolumn{2}{c}{0.138}\\
    \bottomrule
    \end{tabular}
\end{table*}

To further illustrate the benefits of the kernel or distance provided to GEMINIs, we continue the same experiment as in section~\ref{ssec:exp_mnist}. However, we focus this time on the CIFAR10 dataset. As improved distance, we chose a linear kernel and $\ell_2$ norm between features extracted from a pretrained SIMCLR model~\cite{chen_simple_2020}. We provide results for two different architectures: LeNet-5 and ResNet-18 both trained from scratch on raw images, the latter being a common choice of models in deep clustering literature~\cite{van_gansbeke_scan_2020,tao_clustering-friendly_2021}. We report the results in Table~\ref{tab:simclr_kernel} and provide the baseline of MI. We also write the baselines from related works when not using data augmentations to make a fair comparison. Indeed, models trained with GEMINIs do not use data augmentation: only the architecture and the kernel or distance function in the data space plays a role. We observe here that the choice of kernel or distance can be critical in GEMINIs. Indeed, while the Euclidean norm between images does not provide insights on how images of cats and dogs are far as shown by K-Means, features derived from SIMCLR carry much more insight on the proximity of images. This shows that the performances of GEMINIs depend on the quality of distance functions. Interestingly, we observe that for the Resnet-18 using SIMCLR features to guide GEMINIs was not as successful as it has been on the LeNet-5. We believe that the ability of this network to draw any decision boundary makes it equivalent to a categorical distribution model as in Sec.~\ref{ssec:exp_categorical}. Finally, to the best of our knowledge, we are the first to train from scratch a standard discriminative neural network on CIFAR raw images without using labels or direct data augmentations, while getting sensible clustering results. However, other recent methods achieve best scores using data augmentations which we do not~\cite{park_improving_2021}.

\subsection{A practical application with Graph Neural Networks: the Enron email dataset}
\label{ssec:exp_enron}
We focus in this experiment on the clustering of nodes in a graph representing email interaction between individuals in the Enron dataset (available at \url{https://www.cs.cmu.edu/~./enron/}. This famous company was filed for bankruptcy on the 2nd of December 2001 following an investigation for fraud by the Securities and Exchange Commision (SEC). Following Bouveyron et al.~\cite{bouveyron_2018_stochastic}, we focus on exchange of emails in the Enron corporation between the 1st of September 2021 and the 31st of December 2001, which corresponds to the peak period at which the company collapsed. We represent the dataset as a graph where each node corresponds to an employee and each edge represents the sending of at least one email. We choose to let the edges unweighted because the number of emails sent between two persons is not necessarily reflexive of how two persons can be close to each other. After filtering nodes that do not communicate between September and December, the graph comprises 141 nodes and 872 symmetric edges. In order to use the Wasserstein metric, we choose as a distance the all-pairs-shortest path as already used in Section~\ref{ssec:exp_moons} and Appendix~\ref{app:fw_distance}. Due to the necessity of using a \emph{symmetric} distance for the Wasserstein metric, we let the graph undirected. Thus, an edge between two persons represent the exchange of at least one email anyhow.

To cluster the nodes of the graph, we adopt a simple Graph Convolution Network (GCN)~\cite{kipf_2016_variational} inspired by Kipf et al.~\cite{kipf_2016_variational} and Liang et al.\cite{liang_2022_clustering} with a hidden size of dimension 64 and 10 clusters to find at best~\cite{bouveyron_2018_stochastic}. We vary the number of hidden layers from 1 to 3. All models were run 30 times. We perform clustering by maximising the OvO Wasserstein during 1000 epochs with a learning rate of $2\times10^{-3}$ for the Adam optimiser. Indeed, with a long training time, we allow the model to find more clusters despite a stabilised GEMINI. We also experimented using the MI with the same models and hyperparameters. We eventually select the final model using the highest GEMINI value for each different depth of GCN.

Similarly, we run 30 times as well two competitors with a number of clusters to find ranging from 2 to 10: the LPM model~\cite{hoff_2002_latent} and the Deep LPM model~\cite{liang_2022_clustering} which are generative methods based on the assumption of a latent position of the nodes in the graph determining their interaction. Their respective best models were selected according to the lowest bayesian information criterion and the highest evidence lower bound. For Deep LPM, we used the architecture proposed by the authors in~\cite{liang_2022_clustering} with a one-hidden-layer network.

\begin{figure}[!ht]
    \centering
    \subfloat[GEMINI complete graph]{\includegraphics[width=0.48\linewidth]{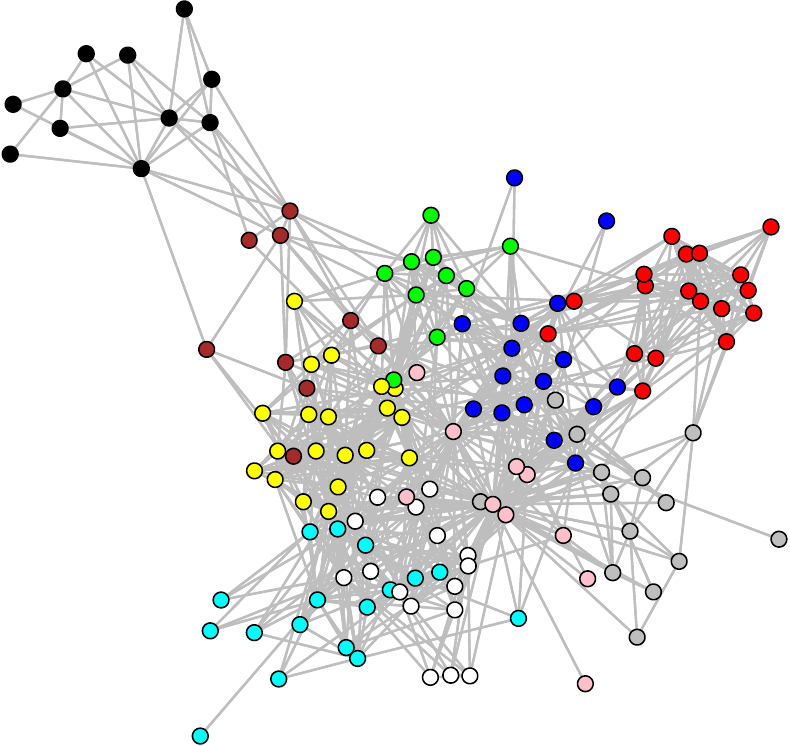}}\hfil
    \subfloat[GEMINI summary graph]{\includegraphics[width=0.48\linewidth]{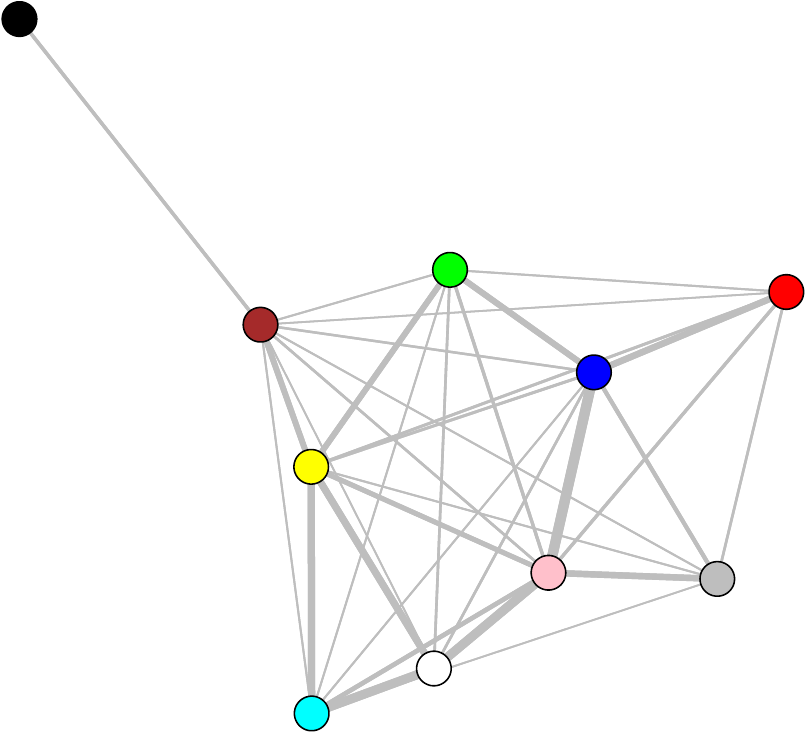}}\\
    \subfloat[Deep LPM complete graph]{\includegraphics[width=0.48\linewidth]{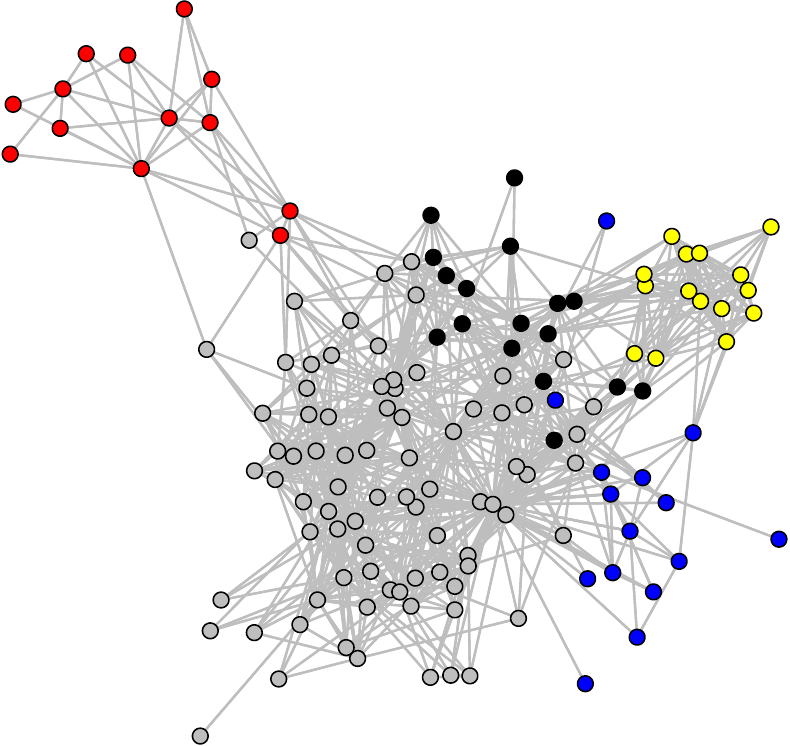}}\hfil
    \subfloat[Deep LPM summary graph]{\includegraphics[width=0.48\linewidth]{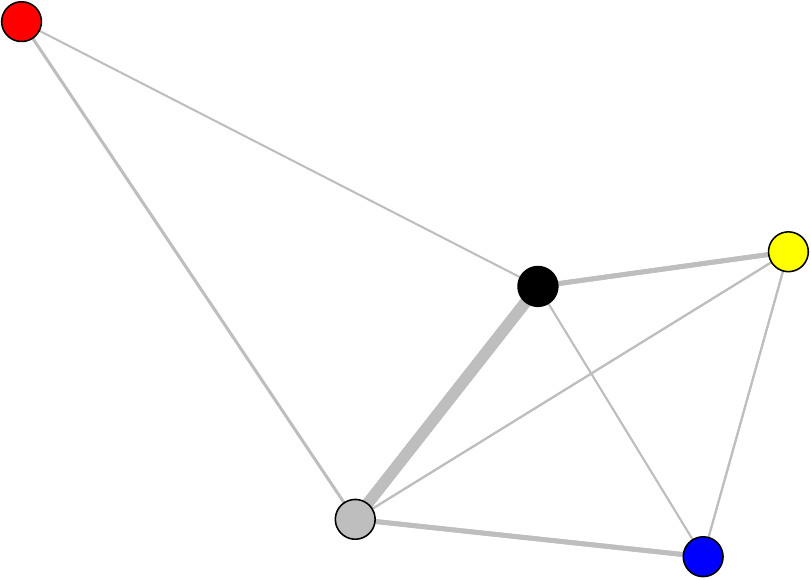}}
    \caption{Fruchterman Reingold representation of the Enron email interaction graph. Nodes are coloured according to clusters. The summary graph is a cluster-wise average of the positions of the nodes with edges as strong as the number of interactions between two clusters.}
    \label{fig:projection_clustering}
\end{figure}
\begin{figure*}[!ht]
    \centering
    \subfloat[GEMINI clustering]{\fbox{ \includegraphics[width=0.3\linewidth]{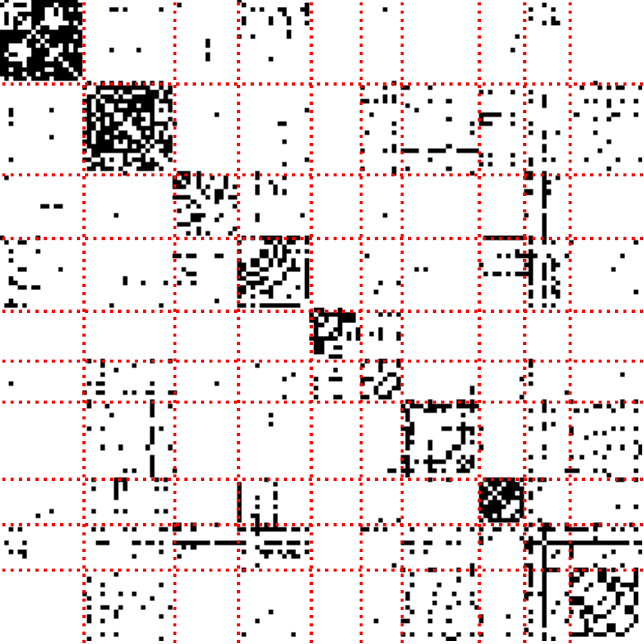}\label{sfig:enron_gemini_interaction}}}\hfil
    \subfloat[LPM clustering]{\fbox{ \includegraphics[width=0.3\linewidth]{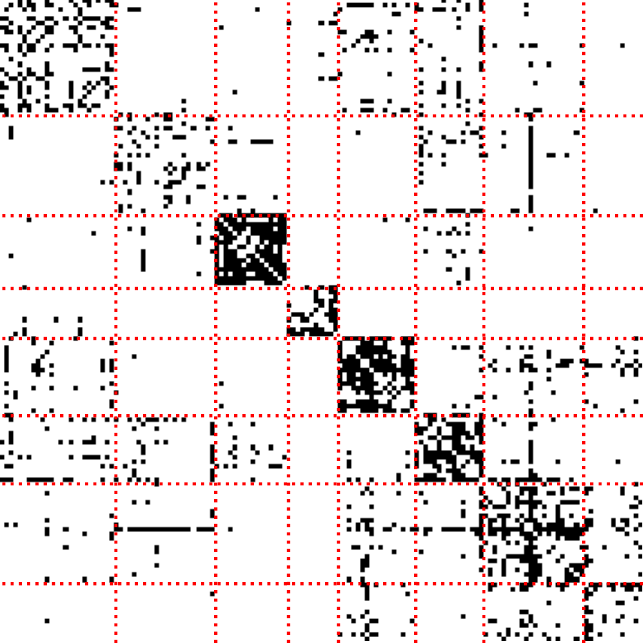}\label{sfig:enron_lpm_interaction}}}\hfil
    \subfloat[Deep LPM clustering]{\fbox{ \includegraphics[width=0.3\linewidth]{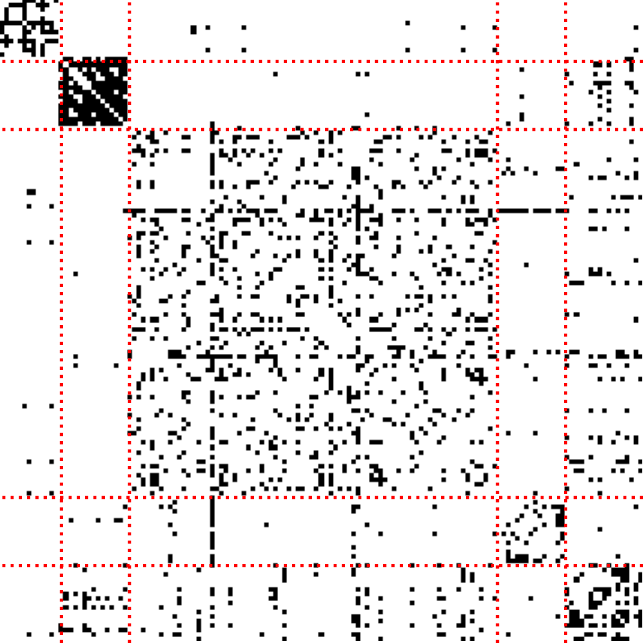}\label{sfig:enron_dlpm_interaction}}}
    \caption{Interaction matrices: each black cell encodes an edge. Nodes were reorganised according to their clustering for each model.}
    \label{fig:interaction_matrices}
\end{figure*}
\begin{figure}[!ht]
    \centering
    \subfloat[MI with 1 hidden layer\\PAC = 0.27]{
        \includegraphics[width=0.45\linewidth]{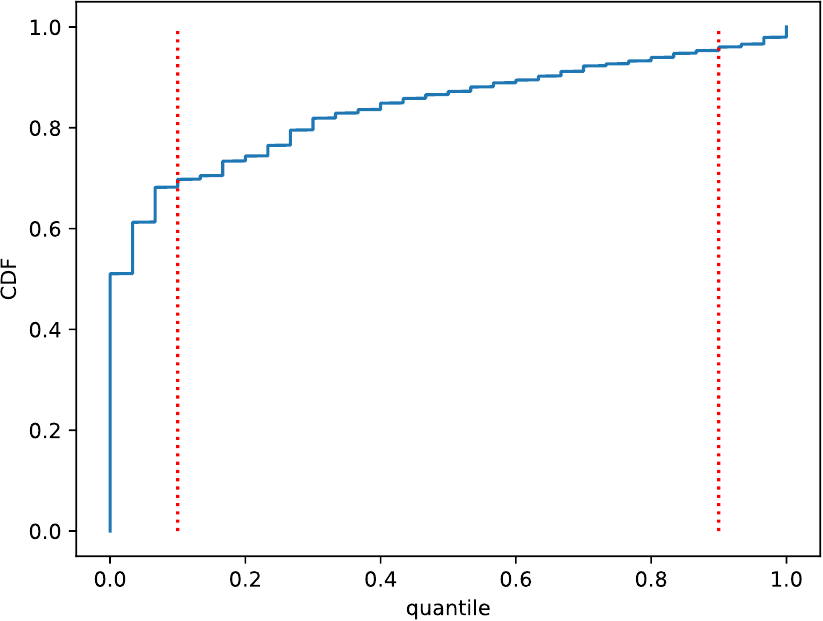}
        \label{sfig:pac_mi}
    }\hfil
    \subfloat[MI with 3 hidden layers\\PAC=0.36]{
        \includegraphics[width=0.45\linewidth]{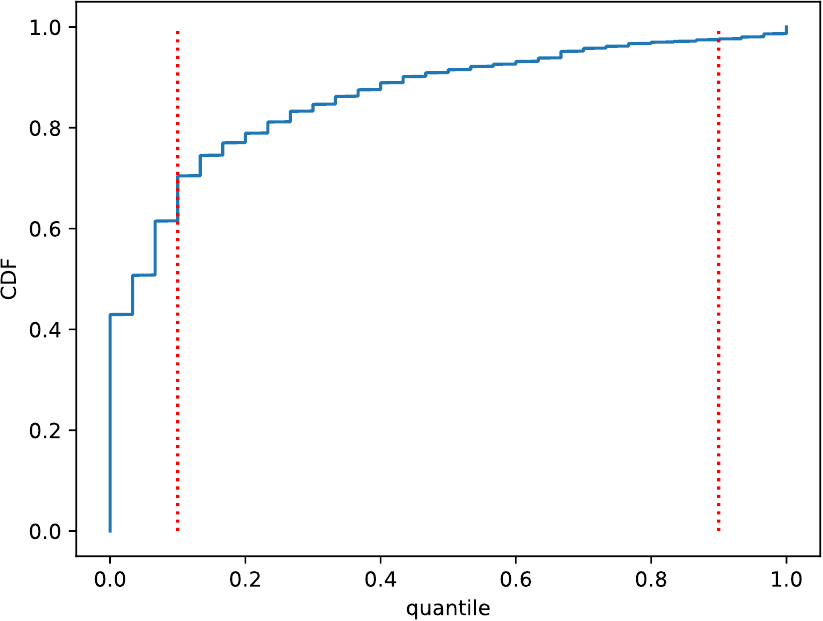}
        \label{sfig:pac_mi_double}
    }\\
    \subfloat[GEMINI with 1 hidden layer\\PAC=0.13]{
        \includegraphics[width=0.45\linewidth]{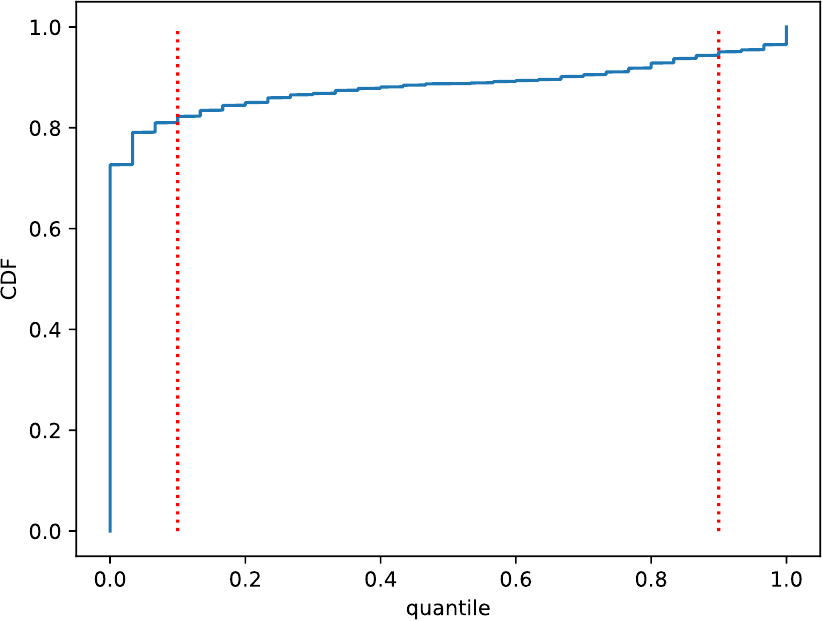}
        \label{sfig:pac_gemini}
    }\hfil
    \subfloat[GEMINI with 3 hidden layers\\PAC=0.22]{
        \includegraphics[width=0.45\linewidth]{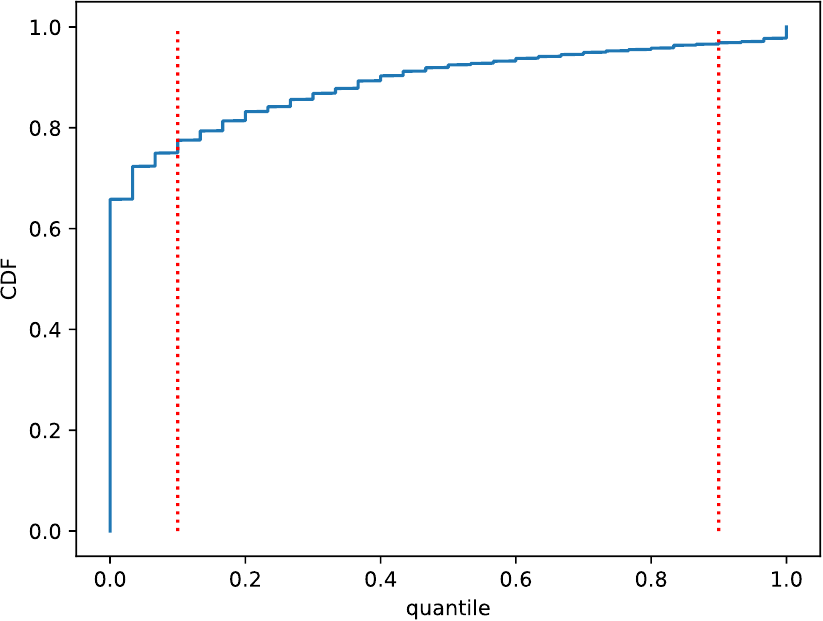}
        \label{sfig:pac_gemini_double}
    }
    \caption{CDFs of consensus between all clusterings produced by the models on the Enron Email dataset. The indicated PAC score is the difference of cdf between the quantiles 10\% and 90\%. A lower PAC score is better and highlights more consistent clustering assignments.}
    \label{fig:pac_scores}
\end{figure}
\begin{table}[!ht]
    \centering
    \caption{Average ARI scores between 30 GEMINI-trained models and the best Deep LPM and LPM clustering according to ELBO criterion. $H$ represents the number of hidden layers in the trained model.}
    \label{tab:ari_enron}
    \begin{tabular}{c c c c }
        \toprule
        GEMINI&$H$&LPM&Deep LPM\\
        \midrule
		\multirow{3}{*}{MI}&1&0.47 (0.08)&{\bf 0.31 (0.10)}\\
		&2&0.31 (0.07)&0.17 (0.09)\\
		&3&0.29 (0.04)&0.16 (0.05)\\        
        \cmidrule(lr){3-4}
		\multirow{3}{*}{OvO Wasserstein}&1&{\bf 0.58 (0.04)}&0.25 (0.03)\\
		&2&0.54 (0.06)&0.22 (0.03)\\
		&3&0.49 (0.06)&0.21 (0.04)\\
        \bottomrule
    \end{tabular}
\end{table}

We start by showing the highest Wasserstein-GEMINI clustering in Figure~\ref{fig:projection_clustering} where the graph nodes are positioned according to the Fruchterman Reingold algorithm~\cite{fruchterman_graph_1991}.

When we look at the interaction matrices of the best models in Figure~\ref{fig:interaction_matrices}, we can observe that the GEMINI clustering applied to graph using the all pair shortest path distance yields dense clusters where nodes intensively connect with each others (Figure~\ref{sfig:enron_gemini_interaction}). The LPM model found fewer clusters that are as well densely connected while the Deep LPM found 6 clusters including one densely connected and another one which contains multiple nodes sparsely connected. Interestingly, GEMINI managed to isolate graph nodes that act as hub e.g. the 9th cluster in Fig.~\ref{sfig:enron_gemini_interaction} with star-shaped interactions while Deep LPM mixes hubs in its 3rd cluster in Figure~\ref{sfig:enron_dlpm_interaction}. A potential explanation for this difference is that LPM models seek to cluster nodes based on supposedly close latent representations whereas the GEMINI clustering uses the Wasserstein distance, hence taking into account the flow of information passing through each graph node. In Table~\ref{tab:ari_enron}, we compared the average ARI between our 30 GEMINI models with the best selected LPM and Deep LPM clustering. It appears indeed that GEMINI models have a stronger ARI with the LPM model than with the Deep LPM, as the former concentrated on dense clusters. Interestingly, we can notice in Table~\ref{tab:ari_enron} that the MI models with one hidden layer are able to produce also satisfying results with an ARI close to or better than the ARI between GEMINI and LPM methods. However, when the models turn deeper, the performances of MI drop whereas the ARI between the Wasserstein-GEMINI decreases more slowly.

To further compare the performances of these models, we chose to evaluate their proportions of ambiguous clusters (PAC score)~\cite{senbabaouglu_2014_critical}. The PAC score is evaluated using two steps: we first build a matrix where each entry $i,j$ contains the proportion of times two samples $i$ and $j$ where clustered together by some model, then arrange these proportions to build a cdf which highlights for given quantiles the proportions of paired samples that can be clustered together. The PAC score corresponds to the difference of cdf between two given quantiles, commonly 10\% and 90\%. Through Figure~\ref{fig:pac_scores}, we can see that the models trained with the Wasserstein GEMINI have a lower PAC score compared to MI-trained models which highlights a more consistent clustering assignment through all 30 runs. We can further observe that for both objective functions, a deeper model leads to higher PAC score which we can explain by the difficulty to repeat the same decision boundary with deep models. Eventually, we may conclude that while the MI may have competitive results with shallow models compared to GEMINI and Deep LPM, it is unable to repeat a consistent clustering as illustrated by a high PAC score.


\section{Conclusions}
\label{sec:conclusions}
We highlighted that the choice of distance at the core of MI can alter the performances of deep learning models when used as an objective for a deep discriminative clustering. We first showed that MI maximisation does not necessarily reflect the best decision boundary in clustering when the clustering model converges to a Dirac distribution. We introduced the GEMINI, a method which only needs the specification of a neural network and a kernel or distance in the data space. Moreover, we showed how the notion of neighborhood built by the neural network can affect the clustering, especially for MI. To the best of our knowledge, this is the first method that trains single-stage neural networks from scratch using neither data augmentations nor regularisations, yet achieving good clustering performances. We emphasised that GEMINIs are only searching for a maximum number of clusters: after convergence some may be empty. However, we do not have insights to explain this convergence which is part of future work. Finally, we introduced several versions of GEMINIs and would encourage the OvA MMD or OvA Wasserstein as a default choice, since it proves to both incorporate knowledge from the data using a kernel or distance while remaining the less complex than OvO MMD and OvO Wasserstein in time and memory. OvO versions could be privileged for fine-tuning steps. As follow-up, we focused on the introduction of sparsity in GEMINI clustering for feature selection~\cite{ohl_sparse_2023}. Future works could focus on the joint learning of distances or kernels~\cite{wu_deep_2020} while maximising the the GEMINI to get both meaningful clustering and metrics in the data space. We will also investigate why sometimes after convergence some clusters end up empty.

\subsubsection*{Acknowledgements}

This work has been supported by the French government, through the 3IA C\^ote d'Azur, Investment in the Future, project managed by the National Research Agency (ANR) with the reference number ANR-19-P3IA-0002. We would also like to thank the France Canada Research Fund (FFCR) for their contribution to the project. This work was partly supported by the Fonds de recherche du Québec – Santé (FRQS) and the Health-Data Hub, through the joint project AORTIC STENOSIS . The authors are grateful to the OPAL infrastructure from Université Côte d'Azur for providing resources and support.

\bibliographystyle{abbrv}
\bibliography{bib}\vfill
\begin{IEEEbiography}[{\includegraphics[width=1in,height=1.25in,keepaspectratio]{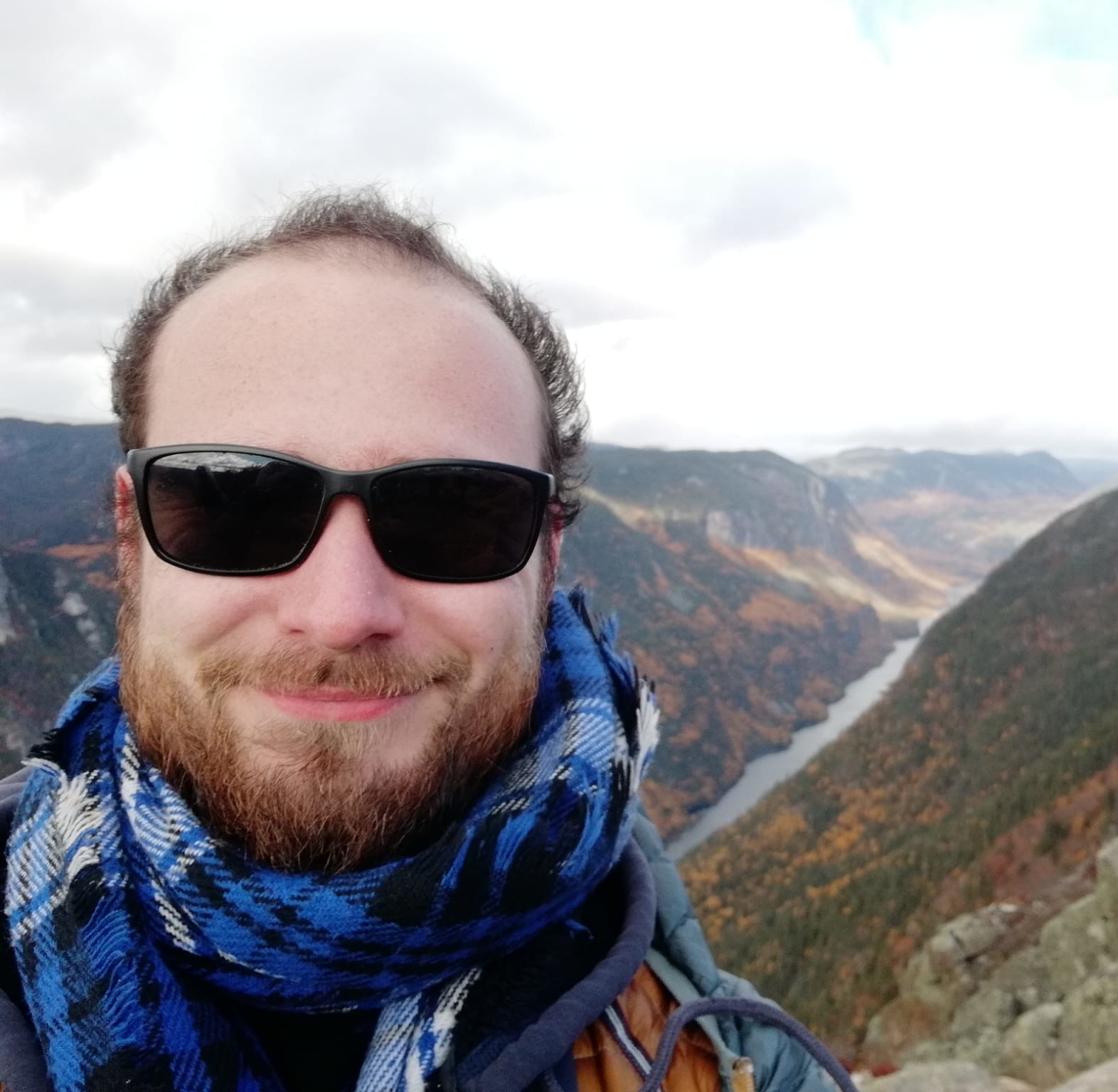}}]{Louis Ohl} 
PhD Student at the Universit\'e C\^ote d'Azur and the Universit\'e Laval from 2021 to 2024 and computer science graduate from the INSA Lyon and KTH Royal Institute of Technology in 2021. His project mainly focuses on clustering and the choices of the decision boundaries to discriminatively separate samples with applications in medical research including identification of phenogroups in aortic stenosis.
\end{IEEEbiography}\vspace{-2\baselineskip}
\begin{IEEEbiography}[{\includegraphics[width=1in,height=1.25in,keepaspectratio]{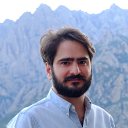}}]{Pierre-Alexandre Mattei}  is a research scientist at Inria, in the Maasai team located in Sophia-Antipolis (France). He graduated from Ecole Normale Supérieure de Cachan in 2014, and received a PhD in applied mathematics from Université Paris Cité in 2017. His research interests mostly revolve around problems with latent variables: deep generative modelling, Bayesian inference, clustering, missing data.
\end{IEEEbiography}\vspace{-2\baselineskip}
\begin{IEEEbiographynophoto}{Charles Bouveyron} Professor of Statistics at Université Côte d'Azur, Nice, France. He holds a Chair on Artificial Intelligence and is the Director of the Institut 3IA Côte d'Azur and head of the research team MAASAI from INRIA. His research interests include statistical learning for clustering, classification and regression in high dimension, statistical learning on networks applied to different domains (medicine, image analysis, astrophysics, humanities...)    
\end{IEEEbiographynophoto}\vspace{-2\baselineskip}
\begin{IEEEbiography}[{\includegraphics[width=1in,height=1.25in,keepaspectratio]{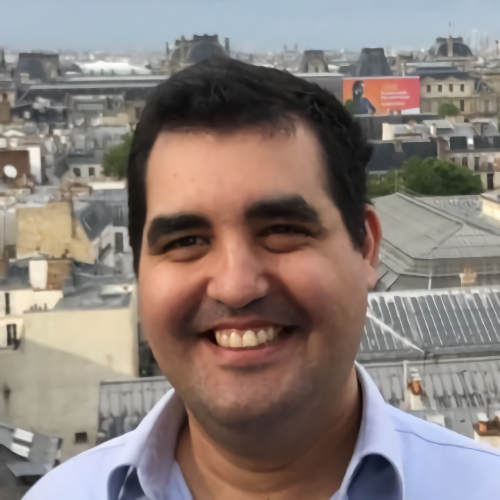}}]{Warith Harchaoui} holds a Ph.D. in Applied Mathematics and is deeply engaged in the intersection of artificial intelligence (AI) and pattern analysis.
He commenced his academic journey at École Normale Supérieure de Cachan, where he received his M.Sc. in 2008. His initial forays into the industry revolved around Computer Vision applications in startups and global enterprises.
Harchaoui extended his academic pursuits at École Normale Supérieure de Paris, contributing to research at the globally recognized Willow laboratory. His scholarly interests encompass a range of topics within image, sound, video, and text processing, a domain where he has consistently leveraged mathematical models for innovative solutions.
From 2014 to 2020, Dr. Harchaoui also embraced a corporate role in Data Science, working with Oscaro.com the e-commerce leader in the automotive parts sector. Balancing industry demands with academic rigor, he completed his Ph.D. research between 2016 and 2020.
Today, Dr. Harchaoui serves at Jellysmack since 2021, where he combines his expertise to help internet video creators through every stage from content production, editing to social media promotion.
\end{IEEEbiography}\vspace{-2\baselineskip}
\begin{IEEEbiography}[{\includegraphics[width=1in,height=1.25in,keepaspectratio]{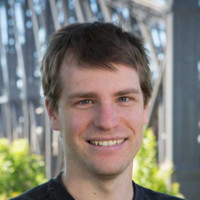}}]{Mickaël Leclercq}
PhD in Computer Science from McGill University, Canada, in 2016. Since then, he's been working as a research associate in the computational biology laboratory of Prof. Droit, researcher at the CHU de Québec - Université Laval. His role is to develop and supervise bioinformatics research projects with an artificial intelligence component. He also contributes to the organization and coordination of the laboratory's research activities and supervises the work of students and young research professionals. He actively participates in the long-term vision of the laboratory's research themes as well as in the development of grant applications and manuscripts.
\end{IEEEbiography}\vfill
\begin{IEEEbiography}[{\includegraphics[width=1in,height=1.25in,keepaspectratio]{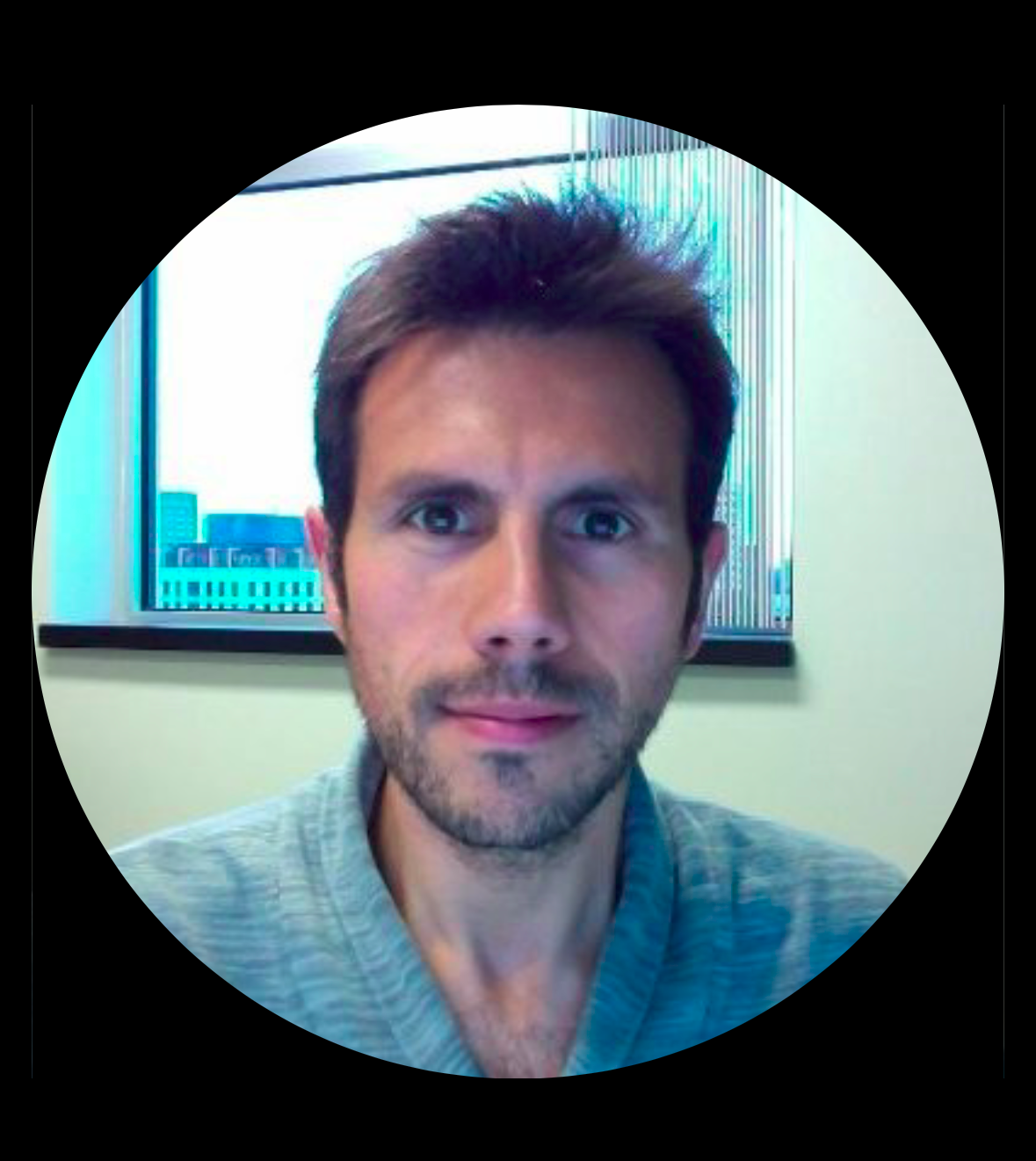}}]{Arnaud Droit}
PhD graduate in 2007 in bioinformatics from Université Laval, Canada. Since 2012, he has been a Professor at Université Laval (Québec, Canada) and leads a bioinformatics and proteomics laboratory at the CHU de Québec. His research focuses on robust bioinformatics approaches, advanced statistical methods, and machine/deep learning strategies to extract information relevant to medical research from large biological datasets. Many of his projects are related to cancer research, where the aim is to discover biomarker signatures associated with various clinical outcomes. His expertise is essential to many research sectors requiring bioinformatics, so he built a large network of international collaborations, including the private sector. 
\end{IEEEbiography}\vspace{-2\baselineskip}
\begin{IEEEbiographynophoto}{Frederic Precioso} 
Full Professor at the Universit\'e C\^ote d'Azur from 2011 and Affiliated Professor at Universit\'e Laval from 2022, his research interests cover active learning, foundation models, hybrid learning combining symbolic and non-symbolic models, clustering, MLOps, applied to different domains (health, biology, multimedia, autonomous vehicles, biodiversity).
\end{IEEEbiographynophoto}\vfill

\newpage
\onecolumn

\appendices
\section{Demonstration of the convergence to 0 of the MI for a Gaussian Mixture}
\label{app:mi_convergence}
\newcommand{\nmuA}{\mathcal{N}(x|\mu_0,\sigma^2)}
\newcommand{\nmuB}{\mathcal{N}(x|\mu_1,\sigma^2)}

\label{app:mi_boundaries}
\subsection{Models definition}
Let us consider a mixture of two Gaussian distributions, both with different means $\mu_0$ and $\mu_1$, s.t. $\mu_0<\mu_1$ and of same standard deviation $\sigma$:

\begin{equation*}
p(x|y=0) = \nmuA, p(x|y=1) = \nmuB ,
\end{equation*}

where $y$ is the cluster assignment. We take balanced clusters proportions, i.e. $p(y=0)=p(y=1)=\frac{1}{2}$. This first model is the basis that generated the complete dataset $p(x)$. When performing clustering with our discriminative model, we are not aware of the distribution. Consequently: we create other models. We want to compute the difference of mutual information between two decision boundaries that discriminative models $p_\theta(y|x)$ may yield.

We define two decision boundaries: one which splits evenly the data space called $p_A$ and another which splits it on a closed set $p_B$:

\begin{equation}\label{eq:p_a}
p_A(y=1|x) = \left\{ \begin{array}{c c}
1-\epsilon&x>\frac{\mu_1-\mu_0}{2}\\
\epsilon&\text{otherwise}
\end{array}\right. ,
\end{equation}

\begin{equation}\label{eq:p_b}
p_B(y=1|x) = \left\{ \begin{array}{c c}
1-\epsilon&x \in [\mu_0, \mu_1]\\
\epsilon&\text{otherwise}
\end{array}\right. .
\end{equation}

Our goal is to show that both models $p_A$ and $p_B$ will converge to the same value of mutual information as $\epsilon$ converges to 0.

\subsection{Computing cluster proportions}
\subsubsection{Cluster proportion of the correct decision boundary}

To compute the cluster proportions, we estimate with samples $x$ from the distribution $p_\text{data}(x)$. Since we are aware for this demonstration of the true nature of the data distribution, we can use $p(x)$ for sampling. Consequently, we can compute the two marginals:

\begin{align*}
p_A(y=1) &= \int_\mathcal{X} p(x) p_A(y=1|x)dx ,\\
&= \int_{-\infty}^{\frac{\mu_1-\mu_0}{2}}p(x)\epsilon dx + \int_{\frac{\mu_1-\mu_0}{2}}^{+\infty}p(x)(1-\epsilon)dx ,\\
&=\epsilon \left(\int_{-\infty}^{\frac{\mu_1-\mu_0}{2}}p(x)dx \right) + (1-\epsilon) \left( \int_{\frac{\mu_1-\mu_0}{2}}^{+\infty}p(x)dx\right) ,\\
&= \frac{1}{2} .
\end{align*}

\subsubsection{Cluster proportion of the misplaced decision boundary}
For the misplaced decision boundary, the marginal is different:

\begin{equation*}\begin{split}
p_B(y=1) &= \int_\mathcal{X} p(x) p_B(y=1|x)dx ,\\
&= \epsilon\left(\int_{-\infty}^{\mu_0}p(x)dx + \int_{\mu_1}^{+\infty}p(x)dx\right) + (1-\epsilon) \int_{\mu_0}^{\mu_1}p(x)dx ,\\
&=\epsilon \left(1-\int_{\mu_0}^{\mu_1}p(x)dx\right) + (1-\epsilon)\int_{\mu_0}^{\mu_1}p(x)dx .
\end{split}\end{equation*}

Here, we simply introduce a new variable named $\beta$ that will be a shortcut for noting the proportion of data between $\mu_0$ and $\mu_1$:

\begin{equation*}
\beta = \int_{\mu_0}^{\mu_1}p(x)dx .
\end{equation*}

And so can we simply write the cluster proportion of decision boundary model B as:

\begin{equation*}\label{eq:py_b}
p_B(y=1) = \epsilon (1-\beta) + (1-\epsilon)\beta ,
\end{equation*}

Leading to the summary of proportions in Table~\ref{tab:proportions}. For convenience, we will write the proportions of model B using the shortcuts:

\begin{equation*}\label{eq:pi_b}
\pi_B = p_B(y=1) = \epsilon + \beta(1-2\epsilon) ,
\end{equation*}
\begin{equation*}
\bar{\pi}_B = p_B(y=0) = 1-\epsilon - \beta(1-2\epsilon) .
\end{equation*}

\begin{table}[hbt]
\caption{Proportions of clusters for models A and B}\label{tab:proportions}
\centering
\begin{tabular}{c c c}
\toprule
$\mathcal{M}$&A&B\\
\cmidrule(lr){2-3}
$p_\mathcal{M}(y=1)$&$\frac{1}{2}$&$\epsilon+\beta(1-2\epsilon)$\\
$p_\mathcal{M}(y=0)$&$\frac{1}{2}$&$1-\epsilon-\beta(1-2\epsilon)$\\
\bottomrule
\end{tabular}
\end{table}

\subsection{Computing the KL divergences}
\subsubsection{Correct decision boundary}

We first start by computing the Kullback-Leibler divergence for some arbitrary value of $x\in\mathbb{R}$:

\begin{equation*}
D_\text{KL}(p_A(y|x)||p_A(y))= \sum_{i=0}^1 p_A(y=i|x) \log{\frac{p_A(y=i|x)}{p_A(y=i)}} .
\end{equation*}

We now need to detail the specific cases, for the value of $p(y=i|x)$ is dependent on $x$. We start $\forall x <\frac{\mu_1-\mu_0}{2}$:

\begin{align*}
D_\text{KL}(p_A(y|x)||p_A(y))&= p_A(y=0|x)\log{\frac{p_A(y=0|x)}{\frac{1}{2}}} + p_A(y=1|x)\log{\frac{p_A(y=1|x)}{\frac{1}{2}}} ,\\
&=(1-\epsilon)\log{2(1-\epsilon)} + \epsilon \log{2\epsilon} .
\end{align*}

The opposite case, $\forall x\geq \frac{\mu_1-\mu_0}{2}$ yields:
\begin{align*}
D_\text{KL}(p_A(y|x)||p_A(y))&= p_A(y=0|x)\log{\frac{p_A(y=0|x)}{\frac{1}{2}}} + p_A(y=1|x)\log{\frac{p_A(y=1|x)}{\frac{1}{2}}} ,\\
&=\epsilon\log{2\epsilon} + (1-\epsilon) \log{2(1-\epsilon)} .
\end{align*}

Since both cases are equal, we can write down:

\begin{equation}\label{eq:correct_kl_div}
D_\text{KL}(p_A(y|x)||p_A(y)) = \epsilon\log{2\epsilon} + (1-\epsilon)\log{2(1-\epsilon)} ,\forall x\in\mathbb{R} .
\end{equation}

\subsubsection{Misplaced boundary}

We proceed to the same detailing of the Kullback-Leibler divergence computation for the misplaced decision boundary. We start with the set $x\in [\mu_0,\mu_1]$:

\begin{align*}
D_\text{KL}(p_B(y|x)||p_B(y))&= p_B(y=0|x) \log{\frac{p_B(y=0|x)}{p_B(y=0)}} + p_B(y=1|x)\log{\frac{p_B(y=1|x)}{p_B(y=1)}} ,\\
&=\epsilon \log{\frac{\epsilon}{\bar{\pi}_B}} + (1-\epsilon) \log{\frac{1-\epsilon}{\pi_B}} .
\end{align*}

When $x$ is out of this set, the divergence becomes:

\begin{align*}
D_\text{KL}(p_B(y|x)||p_B(y))&=p_B(y=0|x) \log{\frac{p_B(y=0|x)}{p_B(y=0)}} + p_B(y=1|x)\log{\frac{p_B(y=1|x)}{p_B(y=1)}} ,\\
&=(1-\epsilon)\log{\frac{1-\epsilon}{\bar{\pi}_B}} + \epsilon \log{\frac{\epsilon}{\pi_B}} .
\end{align*}

To fuse the two results, we will write the KL divergence as such:

\begin{equation}\label{eq:odd_kl_div}
D_\text{KL}(p_B(y|x)||p_B(y)) = \epsilon\log{\epsilon}+ (1-\epsilon)\log(1-\epsilon) - C(x) ,\forall x\in\mathbb{R} ,
\end{equation}

where $C(x)$ is a constant term depending on $x$ defined by:

\begin{equation*}\label{eq:odd_constant}
C(x)=\left\{\begin{array}{c c}
\epsilon\log{\bar{\pi}_B} + (1-\epsilon)\log{\pi_B}&x\in[\mu_0,\mu_1]\\
\epsilon\log{\pi_B} + (1-\epsilon)\log{\bar{\pi}_B}&x\in\mathbb{R}\setminus[\mu_0,\mu_1]\\
\end{array}  .\right.
\end{equation*}

For simplicity of later writings, we will shorten the notations by:

\begin{equation*}
C(x)=\left\{ \begin{array}{c c}
\alpha_1&x\in[\mu_0,\mu_1]\\
\alpha_0&x\in\mathbb{R}\setminus[\mu_0,\mu_1]
\end{array}\right. .
\end{equation*}

\subsection{Evaluating the mutual information}

\subsubsection{Correct decision boundary}

We inject the value of the Kullback-Leibler divergence from Eq.~(\ref{eq:correct_kl_div}) inside an expectation performed over the data distribution $p_\text{data}(x)$:

\begin{align}
\mathcal{I}_A(x;y) &= \mathbb{E}_{x\sim p_\text{data}(x)}\left[ D_\text{KL}(p_A(y|x)||p_A(y))\right] ,\\
&=\int_\mathcal{X}p(x) \left(\epsilon \log(2\epsilon) + (1-\epsilon)\log(2(1-\epsilon)) \right)dx ,\\
&=\epsilon \log(2\epsilon) + (1-\epsilon)\log(2(1-\epsilon)) .\label{eq:mi_a}
\end{align}

Since the KL divergence was independent of $x$, we could leave the constant outside of the integral which is equal to 1.

We can assess the coherence of Eq.~(\ref{eq:mi_a}) since its limit as $\epsilon$ approaches 0 is $\log 2$. In terms of bits, this is the same as saying that the information on $X$ directly gives us information on the $Y$ of the cluster.

\subsubsection{Odd decision boundary}

We inject the value of the KL divergence from Eq.~(\ref{eq:odd_kl_div}) inside the expectation of the mutual information:

\begin{align*}
\mathcal{I}_B(x;y) &= \mathbb{E}_{x\sim p_\text{data}(x)}\left[ D_\text{KL}(p_B(y|x)||p_B(y))\right] ,\\
&=\int_{\mathcal{X}}p(x)\left(\epsilon\log\epsilon +(1-\epsilon)\log(1-\epsilon) -C(x)\right) ,dx\\
&=\epsilon\log\epsilon +(1-\epsilon)\log(1-\epsilon) - \int_{\mathcal{X}}p(x)C(x)dx .
\end{align*}

The first terms are constant with respect to $x$ and the integral of $p(x)$ over $\mathcal{X}$ adds up to 1. We finally need to detail the expectation of the constant $C(x)$ from Eq.~(\ref{eq:odd_constant}):

\begin{align*}
\mathbb{E}_x[C(x)] &= \int_{-\infty}^{\mu_0}C(x)p(x)dx + \int_{\mu_0}^{\mu_1}C(x)p(x)dx + \int_{\mu_1}^{+\infty}C(x)p(x)dx ,\\
&= \alpha_0\left(\int_{-\infty}^{\mu_0}p(x)dx + \int_{\mu_1}^{+\infty}p(x)dx \right) + \alpha_1\int_{\mu_0}^{\mu_1}p(x)dx ,\\
&=\alpha_0(1-\beta) + \alpha_1\beta .
\end{align*}

This can be further improved by unfolding the description of $\alpha_0$ and $\alpha_1$ from Eq.~(\ref{eq:odd_constant}):

\begin{align*}
\alpha_0(1-\beta)+\beta\alpha_1&= \alpha_0 +\beta(\alpha_1-\alpha_0),\\
&=\epsilon\log{\pi_B}+(1-\epsilon)\log{\bar{\pi}_B} + \beta\left[\epsilon\log{\bar{\pi}_B}+(1-\epsilon)\log{\pi_B}\right.\\&\quad\left.-\epsilon\log{\pi_B}- (1-\epsilon)\log{\bar{\pi}_B}\right],\\
&= \left[1-\epsilon+\beta\epsilon-\beta+\beta\epsilon\right]\log{\bar{\pi}_B} + \left[\epsilon+\beta-\beta\epsilon-\beta\epsilon\right]\log{\pi_B},\\
&=\log{\bar{\pi}_B} + \left[2\beta\epsilon-\beta-\epsilon\right]\log{\frac{\bar{\pi}_B}{\pi_B}} .
\end{align*}

We can finally write down the mutual information for the model B:

\begin{equation*}\label{eq:mi_b}
\mathcal{I}_B(x;y) = \epsilon\log{\epsilon} +(1-\epsilon)\log(1-\epsilon) -\log{\bar{\pi}_B} - \left[2\beta\epsilon-\beta-\epsilon\right]\log{\frac{\bar{\pi}_B}{\pi_B}} .
\end{equation*}

\subsection{Differences of mutual information}

Now that we have the exact value of both mutual informations, we can compute their differences:

\begin{align*}
\Delta_\mathcal{I} &= \mathcal{I}_A(x;y) - \mathcal{I}_B(x;y),\\
&= \epsilon\log(2\epsilon) + (1-\epsilon)\log(2(1-\epsilon)) - \epsilon\log{\epsilon} -(1-\epsilon)\log(1-\epsilon)\\&\quad+\log{\bar{\pi}_B} + \left[2\beta\epsilon-\beta-\epsilon\right]\log{\frac{\bar{\pi}_B}{\pi_B}} ,\\
&=\epsilon\log{2}+(1-\epsilon)\log2 +\log{\bar{\pi}_B} + \left[2\beta\epsilon-\beta-\epsilon\right]\log{\frac{\bar{\pi}_B}{\pi_B}} .
\end{align*}

We then deduce how the difference of mutual information evolves as the decision boundary becomes sharper, i.e. when $\epsilon$ approaches 0:

\begin{equation*}
\lim_{\epsilon\rightarrow 0}\Delta_\mathcal{I} = \log{2} + \log{\bar{\pi}_B} -\beta\log{\frac{\bar{\pi}_B}{\pi_B}} .
\end{equation*}

However, the cluster proportions by B $\pi_B$ also take a different value as $\epsilon$ approaches 0. Recalling Eq.~(\ref{eq:p_b}):

\begin{equation*}
\lim_{\epsilon\rightarrow 0} p_B(y=1) = \beta , \lim_{\epsilon\rightarrow 0} p_B(y=0) = 1-\beta .
\end{equation*}

And finally can we write that:

\begin{equation*}\begin{split}
\lim_{\epsilon\rightarrow 0} \Delta_\mathcal{I} &= \log{2} + \log(1-\beta) - \beta \log\frac{1-\beta}{\beta} ,\\
&=\log{2} + (1-\beta)\log(1-\beta) + \beta \log{\beta} ,\\
&=\log{2} -\mathcal{H}(\beta) .
\end{split}\end{equation*}

\begin{figure}[hbt]
    \centering
    \subfloat[Differences of MI between models A and B]{
        \includegraphics[width=0.45\linewidth]{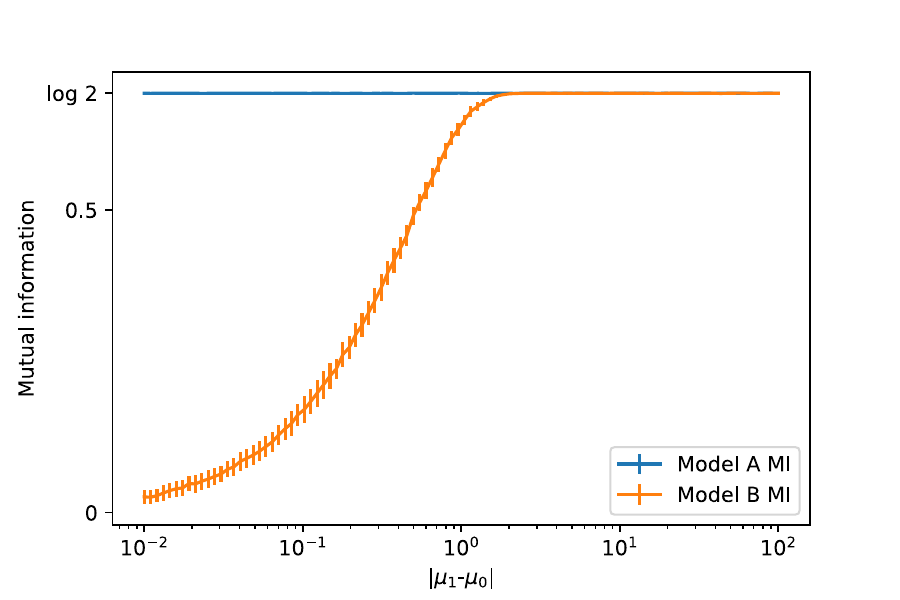}
        \label{sfig:mi_convergence_differences}
    }\hfil
    \subfloat[Gaussian mixture distribution $p(x)$ with proportion $\beta$ in between the two means $\mu_0$ and $\mu_1$]{
        \includegraphics[width=0.45\linewidth]{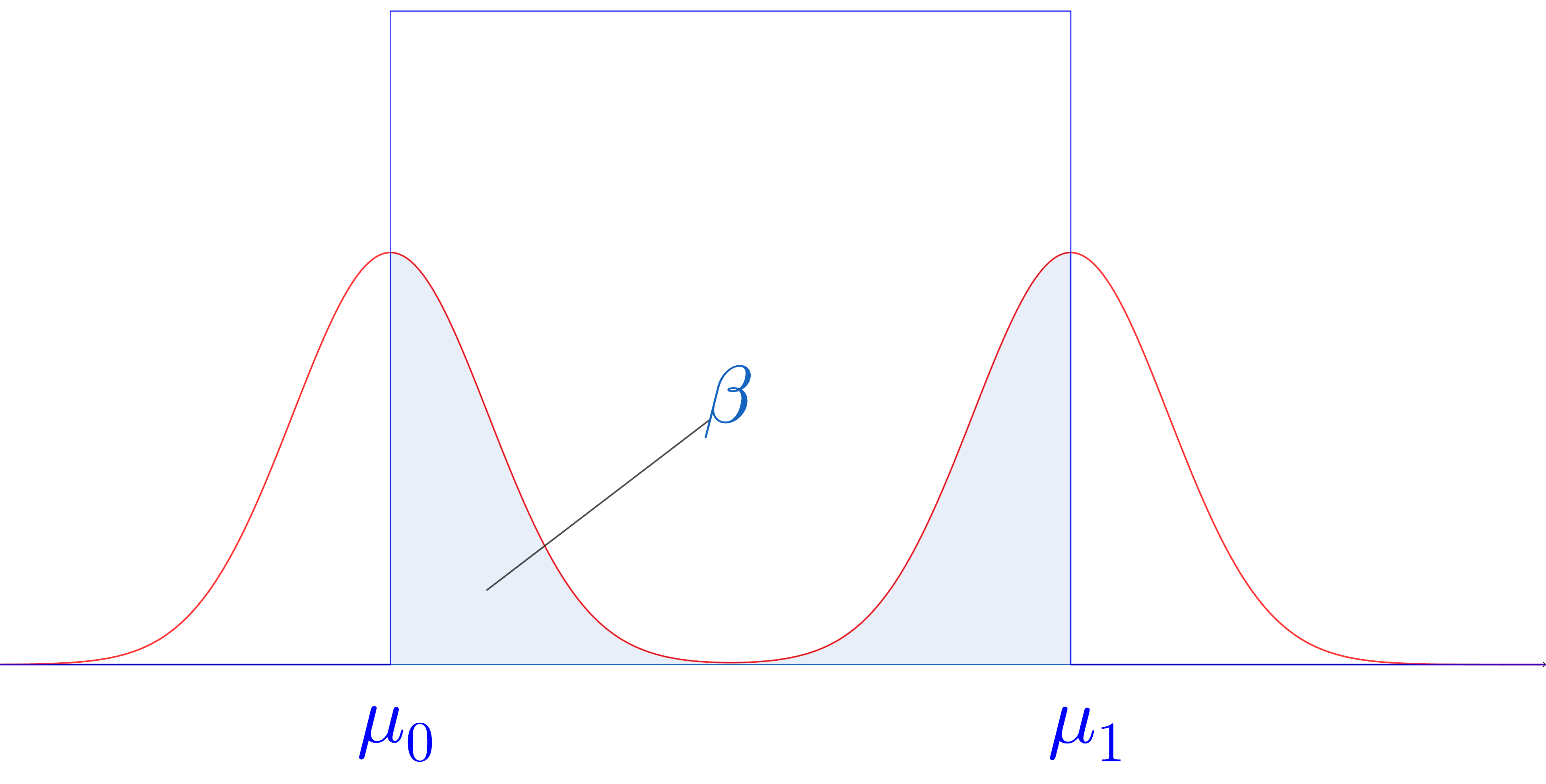}
        \label{sfig:recap_gaussian_mixture}
    }
    \caption{Value of the mutual information for the two models splitting a Gaussian mixture depending on the distance between the two means $\mu_0$ and $\mu_1$ of the two generating Gaussian distributions. We estimate the MI by computing it 50 times on 1000 samples drawn from the Gaussian mixture.}
    \label{fig:mi_convergence_differences}
\end{figure}

To conclude, as the decision boundaries turn sharper, i.e. when $\epsilon$ approaches 0, the difference of mutual information between the two models is controlled by the entropy of proportion of data $\beta$ between the two means $\mu_0$ and $\mu_1$. We know that for binary entropies, the optimum is reached for $\beta=0.5$. In other words having $\mu_0$ and $\mu_1$ distant enough to ensure balance of proportions between the two clusters of model B leads to a difference of mutual information equal to 0. We experimentally highlight this convergence in Figure~\ref{fig:mi_convergence_differences} where we compute the mutual information of models A and B as the distance between the two means $\mu_0$ and $\mu_1$ increases in the Gaussian distribution mixture.

\section{Proof of Proposition~\ref{prop:alpha_div_maximisation}}
\label{app:proof_alpha_div_maximisation}
For the sake of clarity, we will use the notations $\pi_k\equiv p(y=k)$ during the demonstration.

\subsection{Modelisation of the conditional distribution}

We will consider two types of models. The first one is the \emph{generic} clustering model, where the cluster assignment follows a categorical distribution:

\begin{equation*}
y|\vec{x} \sim \text{Cat}(\psi_\theta(\vec{x})),
\end{equation*}
where $\psi_\theta : \mathcal{X} \rightarrow \Delta_{K}$ is a learnable function of parameters $\theta$ and $\Delta_{K}$ is a $K$-simplex.

The second model is a Dirac distribution where the data space $\mathcal{X}$ is divided into a partition $\mathcal{X}_k,\forall k \in \{1,\cdots,K\}$:

\begin{equation*}
\pykx = \pmb{1}_{[\vec{x}\in\mathcal{X}_k]},
\end{equation*}
with $\pmb{1}$ the indicator function. This is simply a sub-case of the generic model.

For both models, we consider that clusters are not empty and that the model is not degenerate, i.e. $\pi_k \in ]0,1[ \forall k \in \{1,K\}$.

\subsection{Value of the GEMINI OvA and upper bounds}

We first unfold the OvA GEMINI for the generic model and $\alpha\in \mathbb{R}\setminus\{0,1\}$:

\begin{align*}
\I^\text{ova}_{D_\alpha}(\vec{x};y) &= \E_{y\sim p(y)} \left[\E_{\vec{x} \sim \px} \left[f_\alpha\left(\frac{p(\vec{x}|y)}{\px}\right) \right] \right],\\
&=\sum_{k=1}^K \pi_k \int_{\mathcal{X}}\px \left(\frac{\pxyk^\alpha \px^{-\alpha}}{\alpha(\alpha-1)} - \frac{\pxyk \px^{-1}}{\alpha-1}+\frac{1}{\alpha}\right)d\vec{x},\\
&= \sum_{k=1}^K \pi_k \int_{\mathcal{X}} \left(\frac{\px\pykx^\alpha }{\pi_k^\alpha \alpha(\alpha-1)} - \frac{\pxyk}{\alpha-1} + \frac{\px}{\alpha}\right)d\vec{x}.
\end{align*}

After distributing the factor $\px$ in the integral, we can notice that the two last terms will be summed to 1, up to a factor depending on $\alpha$.

\begin{align*}
\I^\text{ova}_{D_\alpha}(\vec{x};y) &= \sum_{k=1}^K \pi_k \left(\frac{1}{\alpha}-\frac{1}{\alpha-1} + \frac{1}{\pi_k^\alpha \alpha(\alpha-1)}\mathbb{E}_{\vec{x}\sim \px}\left[\pykx^\alpha\right]\right)\\
&= \frac{-1}{\alpha(\alpha-1)} + \frac{1}{\alpha(\alpha-1)}\sum_{k=1}^K \pi_k^{1-\alpha} \E_{\vec{x}\sim \px}\left[\pykx^\alpha\right],\\
&= \left(\alpha(\alpha-1)\right)^\alpha \left[-1 + \sum_{k=1}^K \pi_k^{1-\alpha}\E_{\vec{x}\sim \px}\left[\pykx^\alpha\right] \right].
\end{align*}

Since we face a categorical distribution, we can affirm that $p(y|\vec{x}) \in [0,1]$. Therefore, depending on the value of $\alpha$, we have either $\pykx^\alpha \in [1, \infty[$ if $\alpha$ is negative, and $\pykx\in [0,1]$ for $\alpha$ positive. We now inspect these different cases.

\subsection{Case of \texorpdfstring{$\alpha \in ]1,+\infty[$}{alpha greater than 1}}

The upper bound we can get on the expectation involves the inequality $\pykx^\alpha \leq \pykx$. Owing to the linearity of the expectation, we can affirm that:

\begin{align*}
\E_{\vec{x}\sim \px} \left[\pykx^\alpha\right] &\leq \E_{\vec{x}\sim \px} [\pykx],\\
&\leq \pi_k.
\end{align*}

This allows us to derive the following upper bound on the OvA GEMINI:

\begin{align*}
\I^\text{ova}_{D_\alpha}(\vec{x};y) &=\left(\alpha(\alpha-1)\right)^{-1} \left[ -1 + \sum_{k=1}^K \pi_k^{1-\alpha} \E_{\vec{x}\sim \px}\left[\pykx^\alpha\right]\right],\\
&\leq \left(\alpha(\alpha-1)\right)^{-1}\left[-1+\sum_{k=1}^K\pi_k^{2-\alpha} \right],\\
&\leq \mathcal{B}_{]1,+\infty[}(\pi_1,\cdots,\pi_K).
\end{align*}

The upper bound $\mathcal{B}_{]1,+\infty[}(\pi_1,\cdots,\pi_K)$ is a convex function that is invariant to permutations of $\pi_k$. Interestingly, this upper bound only depends on the proportions of the clusters. Its maximimum is reached when $\pi_k=K^{-1}$. However, any solution is acceptable for the special case of $\alpha=2$, i.e. the Pearson $\chi^2$-divergence. In this case, the upper bound is a constant: $\mathcal{B}^+_2 = \frac{K-1}{2}$. In this situation, the proportions of the clusters do not matter. Only getting a Dirac model is sufficient to maximise the OvA GEMINI.

We can further conclude that the bound is tight when considering a Dirac model since $1^\alpha=1$ and $0^\alpha=0$, leading to:

\begin{align*}
\E_{\vec{x}\sim \px}\left[ \pmb{1}_{\left[\vec{x}\in\mathcal{X}_k\right]}^\alpha\right] &= \E_{\vec{x}\sim \px}\left[ \pmb{1}_{\left[\vec{x}\in\mathcal{X}_k\right]}\right],\\
&=\pi_k.
\end{align*}

And so do we conclude:

\begin{align*}
    \I^\text{ova}_{D_\alpha}(\vec{x};y) &=\left(\alpha(\alpha-1)\right)^{-1} \left[ -1 + \sum_{k=1}^K \pi_k^{1-\alpha} \E_{\vec{x}\sim \px}\left[\pykx^\alpha\right]\right],\\
    &= \left(\alpha(\alpha-1)\right)^{-1}\left[-1+\sum_{k=1}^K\pi_k^{2-\alpha} \right],\\
    &= \mathcal{B}_{]1,+\infty[}(\pi_1,\cdots,\pi_K)
\end{align*}

\subsection{Case of a \texorpdfstring{$\alpha \in ]0,1[$}{alpha between 0 and 1}}

In this case, the front factor $(\alpha(\alpha-1))^{-1}$ is negative. Thus, we are interested in minimising the second term and finding the lower bound. We can already infer:
\begin{align*}
\pykx &\leq \pykx^\alpha\\
\E_{\vec{x}\sim \px}[\pykx]&\leq \E_{\vec{x}\sim \px} \left[ \pykx^\alpha\right]\\
\pi_k&\leq \E_{\vec{x}\sim \px} \left[ \pykx^\alpha\right].
\end{align*}

This lower bound is tight for a Dirac model. We can finally compute for the OvA GEMINI that:

\begin{align*}
\I^\text{ova}_{D_\alpha}(\vec{x};y) &=\left(\alpha(\alpha-1)\right)^{-1} \left[ -1 + \sum_{k=1}^K \pi_k^{1-\alpha} \E_{\vec{x}\sim \px}\left[\pykx^\alpha\right]\right],\\
&\leq \left(\alpha(\alpha-1)\right)^{-1}\left[-1+\sum_{k=1}^K\pi_k^{2-\alpha} \right],\\
&\leq \mathcal{B}_{]0,1[}.
\end{align*}

Hence, we conclude that $\mathcal{B}_{]0,1[}=\mathcal{B}_{]1,+\infty[} = \mathcal{B}_{\mathbb{R}^{+*}\setminus \{1\}}$. In both cases, using a Dirac model implies that the OvA GEMINI reaches its upper bound.

\subsection{Case of a negative \texorpdfstring{$\alpha$}{alpha}}

The upper bound of the OvA GEMINI in this case is the infinity. Indeed, taking the example of the Dirac model is sufficient to consider regions of the data space $\mathcal{X}$ where the clustering distribution has no support. Thus, the expectation is undefined, or rather drifts towards infinity.

\subsection{Specific case of \texorpdfstring{$\alpha=1$}{alpha equal to 1}, the KL divergence}

In this case, we need to start the computations all over again using the definition $f(t)=t\log{t}$. Indeed, we can skip the term $-t +1$ since it does not affect the value of an $f$-divergence, i.e. for any convex function $f$ s.t. $f(1)=0$ and for any real constant $c$:

\begin{equation*}
D_{f(t)}(p\|q) = D_{f(t)+c(t-1)}(p\|q).
\end{equation*}

We thus get:

\begin{align*}
\I_{D_1}^\text{ova}(\vec{x};y) &= \E_{y\sim p(y)}\left[\E_{\vec{x} \sim \px}\left[\frac{\pxyk}{\px} \log\frac{\pxyk}{\px}\right]\right],\\
&= \sum_{k=1}^K \pi_k \int_\mathcal{X} \pxyk \log\left(\frac{\pxyk}{\px}\right)d\vec{x},\\
&= \sum_{k=1}^K \int_\mathcal{X} \pykx \px \log \left(\frac{\pykx}{\pi_k} \right)d\vec{x}.
\end{align*}

We can then separate the log term to make appear the two different entropies contributing to the mutual information:

\begin{align*}
\I_{D_1}^\text{ova}(\vec{x};y) &= \sum_{k=1}^K \int_\mathcal{X} \pykx \px\log(\pykx)d\vec{x} - \log{\pi_k}\int_\mathcal{X} \pykx \px d\vec{x},\\
&= \sum_{k=1}^K \int_\mathcal{X} \pykx \px\log(\pykx)d\vec{x} - \pi_k\log\pi_k
\end{align*}

We find once again an upper bound on the integral depending on $p(y|\vec{x})$. We know that the function $g: t\mapsto t\log{t}$ is convex and below 0 for $t\in [0,1]$. Hence:

\begin{equation*}
\pykx\log{\pykx} \leq 0,
\end{equation*}

with strict equality iff $\pykx \in \{0,1\}$. This implies that the Dirac model maximises the left integral. We deduce the upper bound of the mutual information:

\begin{align*}
\I_{D_1}^\text{ova}(\vec{x};y) &= \sum_{k=1}^K \int_\mathcal{X} \pykx \px\log(\pykx)d\vec{x} - \pi_k\log\pi_k,\\
&\leq \sum_{k=1}^K -\pi_k \log{\pi_k},\\
&\leq \mathcal{B}_1.
\end{align*}

This shows that the cluster proportion entropy is the upper bound of the mutual information. It is reached for any Dirac model.

\subsection{Last subcase: the null alpha}

In this case, the $\alpha$-divergence is defined by the function $f(t) = -\log{t}$. Let us derive again the OvA GEMINI:

\begin{align*}
\I_{D_0}^\text{ova}(\vec{x};y) &= \E_{y\sim p(y)}\left[\E_{\vec{x} \sim \px}\left[- \log\frac{\pxyk}{\px}\right]\right],\\
&= -\sum_{k=1}^K \pi_k \int_\mathcal{X} \px \log\left(\frac{\pxyk}{\px}\right)d\vec{x},\\
&= -\sum_{k=1}^K \pi_k\int_\mathcal{X} \px \log \left(\frac{\pykx}{\pi_k} \right)d\vec{x}.
\end{align*}

We expand again the logarithm and compute the integral over constant terms factorised by $\px$:

\begin{align*}
\I_{D_0}^\text{ova}(\vec{x};y) &= \sum_{k=1}^K \pi_k\log{\pi_k} - \pi_k\int_\mathcal{X} \px\log{\pykx}d\vec{x}.
\end{align*}

Now we can see that this OvA GEMINI may converge to infinity. Indeed, for the example of the Dirac model, we evaluate the integral with terms worth $\lim_{t\rightarrow 0}\log{t}$. We cannot conclude on the upper bounds of this case.

\subsection{Maximal upper bound}

We have shown so far that for $\alpha >0$, we can derive two different upper bounds that only depend on the proportions of the clusters $\pi_k$. These upper bounds can be reached by Dirac model of type $\pykx=\pmb{1}_{[\vec{x}\in\mathcal{X}_k]}$.

We can now question for the two upper bounds, $\mathcal{B}_{\mathbb{R}^{+*}\setminus\{1\}}$ and $\mathcal{B}_1$ what are the optimal cluster proportions $\pi_k$. By adding a Lagrangian constraint to enforce $\sum_{k=1}^K \pi_k=1$ in each upper bound, we can show that the maximal upper bound is reached iff $\pi_k=K^{-1} \forall k$. This concludes our proof.

\section{Proof of Proposition~\ref{prop:ovo_greater_ova}}
\label{app:proof_ovo_greater_ova}
Let us consider the value of the OvO GEMINI when the distance $D$ is an $f$-divergence or an IPM.

\subsection{Demonstration for \texorpdfstring{$f$}{f}-divergences}
We first need to highlight that $f$-divergences come with a conjugate convex function $g$. This conjugate enables the inversion of the arguments of the $f$-divergence:

\begin{equation*}
D_f(p\|q) = D_g(q\|p),
\end{equation*}
for any distributions $p$ and $q$. We can use this trick to revert first the $f$-divergence between the distribution $\pykx$ and $\px$:

\begin{equation*}
D_f(\pxyk\|\px) = D_g(\px\|\pxyk).
\end{equation*}

We then write $\px$ as a sum marginalising the $y$ variable. Using the convexity of the function $g$, we get a weighted upper bound of this divergence:

\begin{align*}
D_g(\px\|\pxyk)&= D_g\left(\sum_{k^\prime=1}^K p(y=k^\prime)p(\vec{x}|y=k^\prime) \| \left(\sum_{k^\prime=1}^K p(y=k^\prime)\right) \pxyk\right),\\
&\leq \sum_{k^\prime=1}^K p(y=k^\prime) D_g(p(\vec{x}|y=k^\prime) \| \pxyk),\\
&\leq \E_{k^\prime \sim p(y)} \left[ D_g(p(\vec{x}|y=k^\prime)\|\pxyk)\right].
\end{align*}

To retrieve the OvO form, we can compute the expectation of this inequality over all possible combinations of $p(y)$:

\begin{align*}
\E_{y \sim p(y)}\left[D_f(\px\|\pxyk)\right] &\leq \E_{y_1,y_2 \sim p(y)} \left[ D_g (p(\vec{x}|y_1)\|p(\vec{x}|y_2))\right],\\
\I^\text{ova}_{D_f}(\vec{x};y) &\leq \I^\text{ovo}_{D_g}(\vec{x};y).
\end{align*}

Then, owing to the conjugate convex functions, we can observe that for any $k, k^\prime \in \{1,\cdots, K\}$:

\begin{multline*}
D_g(\pxyk\|p(\vec{x}|y=k^\prime)) + D_g(p(\vec{x}|y=k^\prime)\|\pxyk) = D_f(p(\vec{x}|y=k^\prime)\|\pxyk)\\+ D_f(\pxyk\|p(\vec{x}|y=k^\prime)).
\end{multline*}

Consequently, the symmetry of OvO in its double expectation implies that:

\begin{equation*}
\I^\text{ovo}_{D_f}(\vec{x};y) = \I^\text{ovo}_{D_g}(\vec{x};y).
\end{equation*}

And so do we conclude that:

\begin{equation}
\I^\text{ova}_{D_f}(\vec{x};y) \leq \I^\text{ovo}_{D_f}(\vec{x};y).
\end{equation}

\subsection{Demonstration for IPMs}

For IPMs, we start from the OvA distance between the distribution of an arbitrary cluster $i$ among $K$:

\begin{equation*}
D_\text{IPM}(p(\vec{x}|y=i)\|\px) = \sup_{f\in\mathcal{F}} \E_{\vec{x}\sim p(\vec{x}|y=i)}[f(\vec{x})] - \E_{\vec{x}\sim \px}[f(\vec{x})].
\end{equation*}

We can extend the expectation of the data distribution over all clusters using the sum rules of probabilities:

\begin{align*}
D_\text{IPM}(p(\vec{x}|y=i)\|\px)&= \sup_{f\in\mathcal{F}} \E_{\vec{x}\sim p(\vec{x}|y=i)}[f(\vec{x})] - \sum_{k=1}^Kp(y=k)\E_{\vec{x}\sim \pxyk}[f(\vec{x})],\\
&=\sup_{f\in\mathcal{F}} \sum_{k=1}^K p(y=k) \left(\E_{\vec{x}\sim p(\vec{x}|y=i)}[f(\vec{x})] - \E_{\vec{x}\sim \pxyk}[f(\vec{x})]\right).
\end{align*}

We used in the second line the fact that the sum of $\sum_{k=1}^Kp(y=k)=1$ to factorise the first expectation. Finally, we can use the property that the supremum of a sum is lower or equal than a sum of suprema:

\begin{align*}
D_\text{IPM}(p(\vec{x}|y=i)\|\px) &\leq \sum_{k=1}^K \sup_{f\in\mathcal{F}} p(y=k)\left(\E_{\vec{x}\sim p(\vec{x}|y=i)}[f(\vec{x})] - \E_{\vec{x}\sim \pxyk}[f(\vec{x})]\right),\\
&\leq \sum_{k=1}^K p(y=k) D_\text{IPM} (p(\vec{x}|y=i)\|\pxyk).
\end{align*}

This upper bound corresponds to the expectation of the IPM between a specific cluster distribution $i$ and all other cluster distributions. Finally, by performing the expectation over all cluster of index $i$,  we can conclude that:

\begin{equation}
\I^\text{ova}_\text{IPM}(\vec{x};y) \leq \I^\text{ovo}_\text{IPM}(\vec{x};y).
\end{equation}

\section{Proof of Proposition~\ref{prop:equality_ova_ovo_ipm}}
\label{app:proof_equality_ova_ovo_ipm}
We will show here that when the cluster variable $y$ is binary, the OvA and OvO GEMINIs are equal if we use IPMs for distance between distributions. Indeed we can first unfold both equations:

\begin{equation}\label{eq:unfold_ova}\begin{split}
\I_\text{IPM}^\text{ova}(\vec{x},y) &= \E_{y\sim p(y)}\left[D_\text{IPM}(p(\vec{x}|y)\|\px\right]\\
&=p(y=0)D_\text{IPM}(p(\vec{x}|y=0)\|p(\vec{x})) + p(y=1)D_\text{IPM}(p(\vec{x}|y=1)\|\px)\\
&= p(y=0)\sup_{f\in\mathcal{F}} \{ \E_{\vec{x}\sim p(\vec{x}|y=0)}[f(\vec{x})] - \E_{\vec{x}\sim \px}[f(\vec{x})] \} \\&\quad+ p(y=1) \sup_{g\in\mathcal{F}} \{\E_{\vec{x}\sim p(\vec{x}|y=1)}[g(\vec{x})] - \E_{\vec{x}\sim \px}[g(\vec{x})]\},
\end{split}\end{equation}

and for the OvO, we simply use the symmetric property of IPMs:

\begin{equation}\label{eq:unfold_ovo}\begin{split}
\I_\text{IPM}^\text{ovo}(\vec{x},y) &=\E_{\ya,\yb \sim p(y)}\left[D_\text{IPM}(p(\vec{x}|\ya)\|p(\vec{x}|\yb))\right]\\
&=p(y=0)p(y=1)D_\text{IPM}(p(\vec{x}|y=0)||p(\vec{x}|y=1)) \\&\qquad+ p(y=1)p(y=0)D_\text{IPM}(p(\vec{x}|y=1)||p(\vec{x}|y=0))\\
&= 2 p(y=0)p(y=1) D_\text{IPM}(p(\vec{x}|y=0)||p(\vec{x}|y=1)).
\end{split}\end{equation}

Notice that in Eq. (\ref{eq:unfold_ovo}), we skipped the terms where both the random variables $y_1$ and $y_2$ are equal, since the implied distance is necessarily 0.

Now, to show the equivalence of both equations (\ref{eq:unfold_ova}) and (\ref{eq:unfold_ovo}), we simply need to write the sum rule of probabilities leading to the marginalisation of $\vec{x}$:

\begin{equation*}
\px=p(\vec{x}|y=0)p(y=0) + p(\vec{x}|y=1)p(y=1).
\end{equation*}

Thus, we can rewrite the expectations depending on the distribution $\px$ with other distributions for any function $f$:

\begin{equation*}
\E_{\vec{x}\sim \px}[f(\vec{x})] = p(y=0)\E_{\vec{x} \sim p(\vec{x}|y=0)}[f(\vec{x})] + p(y=1)\E_{\vec{x}\sim p(\vec{x}|y=1)} [f(\vec{x})]],
\end{equation*}

which we can incorporate back into Eq. (\ref{eq:unfold_ova}) to get:

\begin{multline*}
\I_\text{IPM}^\text{ova}(\vec{x},y) = p(y=0)\sup_{f\in\mathcal{F}}\{(1-p(y=0))\E_{\vec{x}\sim p(\vec{x}|y=0)}[f(\vec{x})] - p(y=1)\E_{\vec{x}\sim p(\vec{x}|y=1)} [f(\vec{x})]]\}\\+p(y=1)\sup_{g\in\mathcal{F}} \{ (1-p(y=1))\E_{\vec{x}\sim p(\vec{x}|y=1)}[g(\vec{x})] - p(y=0)\E_{\vec{x} \sim p(\vec{x}|y=0)}[g(\vec{x})]\}.
\end{multline*}

Since we only use two clusters, we know that $p(y=1)=1-p(y=0)$. This helps us factorising terms inside the sup expressions:

\begin{equation*}\begin{split}
\I_\text{IPM}^\text{ova}(\vec{x},y) &= p(y=0)\sup_{f\in\mathcal{F}} \left\{p(y=1)\left[ \E_{\vec{x}\sim p(\vec{x}|y=0)}[f(\vec{x})] - \E_{\vec{x}\sim p(\vec{x}|y=1)}[f(\vec{x})]\right] \right\}\\
&\qquad p(y=1)\sup_{g\in\mathcal{F}} \left\{ p(y=0)\left[ \E_{\vec{x}\sim p(\vec{x}|y=1)}[g(\vec{x})] - \E_{\vec{x}\sim p(\vec{x}|y=0)}[g(\vec{x})]\right]\right\}.
\end{split}\end{equation*}

Eventually, the factors $p(y=0)$ and $p(y=1)$ do not depend on the functions $f$ and $g$, so we can pull them out of the supremum. The remaining expressions are then symmetric and can be thus factorised:

\begin{equation*}\label{eq:final_ova}\begin{split}
\I_\text{IPM}^\text{ova}(\vec{x},y) &= 2p(y=0)p(y=1)\sup_{f\in\mathcal{F}} \left\{ \E_{\vec{x}\sim p(\vec{x}|y=0)}[f(\vec{x})] - \E_{\vec{x}\sim p(\vec{x}|y=1)}[f(\vec{x})]\right\}\\
&= 2p(y=0)p(y=1)D_\text{IPM}(p(\vec{x}|y=0)||p(\vec{x}|y=1))\\
&=\I_\text{IPM}^\text{ovo}(\vec{x},y)
\end{split}\end{equation*} 

This concludes the proof.

\section{Proof of Proposition~\ref{prop:ovo_fdiv_maximisation}}
\label{app:proof_ovo_fdiv_maximisation}
For any $f$-divergence and two distribution $p$ and $q$ taking value in the space $\mathcal{X}$, then disjoint support between $p$ and $q$ implies the maximisation of the $f$-divergence. Indeed, the bounds of an $f$-divergence are:

\begin{equation*}
0 \leq D_f(p,q) \leq f(0)+g(0),
\end{equation*}
where the upper bound can be infinity depending on $f$ and its convex conjugate $g: t\longrightarrow tf(1/t)$. Thus, for any two different clusters $k\neq k^\prime$:

\begin{equation*}
0\leq D_f(p(\vec{x}|y=k)\|p(\vec{x}|y=k^\prime)) \leq f(0)+g(0).
\end{equation*}

However, distributions for the same clusters have an $f$-divergence of 0. We can therefore sum all the terms and their respective upper bounds:

\begin{align*}
\sum_{k\neq k^\prime}^K p(y=k)p(y=k^\prime) D_f(p(\vec{x}|y=k)\|p(\vec{x}|y=k^\prime)) &\leq \sum_{k\neq k^\prime}^K p(y=k)p(y=k^\prime) (f(0)+g(0))\\
\mathcal{I}_{D_f}^\text{ovo}(\vec{x},y) &\leq \sum_{k=1}^K p(y=k)p(y\neq k)(f(0)+g(0)),
\end{align*}

Following~\cite[theorem 5]{caglar2014divergence}, disjoint supports between the distribution $p(\vec{x}|y=k)$ and $p(\vec{x}|y=k^\prime)$ implies the equality with the upper bound. Assume that the data space $\mathcal{X}$ is separated into $K$ disjoint and supplementary spaces $\mathcal{X}_k$. To each subspace $\mathcal{X}_k$ corresponds a cluster distribution $p(\vec{x}|y=k)$. This disjoint supports are achieved for any model of the form:

\begin{equation*}
p(y=k|\vec{x}) = \pmb{1}[\vec{x}\in\mathcal{X}_k],
\end{equation*}
which implies the disjoint distributions:

\begin{equation*}
p(\vec{x}|y=k) \propto \pmb{1}[\vec{x}\in\mathcal{X}_k]
\end{equation*}

Each of these spaces control the proportion of data in the cluster $k$, and hence controls $p(y=k)$. Thus, the OvO GEMINI is equal to its upper bound owing to disjoint supports:

\begin{equation*}
\mathcal{I}^\text{ovo}_{D_f}(\vec{x},y) = \sum_{k=1}^K p(y=k)p(y\neq k)(f(0)+g(0))
\end{equation*}

We need to maximise the upper bound. This will be the maximum value of OvO GEMINI reachable for models with disjoint supports. Adding a Lagrangian term to respect the contraint of $\sum_{k=1}^K p(y=k)=1$ leads to the optimal solution $p(y=k)=\frac{1}{K}$. This concludes the proof.

\section{Deriving GEMINIs}
\label{app:deriving_geminis}
We show in this appendix how to derive all estimable forms of the GEMINI.

\subsection{\texorpdfstring{$f$}{f}-divergence GEMINI}

We detail here the derivation for 3 $f$-divergences that we previously chose: the KL divergence, the TV distance and the squared Hellinger distance, as well as the generic scenario for any function $f$.

\subsubsection{Generic scenario}

First, we recall that the definition of an $f$-divergence involves a convex function:

\begin{align*}
    f:\mathbb{R}^+ &\rightarrow \mathbb{R}\\
    x&\rightarrow f(x)\\
    \text{s.t.}\quad f(1)&=0,
\end{align*}

between two distributions $p$ and $q$ as described:

\begin{equation*}
    D_\text{f-div}(p,q) = \E_{\vec{x} \sim q} \left[ f\left(\frac{p(\vec{x})}{q(\vec{x})}\right)\right].
\end{equation*}

We simply inject this definition in the OvA GEMINI and directly obtain both an expectation on the cluster assignment $y$ and on the data variable $\vec{x}$. We then merge the writing of the two expectations for the sake of clarity.

\begin{align*}
    \mathcal{I}^\text{ova}_\text{f-div}(x;y) &= \E_{\py} \left[ D_\text{f-div}(\pxy || \pdata)\right],\\
    &= \E_{\py} \left[ \E_{\pdata}\left[ f\left(\frac{\pxy}{\pdata}\right)\right]\right],\\
    &= \E_{\py,\pdata} \left[ f\left(\frac{\pyx}{\py}\right)\right].\\
\end{align*}

Injecting the $f$-divergence in the OvO GEMINI first yields:

\begin{align*}
    \mathcal{I}^\text{ovo}_\text{f-div}(x;y) &= \E_{\pya,\pyb} \left[ D_\text{f-div}(\pxya || \pxyb)\right],\\
    &= \E_{\pya,\pyb} \left[ \E_{\pxyb}\left[ f\left(\frac{\pxya}{\pxyb}\right)\right]\right].\\
\end{align*}

Now, by using Bayes theorem, we can perform the inner expectation over the data distribution. We then merge the expectations for the sake of clarity.

\begin{align*}
    \mathcal{I}^\text{ovo}_\text{f-div}(x;y) &= \E_{\pya,\pyb} \left[ \E_{\pdata} \left[ \frac{\pyxb}{\pyb}f\left(\frac{\pxya}{\pxyb}\right)\right]\right],\\
    &= \E_{\pya,\pyb,\pdata} \left[ \frac{\pyxb}{\pyb}f\left(\frac{\pyxa\pyb}{\pyxb\pya}\right)\right].\\
\end{align*}

Notice that we also changed the ratio of conditional distributions inside the function by a ratio of posteriors through Bayes' theorem, weighted by the relative cluster proportions.

Now, we can derive into details these equations for the 3 $f$-divergences we focused on: the KL divergence, the TV distance and the squared Hellinger distance.

\subsubsection{Kullback-Leibler divergence}

The function for Kullback-Leibler is $f(t) = t\log t$. We do not need to write the OvA equation: it is straightforwardly the usual MI. For the OvO, we inject the function definition by replacing:

\begin{equation*}
    t=\frac{\pyxa\pyb}{\pyxb\pya},
\end{equation*}

in order to get:

\begin{align*}
    \mathcal{I}^\text{ovo}_\text{KL}(x;y) &=  \E_{\pya,\pyb,\pdata} \left[ \frac{\pyxb}{\pyb}\times\frac{\pyxa\pyb}{\pyxb\pya}\log{\frac{\pyxa\pyb}{\pyxb\pya}}\right].
\end{align*}

We can first simplify the factors outside of the logs:

\begin{align*}
    \mathcal{I}^\text{ovo}_\text{KL}(x;y)&=\E_{\pya,\pyb,\pdata} \left[ \frac{\pyxa}{\pya}\log{\frac{\pyxa\pyb}{\pyxb\pya}}\right].
\end{align*}

If we use the properties of the log, we can separate the inner term in two sub-expressions:

\begin{align*}
    \mathcal{I}^\text{ovo}_\text{KL}(x;y) = \E_{\pya,\pyb,\pdata} \left[ \frac{\pyxa}{\pya}\log{\frac{\pyxa}{\pya}} + \frac{\pyxa}{\pya}\log{\frac{\pyb}{\pyxb}}\right].
\end{align*}

Hence, we can use the linearity of the expectation to separate the two terms above. The first term is constant w.r.t. $\yb$, so we can remove this variable from the expectation among the subscripts:

\begin{equation*}
    \mathcal{I}^\text{ovo}_\text{KL}(\vec{x},y) = \E_{\pya,\pdata} \left[ \frac{\pyxa}{\pya}\log{\frac{\pyxa}{\pya}} \right] + \E_{\pya,\pyb,\pdata} \left[ \frac{\pyxa}{\pya}\log{\frac{\pyb}{\pyxb}} \right].
\end{equation*}

Since the variables $\ya$ and $\yb$ are independent, we can use the fact that:

\begin{equation*}
    \E_{\pya}\left[\frac{\pyxa}{\pya}\right] = \int \pya \frac{\pyxa}{\pya} d\ya = 1,
\end{equation*}

inside the second term to reveal the final form of the equation:

\begin{align*}
    \mathcal{I}^\text{ovo}_\text{KL}(x;y) = \E_{\pdata,\py} \left[ \frac{\pyx}{\py}\log{\frac{\pyx}{\py}} \right] + \E_{\pdata,\py} \left[ \log{\frac{\py}{\pyx}}\right].
\end{align*}

Notice that since both terms did not compare one cluster assignment $\ya$ against another $\yb$, we can switch to the same common variable $y$. Both terms are in fact KL divergences depending on the cluster assignment $y$. The first is the reverse of the second. This sum of KL divergences is sometimes called the \emph{symmetric} KL, and so can we write in two ways the OvO KL-GEMINI:

\begin{align*}
    \mathcal{I}^\text{ovo}_\text{KL}(x;y) &= \E_{\pdata} \left[ D_\text{KL} (\pyx || \py)\right] + \E_{\pdata} \left[ D_\text{KL} (\py || \pyx)\right],\\
    &= \E_{\pdata} \left[ D_\text{KL-sym}(\pyx || \py)\right].\\
\end{align*}

We can also think of this equation as the usual MI with an additional term based on the reversed KL divergence.

\subsubsection{Total Variation distance}

For the total variation, the function is $f(t)=\frac{1}{2} |t-1|$. Thus, the OvA GEMINI is:

\begin{equation*}
    \mathcal{I}^\text{ova}_\text{TV}(x;y) = \frac{1}{2}\E_{\py,\pdata} \left[ |\frac{\pyx}{\py}-1|\right].
\end{equation*}

And the OvO is:

\begin{align*}
    \mathcal{I}^\text{ovo}_\text{TV}(x;y) &=\frac{1}{2}\E_{\pya,\pyb,\pdata} \left[ \frac{\pyxb}{\pyb}|\frac{\pyxa\pyb}{\pyxb\pya}-1|\right],\\
    &=\frac{1}{2}\E_{\pya,\pyb,\pdata} \left[ |\frac{\pyxa}{\pya} - \frac{\pyxb}{\pyb} | \right].
\end{align*}

We did not find any further simplification of these equations.

\subsubsection{Squared Hellinger distance}

Finally, the squared Hellinger distance is based on $f(t)=2(1-\sqrt{t})$. Hence the OvA unfolds as:

\begin{align*}
    \mathcal{I}^\text{ova}_{\text{H}^2}(x;y)&= \E_{\py,\pdata} \left[ 2\left(1-\sqrt{\frac{\pyx}{\py}}\right)\right],\\
    &= 2-2\E_{\pdata,\py} \left[\sqrt{\frac{\pyx}{\py}}\right].
\end{align*}

The idea of the squared OvA Hellinger-GEMINI is therefore to minimise the expected square root of the relative certainty between the posterior and cluster proportion.

For the OvO setting, the definition yields:

\begin{equation*}
    \mathcal{I}^\text{ovo}_{\text{H}^2}(x;y)=\E_{\pya,\pyb,\pdata} \left[ \frac{\pyxb}{\pyb}\times 2 \times\left(1-\sqrt{\frac{\pyxa\pyb}{\pyxb\pya}}\right)\right],
\end{equation*}

which we can already simplify by putting the constant 2 outside of the expectation, and by inserting all factors inside the square root before simplifying and separating the expectation:

\begin{align*}
    \mathcal{I}^\text{ovo}_{\text{H}^2}(x;y) &= 2\E_{\pya,\pyb,\pdata} \left[ \frac{\pyxb}{\pyb} - \frac{\pyxb}{\pyb}\sqrt{\frac{\pyxa\pyb}{\pya\pyxb}}\right],\\
    &= 2\E_{\pya,\pyb,\pdata} \left[ \frac{\pyxb}{\pyb} \right] - 2\E_{\pya,\pyb,\pdata} \left[ \sqrt{\frac{\pyxa \pyxb}{\pya \pyb}}\right].
\end{align*}

We can replace the first term by the constant 1, as shown for the OvO KL derivation. Since we can split the square root into the product of two square roots, we can apply twice the expectation over $\ya$ and $\yb$ because these variables are independent:

\begin{equation*}
    \mathcal{I}^\text{ovo}_{\text{H}^2}(x;y) = 2-2\E_{\pdata} \left[ \E_{\py} \left[\sqrt{\frac{\pyx}{\py}}\right]^2\right].
\end{equation*}

To avoid computing this squared expectation, we use the equation of the variance $\mathbb{V}$ to replace it. Thus:

\begin{align*}
    \mathcal{I}^\text{ovo}_{\text{H}^2}(x;y) &= 2-2\E_{\pdata} \left[ \E_{\py}\left[\frac{\pyx}{\py}\right] - \mathbb{V}_{\py}\left[\sqrt{\frac{\pyx}{\py}}\right]\right],\\
    &= 2 - 2\E_{\pdata}\left[ \E_{\py} \left[ \frac{\pyx}{\py}\right]\right] + 2 \E_{\pdata} \left[ \mathbb{V}_{\py} \left[ \sqrt{\frac{\pyx}{\py}}\right]\right].
\end{align*}

Then, for the same reason as before, the second term is worth 1, which cancels the first constant. We therefore end up with:

\begin{equation*}
    \mathcal{I}^\text{ovo}_{\text{H}^2}(x;y)= 2\E_{\pdata} \left[ \mathbb{V}_{\py}\left[\sqrt{\frac{\pyx}{\py}}\right]\right].
\end{equation*}

Similar to the OvO KL case, the OvO squared Hellinger converges to an OvA setting, i.e. we only need information about the cluster distribution itself without comparing it to another. Furthermore, the idea of maximising the variance of the cluster assignments is straightforward for clustering.

\subsection{Maximum Mean Discrepancy}

When using an IPM with a family of functions that project an input of $\mathcal{X}$ to the unit ball of an RKHS $\mathcal{H}$, the IPM becomes the MMD distance.

\begin{align*}
\text{MMD}(p,q) &= \sup_{f: ||f||_\mathcal{H}\leq 1} \E_{\xa\sim p}[f(\xa)] - \E_{\xb \sim q} [f(\xb)],\\
&= \| \E_{\xa\sim p} [\varphi(\xa)] - \E_{\xb \sim q}[\varphi(\xb)]\|_{\mathcal{H}},\\
\end{align*}

where $\varphi$ is a embedding function of the RKHS.

By using a kernel function $\kappa(\xa,\xb) = <\varphi(\xa), \varphi(\xb)>$, we can express the square of this distance thanks to inner product space properties~\cite{gretton_kernel_2012}:

\begin{align*}
\text{MMD}^2 (p,q) &= \E_{\xa, \xa^\prime \sim p}[\kappa(\xa, \xa^\prime)] + \E_{\xb,\xb^\prime \sim q}[\kappa(\xb, \xb^\prime)] - 2 \E_{\xa\sim p, \xb\sim q}[\kappa(\xa, \xb)].
\end{align*}


Now, we can derive each term of this equation using our distributions $p\equiv\pxy$ and $q\equiv \pdata$ for the OvA case, and $p\equiv\pxya, q\equiv\pxyb$ for the OvO case. In both scenarios, we aim at finding an expectation over the data variable $x$ using only the respectively known and estimable terms $\pyx$ and $\py$.

\paragraph{OvA scenario}

For the first term, we use Bayes' theorem twice to get an expectation over two variables $\xa$ and $\xb$ drawn from the data distribution.

\begin{equation*}
    \begin{split}
        \E_{\xa, \xa^\prime \sim p}&= \E_{\xa, \xa^\prime \sim \pxy} \left[ \kappa(\xa,\xa^\prime)\right],\\
        &= \E_{\xa,\xa^\prime \sim \pdata} \left[ \frac{\p(y|\xa)\p(y|\xa^\prime)}{\py^2} \kappa(\xa,\xa^\prime)\right].
    \end{split}
\end{equation*}

For the second term, we do not need to perform anything particular as we directy get an expectation over the data variabes $\xa$ and $\xb$.

\begin{equation*}
    \E_{\xb, \xb^\prime \sim q}= \E_{\xb, \xb^\prime \sim \pdata} \left[ \kappa(\xb,\xb^\prime)\right].
\end{equation*}

The last term only needs Bayes theorem once, for the distribution $q$ is directly replaced by the data distribution $\pdata$:

\begin{align*}
    \E_{\xa\sim p, \xb \sim q}&=\E_{\xa \sim \pxy, \xb \sim \pdata}\left[ \kappa(\xa,\xb)\right],\\
    &= \E_{\xa, \xb \sim px} \left[ \frac{\p(y|\xa)}{\py} \kappa(\xa,\xb)\right].
\end{align*}

Note that for the last term, we could replace $\p(y|\xa)$ by $\p(y|\xb)$; that would not affect the result since $\xa$ and $\xb$ are independently drawn from $\pdata$. We thus replace all terms, and do not forget to put a square root on the entire sum since we have computed so far the squared MMD:

\begin{align*}
        \mathcal{I}_\text{MMD}^\text{ova}(x;y) &= \E_{\py} \left[ \text{MMD}(\pxy,\pdata)\right],\\
        &=\E_{\py} \left[ \left(\E_{\xa,\xa^\prime \sim \pdata} \left[ \frac{\p(y|\xa)\p(y|\xa^\prime)}{\py^2} \kappa(\xa,\xa^\prime)\right] \right.\right.\\
        &\left.\left.\qquad+ \E_{\xb, \xb^\prime \sim \pdata} \left[ \kappa(\xb,\xb^\prime)\right] -2 \E_{\xa, \xb \sim \pdata} \left[ \frac{\p(y|\xa)}{\py} \kappa(\xa,\xb)\right] \right)^{\frac{1}{2}}  \right].\\
\end{align*}

Since all variables $\xa$, $\xa^\prime$, $\xb$ and $\xb^\prime$ are independently drawn from the same distribution $\pdata$, we can replace all of them by the variables $\vec{x}$ and $\vec{x}^\prime$. We then use the linearity of the expectation and factorise by the kernel $\kappa(\vec{x},\vec{x}^\prime)$:

\begin{equation*}
        \mathcal{I}_\text{MMD}^\text{ova}(x;y) = \E_{\py} \left[ \E_{\vec{x},\vec{x}^\prime \sim \pdata}\left[ \kappa(\vec{x},\vec{x}^\prime) \left( \frac{\p(y|\vec{x})\p(y|\vec{x}^\prime)}{\py^2} + 1 - 2\frac{\p(y|\vec{x})}{\py}\right) \right]^{\frac{1}{2}}\right].
\end{equation*}

\paragraph{OvO scenario}

The two first terms of the OvO MMD are the same as the first term of the OvA setting, with a simple subscript $a$ or $b$ at the appropriate place. Only the negative term changes. We once again use Bayes' theorem twice:

\begin{align*}
        \E_{\xa \sim p, \xb \sim q} [\kappa(\xa, \xb)]&= \E_{\xa \sim \pxya, \xb \pxyb} \left[ \kappa(\xa,\xb) \right],\\
        &= \E_{\xa, \xb \sim \pdata} \left[\frac{\p(\ya|\xa)}{\pya}\frac{\p(\yb|\xb)}{\pyb} \kappa(\xa,\xb) \right].
\end{align*}

The final sum is hence similar to the OvA:

\begin{align*}
        \mathcal{I}_\text{MMD}^\text{ovo}(x;y) &= \E_{\pya, \pyb} \left[ \text{MMD} (\pxya, \pxyb)\right],\\
        &= \E_{\pya, \pyb} \left[ \E_{\vec{x},\vec{x}^\prime \sim \pdata} \left[ \kappa(\vec{x},\vec{x}^\prime) \left( \frac{\pyxa \p(\ya|\vec{x}^\prime)}{\pya^2} + \frac{\pyxb \p(\yb|\vec{x}^\prime)}{\pyb^2}\right.\right.\right.\\&\quad\left.\left.\left.-2 \frac{\pyxa \p(\yb|\vec{x}^\prime) }{\pya \pyb} \right) \right]^{\frac{1}{2}}\right].
\end{align*}

\subsection{Wasserstein distance (Proof of Prop.~\ref{prop:wasserstein_convergence})}
\label{app:wasserstein_convergence}

To compute the Wasserstein distance between the distributions $\p(\vec{x}|y=k)$, we estimate it using approximate distributions. We replace $\p(\vec{x}|y=k)$ by a weighted sum of Dirac measures on specific samples $\vec{x}_i$: $p_N^k$:

\begin{equation*}
    \p(\vec{x}|y=k) \approx \sum_{i=1}^N m_i^k\delta_{\vec{x}_i} = p_N^k,
\end{equation*}
where $\{m_i^k\}_{i=1}^N$ is the set of weights. We now show that computing the Wasserstein distance between these approximates converges to the correct distance. We first need to show that $p_N^k$ weakly converges to $p$. To that end, we will use the Portmanteau theorem~\cite{billingsley_convergence_2013}. Let $f$ be any bounded and continuous function. Computing the expectation of such through $\p$ is:

\begin{equation*}
\E_{\vec{x}\sim \p(\vec{x}|y=k)}[f(\vec{x})] = \int_{\mathcal{X}}f(\vec{x})\p(\vec{x}|y=k)d\vec{x},
\end{equation*}
which can be estimated using self-normalised importance sampling~\cite[Chapter 9]{owen_monte_2009}. The proposal distribution we take for sampling is $\pdata$. Although we cannot evaluate both $\pxy$ and $\pdata$ up to a constant, we can evaluate their ratio up to a constant which is sufficient:

\begin{align*}
\E_{\vec{x}\sim \p(\vec{x}|y=k)}[f(\vec{x})]&= \int_{\mathcal{X}}f(\vec{x})\frac{\p(\vec{x}|y=k)}{\pdata}\pdata d\vec{x},\\
&= \int_{\mathcal{X}}f(\vec{x})\frac{\p(y=k|\vec{x})}{\p(y=k)}\pdata d\vec{x},\\
&\approx \sum_{i=1}^N f(\vec{x}_i) \frac{\p(y=k|\vec{x}=\vec{x}_i)}{\sum_{j=1}^N \p(y=k|\vec{x}=\vec{x}_j)}.
\end{align*}

Now, by noticing in the last line that the importance weights are self normalised and add up to 1, we can identify them as the point masses of our previous Dirac approximations:

\begin{equation*}
m_i^k = \frac{\p(y=k|\vec{x}=\vec{x}_i)}{\sum_{j=1}^N \p(y=k|\vec{x}=\vec{x}_j)}.
\end{equation*}

This allows to write that the Monte Carlo estimation through importance sampling of the expectation w.r.t $\p(\vec{x}|y=k)$ is directly the expectation taken on the discrete approximation $p_N^k$. We can conclude that their is a convergence between the two expectations owing to the law of large numbers:

\begin{equation*}
\lim_{N\rightarrow +\infty}\E_{\vec{x}\sim p_N^k}[f(\vec{x})]  = \E_{\vec{x}\sim \p(\vec{x}|y=k)}[f(\vec{x})].
\end{equation*}

Since $f$ is bounded and continuous, the portmanteau theorem~\cite{billingsley_convergence_2013} states that $p_N^k$ weakly converges to $\p(\vec{x}|y=k)$ when defining the importance weights as the normalised predictions cluster-wise.

To conclude, when two series of measures $p_N$ and $q_N$ weakly converge respectively to $p$ and $q$, so does their Wasserstein distance ~\cite[Corollary 6.9]{villani_optimal_2009}, hence:

\begin{equation}
    \lim_{N\rightarrow+\infty}\mathcal{W}_c(p_N^{k_1},p_N^{k_2})= \mathcal{W}_c\left(\p(\vec{x}|y=k_1)\|\p(\vec{x}|y=k_2)\right).
\end{equation}

For the one-vs-all Wasserstein GEMINI, we simply need to set the second distribution to the empirical data distribution: $m_i=1/N$.

\section{Other maximisations of the Wasserstein distances}
\label{app:other_wasserstein_distances}
The Wasserstein-1 metric can be considered as an IPM defined over a set of 1-Lipschitz functions. Indeed, such writing is the dual representation of the Wasserstein-1 metric:

\begin{equation*}
    W_c(p\|q) = \sup_{f, \|f\|_L\leq 1} \E_{x\sim p}[f(x)] - \E_{z\sim q}[f(z)].
\end{equation*}

Yet, evaluating a supremum as an objective to maximise is hardly compatible with the usual backpropagation in neural networks. This definition was used in attempts to stabilise GAN training~\cite{arjovsky_wasserstein_2017} by using 1-Lipschitz neural networks~\cite{gouk_regularisation_2021}. However, the Lipschitz continuity was achieved at the time by weight clipping, whereas other methods such as spectral normalisation~\cite{miyato_spectral_2018} now allow arbitrarily large weights. The restriction of 1-Lipscthiz functions to 1-Lipschitz neural networks does not equal the true Wasserstein distance, and the term "neural net distance" is sometimes preferred~\cite{arora_generalization_2017}. Still, estimating the Wasserstein distance using a set of Lipschitz functions derived from neural networks architectures brings more difficulties to actually leverage the true distance according to the energy cost $c$ of Eq.~\ref{eq:wasserstein_definition}.

Globally, we hardly experimented the generic IPM for GEMINIs. Our efforts for defining a set of 1-Lipschitz critics, one per cluster for OvA or one per pair of clusters for OvO, to perform the neural net distance~\cite{arora_generalization_2017} were not fruitful. This is mainly because such an objective requires the definition of one neural network for the posterior distribution $\pyx$ and $K$ (resp. $K(K-1)/2$) other 1-Lipschitz neural networks for the OvA (resp. OvO) critics, i.e. a large number of parameters. Moreover, this brings the problem of designing not only one but many neural networks while the design of one accurate architecture is already a sufficient problem.

\section{Choosing a GEMINI}
\label{app:exp_complexity}
\begin{figure}
    \centering
    \subfloat[10 samples per batch]{
        \includegraphics[width=0.48\linewidth]{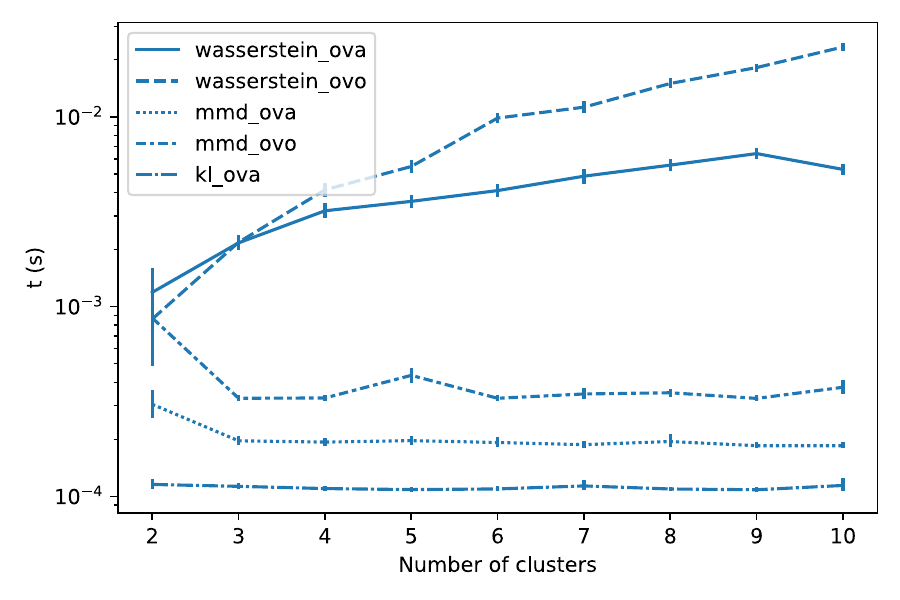}
        \label{sfig:time_performances_batch10}
    }\hfil
    \subfloat[500 samples per batch]{
        \includegraphics[width=0.48\linewidth]{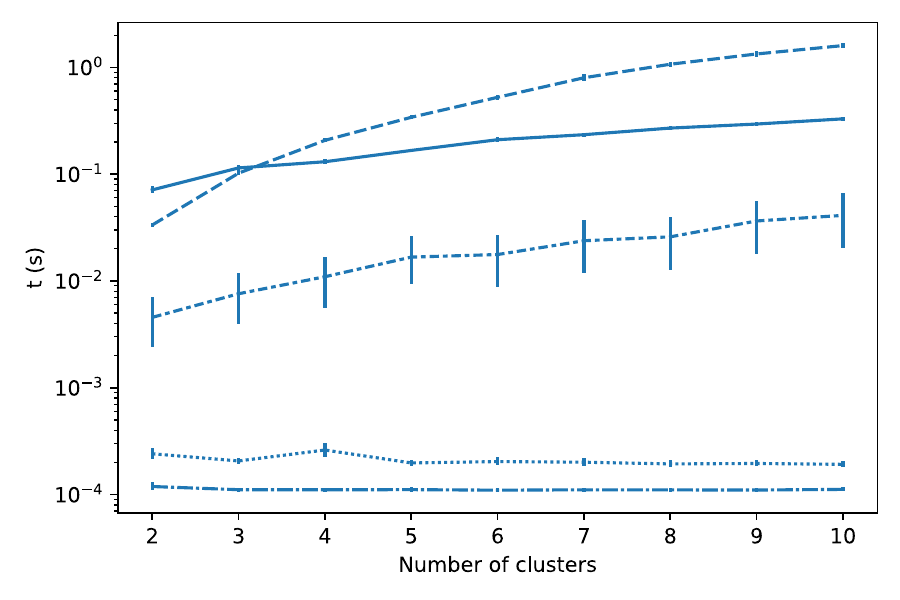}
        \label{sfig:time_performances_batch500}
    }
    \caption{Average performance time (in seconds) of GEMINIs as the number of tasked clusters grows for batches of size 10 and 500 samples.}
    \label{fig:time_performances}
\end{figure}

The complexity of GEMINI increases with the distances previously mentionned depending on the number of clusters $K$ and the number of samples per batch $N$. It ranges from $\mathcal{O}(NK)$ for the usual MI to $\mathcal{O}(K^2N^3\log{N})$ for the OvO Wasserstein-GEMINI. As an example, we show in Figure~\ref{fig:time_performances} the average time of GEMINI as the number of tasked clusters increases for both 10 samples per batch (Figure~\ref{sfig:time_performances_batch10}) and 500 samples (Figure~\ref{sfig:time_performances_batch500}). The batches consists in randomly generated prediction and distances or kernel between randomly generated data.

The OvO Wasserstein is the most complex, and so its usage should remain for 10 clusters or less overall. The second most time-consuming loss is the OvA Wasserstein, however its tendency in optimisation to only find 2 clusters makes it a inappropriate. The main difference also to notice between the following MMD is regarding their memory complexity. The OvA MMD requires only $\mathcal{O}(KN^2)$ while the OvO MMD requires $\mathcal{O}(K^2N^2)$. This memory complexity should be the major guide to choosing one MMD-GEMINI or the other. Thus, the minimal time-consuming and resource-demanding GEMINI is the OvA MMD if we consider GEMINIs that incorporates knowledge of data through kernels and distances. Other versions involving $f$-divergences have in fact the same complexity as MI in our implementations, apart from the OvO TV which reaches $\mathcal{O}(K^2N)$ in our implementation.

\section{Further speed-ups for the Wasserstein-OvO}
\label{app:wasserstein_speedup}
The complexity of the Wasserstein metric is $\mathcal{O}(N^3\log{N})$ for a batch of size $N$. Consequently, the complexity of the OvO Wasserstein reaches $\mathcal{O}(K^2N^3\log{N})$ for $K$ clusters. This implies that this GEMINI is hardly usable for a high number of clusters. To tackle this complexity, we propose instead to sample $M$ independent pairs among the $K(K-1)/2$ pairs of clusters to compare. We evaluate the OvO Wasserstein on these pairs and scale it to the same order as if performed on all pairs. This is stochastically equivalent to maximising the metric for all pairs:

\begin{equation}
    \hat{\I}^\text{ovo}_\mathcal{W}(\vec{x};y) = \frac{2M}{K(K-1)}\times\sum_{k,k^\prime \in I}\p(y=k)\p(y=k^\prime)\times \mathcal{W}_\delta\left(\{m_i^k\}_{i=1}^N, \{m_i^{k^\prime}\}_{i=1}^N \right),
\end{equation}

with

$$I = \{(k_n, k^\prime_n)\}_{n=1}^M\quad; (k_n,k^\prime_n) \sim \mathcal{U}\left(\{1, \cdots, K\}^2 \right),$$ a set of uniformly drawn pairs of clusters.

This optimisation requires however longer training time as a tradeoff for a controlled complexity of $\mathcal{O}(MN^3)$. Note that the same optimisation can be applied to the OvA Wasserstein-GEMINI.

\section{All pair shortest paths distance}
\label{app:fw_distance}
Sometimes, using distances such as the $\ell_2$ may not capture well the shape of manifolds. To do so, we derive a metric using the all pair shortest paths. Simply put, this metric consists in considering the number of closest neighbors that separates two data samples. To compute it, we first use a sub-metric that we note $d$, say the $\ell_2$ norm. This allows us to compute all distances $d_{ij}$ between every sample $i$ and $j$. From this matrix of sub-distances, we can build a graph adjacency matrix $W$ following the rules:

\begin{equation*}
    W_{ij} = \left\{\begin{array}{cr}
        1 & d_{ij}\leq \epsilon \\
        0 & d_{ij}> \epsilon
    \end{array}\right.,
\end{equation*}

where $\epsilon$ is a chosen threshold such that the graph has sparse edges. Our typical choice for $\epsilon$ is the 5\% quantile of all $d_{ij}$.

We chose the graph adjacency matrix to be undirected, owing to the symmetry of $d_{ij}$ and unweighted. Indeed, solving the all-pairs shortest paths involves the Floyd-Warshall algorithm~\cite{warshall_theorem_1962,roy_transitivite_1959} which complexity $\mathcal{O}(n^3)$ is not affordable when the number of samples $n$ becomes large. An undirected and unweighted graph leverages performing $n$ times the breadth-first-search algorithm, yielding a total complexity of $\mathcal{O}(n^2+ne)$ where $e$ is the number of edges. Consequently, setting a good threshold $\epsilon$ controls the complexity of the shortest paths to finds. Our final distance between two nodes $i$ and $j$ is eventually:

\begin{equation}
    c_{ij} = \left\{\begin{array}{c r}\text{Shortest-path}^W(i,j)&\text{if it exists.}\\
    n&\text{otherwise}
    \end{array}\right..
\end{equation}

This metric $c$ can then be incorporated inside the Wasserstein-GEMINI.

\section{Packages for experiments}
\label{app:requirements}
For the implementation details, we use several packages with a python 3.8 version.
\begin{itemize}
    \item We use PyTorch~\cite{pytorch} for all deep learning models and automatic differentiation, as well as NumPy~\cite{numpy} for arrays handling.
    \item We use Python Optimal Transport's function \verb+emd2+\cite{flamary_pot_2021} to compute the Wasserstein distances between weighted sums of Diracs.
    \item We used the implementation of SIMCLR from PyTorch Lightning~\cite{pytorch_lightning}.
    \item Small datasets such as isotropic Gaussian Mixture of score computations are performed using scikit-learn\cite{scikit-learn}.
\end{itemize}

\end{document}